\documentclass[10pt,twocolumn,letterpaper]{article}

\usepackage[pagenumbers]{wacv} % To force page numbers, e.g. for an arXiv version

\usepackage{graphicx}
\usepackage{amsmath}
\usepackage{amssymb}
\usepackage{booktabs}
\usepackage{multirow}
\usepackage{tabularx}
\usepackage{amssymb}
\usepackage{csquotes}
\usepackage{pifont}
\usepackage{stfloats}
\usepackage[pagebackref,breaklinks,colorlinks,allcolors=wacvblue]{hyperref}

\definecolor{wacvblue}{rgb}{0.21,0.49,0.74}

\title{NovaSplat: Model-Free 6D Object Pose Estimation via 3D Gaussian Splatting Onboarding and Refinement}

\author{
    Georgios Floros \qquad Evangelos Sartinas \qquad Orestis Vaggelis \qquad Maria Pateraki \\
    \and
    National Technical University of Athens\\
    {\tt\small \{gfloros, vsartinas, orestisvaggelis, mpateraki\}@mail.ntua.gr}
}

\newcommand{\PAR}[1]{\vskip4pt \noindent {\bf #1~}}
\newcommand{\cmark}{\ding{51}}%
\newcommand{\xmark}{\ding{55}}%

\begin{document}
\maketitle
{
    \begin{abstract}
6D pose estimation of novel objects without CAD models remains an open challenge: model-free methods must build the object representation from scratch during onboarding and have long trailed their model-based counterparts. We present NovaSplat, a model-free pipeline that closes this gap. From a short smartphone video captured at onboarding, we train a 3D Gaussian Splatting (3DGS) model that serves both as a template source for detection and coarse pose estimation and as a differentiable backbone for pose refinement. Our refiner optimizes the object-to-camera pose with combined photometric and geometric losses under a continuous rotation parametrization free of gradient singularities. On a household-object benchmark, NovaSplat surpasses all model-based methods. On a challenging industrial benchmark with textureless and specular surfaces, it outperforms all model-free competitors while substantially narrowing the gap to model-based ones. We further introduce NovaSplat-Lite, an inference variant that is $3.1\times$ faster at negligible accuracy cost. Code will be made available at \href{https://github.com/POSE-Lab/novasplat}{https://github.com/POSE-Lab/novasplat}.
\end{abstract}
}
{
    \section{Introduction}
\label{sec:introduction} 

6D object pose estimation, recovering the full position and orientation of an object relative to a camera, is a core capability for robotic manipulation, augmented reality, and industrial inspection. The highest-performing methods rely on a 3D CAD model of the object: they render templates at many viewpoints and extract geometric point clouds, which are then matched against the query image using powerful frozen foundation models~\cite{Oernek_ECCV_2024,Caraffa_ECCV_2024}. In practice, however, CAD models are rarely available outside controlled environments, making these approaches impractical for real-world deployment. This motivates the \emph{model-free} setting: given only a short video of a novel object captured during an onboarding phase, estimate its 6D pose in new scenes without any 3D model.

\begin{figure}[t]
\centering
\includegraphics[width=\linewidth]{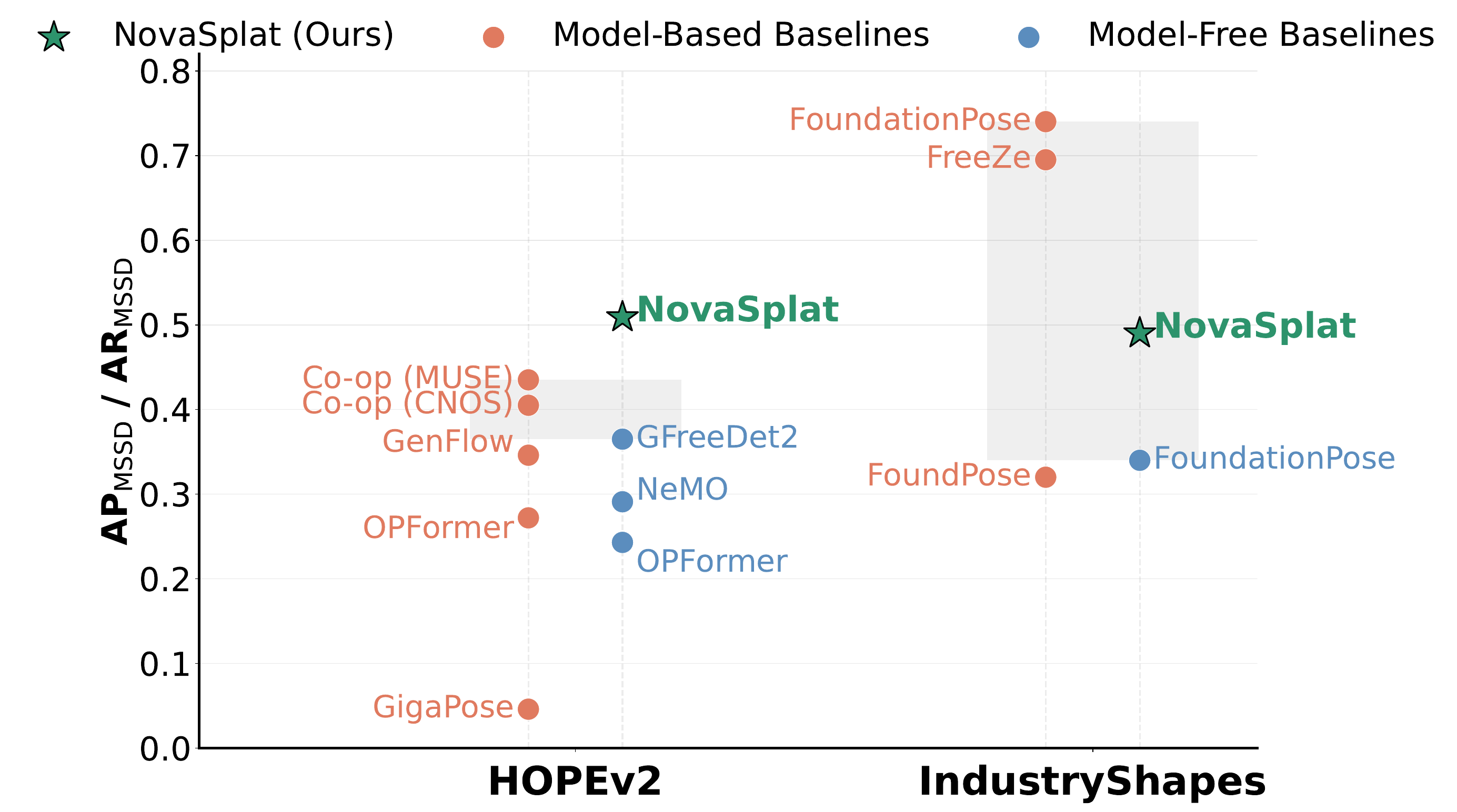}
\caption{AP$_\text{MSSD}$ (\textbf{HOPEv2}) and AR$_\text{MSSD}$ (\textbf{IndustryShapes}, averaged over Classic and Extended splits) for model-free (\textcolor[HTML]{5B8DBE}{blue}) and model-based (\textcolor[HTML]{E07A5F}{orange}) methods. The shaded region marks the gap between the best model-free baseline and the best model-based method. NovaSplat (\textcolor[HTML]{2D936C}{green}) surpasses all model-based methods on HOPEv2, effectively closing the gap on that benchmark. In IndustryShapes NovaSplat sits within the gap, outperforming all model-free competitors while narrowing the distance to model-based methods.}
\label{fig:bridging_gap}
\end{figure}

Model-free methods face a fundamental reconstruction problem. Where model-based methods receive the object geometry by definition, model-free methods must recover it from video, a noisier and less complete substitute that degrades every downstream stage, from template rendering to correspondence matching. As illustrated in Figure~\ref{fig:bridging_gap}, this has historically resulted in a substantial performance gap between the two paradigms, limiting the practical utility of model-free approaches.

We propose \textbf{NovaSplat}, a complete model-free pipeline that closes this gap and in some cases reverses it. During onboarding, we reconstruct a dense 3D representation of the object from video and train a 3D Gaussian Splatting (3DGS) model. This representation serves two roles simultaneously: rendering high-quality templates for object detection and coarse pose estimation and acting as a differentiable backbone for gradient-based pose refinement. Our 3DGS refiner freezes the learned scene representation and directly optimizes the object-to-camera pose by jointly minimizing appearance and geometric losses, using a continuous rotation parameterization that avoids singularities. This joint optimization proves considerably more robust than geometry-only refinement under the noisy reconstructions typical of video-based onboarding.

We make the following contributions:

\begin{itemize}
    \item A complete model-free onboarding pipeline that produces a high-quality 3DGS object representation from a short video, enabling model-based pose estimation methods to operate without a CAD model.
    \item A novel 3DGS-based pose refiner that jointly optimizes appearance and geometry, advancing model-free pose refinement.
    \item State-of-the-art results on two benchmarks spanning household and industrial objects, demonstrating that model-free methods can match and exceed model-based performance without CAD models.
\end{itemize}
}
{
    \section{Related Work}
\label{sec:related-work}

\subsection{Taxonomy of 6D Object Pose Estimation}
\label{sec:taxonomy}

Existing 6D object pose estimation methods can be broadly categorized by the level of prior object knowledge they require. \textbf{Instance-based methods}~\cite{Hodan_CVPR_2020, Tremblay_CORL_2018, Su_CVPR_2022} leverage 3D object models during training to achieve high accuracy but are restricted to previously seen objects. \textbf{Novel object methods}~\cite{Oernek_ECCV_2024, Caraffa_ECCV_2024, Nguyen_CVPR_2024, Moon_CVPR_2025, He_NEURIPS_2022, Liu_ECCV_2022, Jung_3DV_2026, Lee_CVPR_2025, Li_ICCVW_2023, Wen_CVPR_2023, Wen_CVPR_2024, Cao_CVPR_2025, Wang_ICRA_2026} generalize to unseen objects and can be further divided into category-based~\cite{Wang_CVPR_2019, Tian_ECCV_2020} and unseen object approaches. Within the unseen object paradigm, \textbf{model-based methods}~\cite{Oernek_ECCV_2024, Caraffa_ECCV_2024, Nguyen_CVPR_2024, Moon_CVPR_2024, Moon_CVPR_2025} require a CAD model at inference time, exploiting geometric priors from rendered templates and 3D point clouds, while \textbf{model-free methods}~\cite{ Sun_CVPR_2022, Wang_ICRA_2026} build an object representation from a short video sequence during an \emph{onboarding} phase, without any 3D model. This work focuses on the model-free unseen object setting.

\subsection{Model-Based Novel Object Methods}
\label{sec:model-based-methods}

Several recent methods follow a correspondence-based pipeline over CAD-rendered templates. FoundPose~\cite{Oernek_ECCV_2024} matches DINOv2~\cite{Oquab_TMLR_2024} descriptors between the query and rendered templates but relies solely on visual features. FreeZe~\cite{Caraffa_ECCV_2024} and FreeZeV2~\cite{Caraffa_ARXIV_2025} extend to depth input by fusing visual and geometric (GeDi~\cite{Poiesi_PAMI_23}) features for 3D-3D correspondence matching and refine the final pose using the Iterative Closest Point (ICP) algorithm, ranking among the top BOP methods. We build directly on FreeZeV2 as our coarse pose estimator. GigaPose~\cite{Nguyen_CVPR_2024} is fast and supports both model-based and model-free settings on single RGB frames, though lacking depth limits its performance on small or occluded objects. GenFlow~\cite{Moon_CVPR_2024} and its successor Co-op~\cite{Moon_CVPR_2025} refine poses via optical flow and semi-dense correspondences, respectively, the latter using a CroCo-based~\cite{Weinzaepfel_ICCV_2023} ViT with a differentiable PnP layer. Pos3R~\cite{Deng_CVPR_2025} likewise matches over CAD-rendered templates, employing MASt3R~\cite{Leroy_ECCV_2024}, a 3D reconstruction foundation model, to establish dense correspondences, select the best template by feature similarity, and recover pose via PnP-RANSAC, with optional refinement via render-and-compare. WALDO~\cite{Pakdamansavoji_WACV_2026} extends the detect-segment-hypothesize-refine pipeline with dynamic non-uniform sampling, multi-hypothesis inference, and iterative refinement to exhibit robustness under occlusions. All these methods require a 3D CAD model at inference time. NovaSplat operates without one at any stage, replaces ICP with a differentiable 3DGS-based refiner, and surpasses those of them evaluated on HOPEv2 (Table~1).

\subsection{Model-Free Novel Object Methods}
\label{sec:model-free-methods}

Model-free methods can be organized into three categories. \emph{Visual localization} approaches~\cite{Sun_CVPR_2022, He_NEURIPS_2022} apply Structure-from-Motion (SfM) during onboarding to build a sparse 3D point cloud, recovering pose via 2D-3D correspondences. They struggle, however, with low-texture objects and large viewpoint changes. \emph{Template-based} methods~\cite{Liu_ECCV_2022} use viewpoint retrieval over posed reference images for a coarse hypothesis before refinement but are sensitive to detection and bounding box quality. \emph{Few-shot} methods operate from only one or a handful of RGB or RGB-D images:~\cite{Jung_3DV_2026} generate a geometry-aware encoding from a few RGB views,~\cite{Liu_CVPR_2025} compress a single RGB-D reference into a pose embedding via latent diffusion, and~\cite{Lee_CVPR_2025} combine 2D-3D alignment with render-and-compare refinement from a single RGB-D anchor. Despite their practical appeal, these methods rely on strong assumptions, such as symmetry priors or scale constraints, to compensate for limited onboarding data, restricting generalization across object geometries.

\subsection{Neural Rendering for Object Pose Estimation}
\label{sec:neural-rendering-methods}

\emph{Neural rendering} approaches encode objects into differentiable implicit representations for CAD-free novel-view synthesis. Early NeRF-based methods such as LatentFusion~\cite{Park_CVPR_2020} and NeRF-Pose~\cite{Li_ICCVW_2023} use latent voxel grids and radiance fields, respectively. BundleSDF~\cite{Wen_CVPR_2023} builds a Neural Object Field on-the-fly from an RGB-D stream, concurrently optimizing pose and a 3D representation. FoundationPose~\cite{Wen_CVPR_2024} extends this into a unified foundation model supporting both model-based and model-free setups via large-scale synthetic training. More recently, 3D Gaussian Splatting (3DGS)~\cite{Kerbl_TOG_2023} has largely displaced NeRF-based representations owing to superior rendering quality and real-time speed. GS-Pose~\cite{Cai_3DV_2025}, iG-6DoF~\cite{Cao_CVPR_2025}, and 6DOPE-GS~\cite{Jin_2025_ICCV} all leverage 3DGS for render-and-compare refinement or online tracking but evaluate primarily on simpler benchmarks (Linemod~\cite{Hinterstoisser_ACCV_12}, HO3D~\cite{Hampali_CVPR_2020}, YCBInEOAT~\cite{Wen_IROS_2020}) without testing on challenging multi-object scenes such as HOPEv2. The work that is most closely related to ours is GFreeDet2~\cite{Wang_ICRA_2026}, which combines a 3DGS onboarding pipeline with template-based coarse pose estimation but relies exclusively on visual features, lacking the geometric integration we show is critical for accurate pose estimation on challenging objects.
}
{
    \section{Method}
\label{sec:method} 

\begin{figure*}[t]
  \centering
  \includegraphics[width=\linewidth]{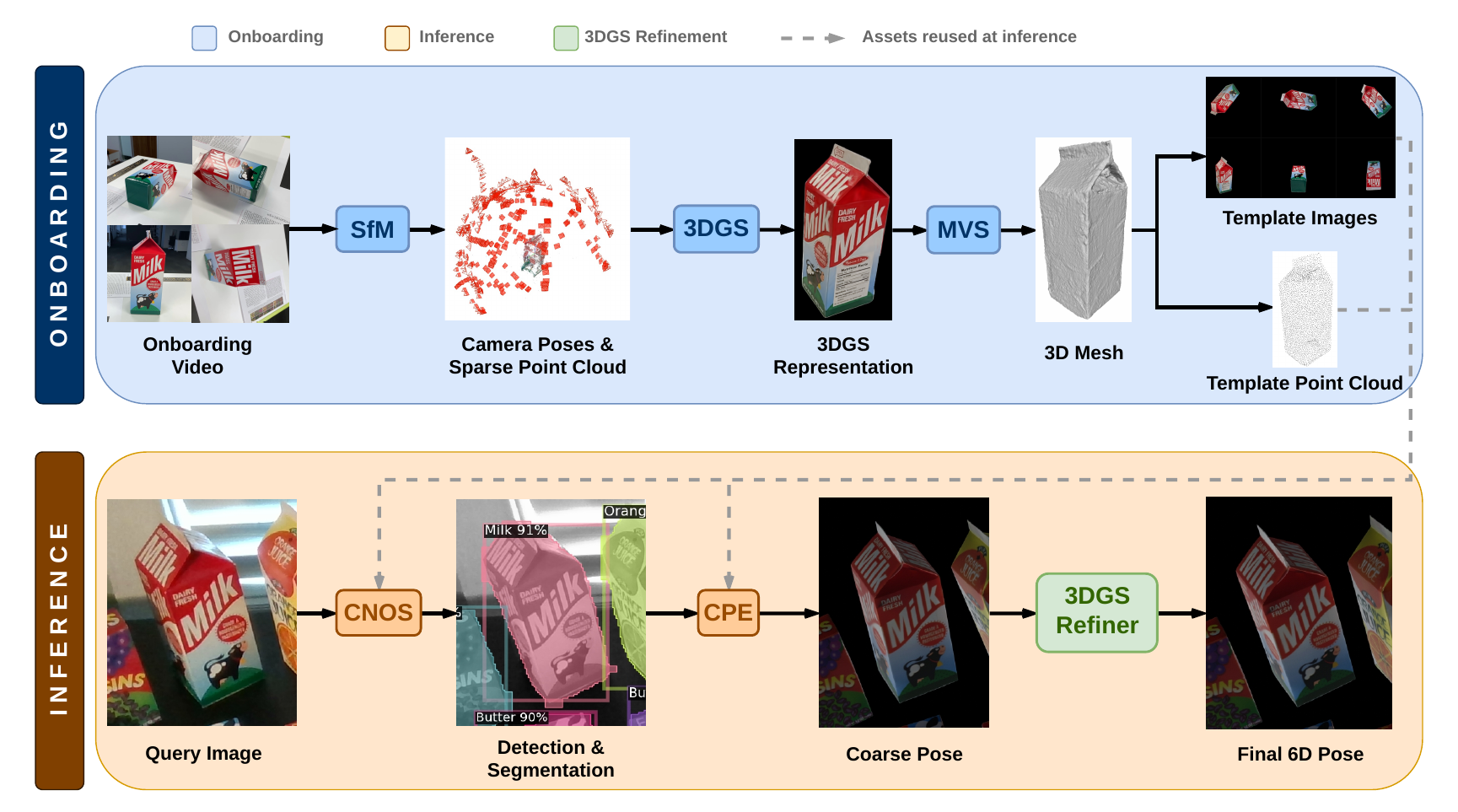}
  \caption{\textbf{NovaSplat pipeline.}
  During \textit{onboarding} (top), a short smartphone video is processed through SfM, 3DGS training, and MVS to produce a 3DGS representation and a 3D mesh. Template images $\{\mathcal{I}_{\text{templ}}^i\}_{i=1}^{N_{\text{templ}}}$ are rendered from the 3DGS model at icosphere-sampled viewpoints, and a template point cloud $\mathcal{T}$ is sampled from the mesh via Poisson disk sampling. During \textit{inference} (bottom), a query image is processed by a modified CNOS module to detect the object and produce a segmentation mask $M_{\text{seg}}$. A FreeZe-based Coarse Pose Estimator (CPE) uses $\{\mathcal{I}_{\text{templ}}^i\}_{i=1}^{N_{\text{templ}}}$ and $\mathcal{T}$ for joint visual and geometric feature matching to produce $P_{\text{coarse}}$. Our 3DGS-based refiner then optimizes $P_{\text{coarse}}$ via a photometric loss $\mathcal{L}_{\text{photo}}$ and a geometric Chamfer loss $\mathcal{L}_{\text{geo}}$ to yield the final 6D pose. Dashed arrows indicate onboarding assets reused at inference.}
  \label{fig:pipeline}
\end{figure*}

\subsection{Overview}
\label{sec:overview}

Our pipeline consists of two stages: \textbf{onboarding} and \textbf{inference} (Figure~\ref{fig:pipeline}).

During \textit{onboarding} (Section~\ref{ref:onboarding}), the goal is to build an object representation from a short video of the object captured from multiple viewpoints with a handheld device such as a smartphone. Given the video frames and corresponding object masks, we run a classical 3D reconstruction pipeline comprising SfM, Multi-View Stereo (MVS), and 3DGS training to produce a 3D mesh and an optimized 3DGS object representation.

During \textit{inference}, given an RGB-D query image containing instances of onboarded objects, the goal is to detect each object and estimate its 6D pose. We first run a modified CNOS~\cite{Nguyen_ICCVW_2023} detection and segmentation module (Section~\ref{ref:detection_segmentation}) to localize each object and produce its segmentation mask. We then employ the coarse pose estimation module of FreeZeV2~\cite{Caraffa_ECCV_2024,Caraffa_ARXIV_2025} (Section~\ref{ref:coarse_pose_estimation}), adapted to use templates rendered from the 3DGS representation and a 3D point cloud sampled from the mesh model, to obtain an initial 6D pose estimate. Finally, each coarse pose is passed to our 3DGS-based hybrid refiner (Section~\ref{sec:refinement}), which produces the final 6D pose. Implementation details of the onboarding and inference stages are provided in Section~\ref{sec:implementation}. 

\subsection{Onboarding}
\label{ref:onboarding}

We assume access to a set of RGB video frames depicting the object from all viewpoints, along with corresponding object segmentation masks. When masks are unavailable, they can be obtained automatically using an off-the-shelf video segmentation model such as SAM3~\cite{Carion_ARXIV_2025}.

\PAR{Structure-from-Motion.}
The raw frames are passed to a SfM pipeline~\cite{Schoenberger_CVPR_2016, Pan_ECCV_2024} to estimate camera poses and a sparse 3D point cloud. The object masks are then applied to isolate the object point cloud from the background.

\PAR{3D Gaussian Splatting.}
The masked frames, camera poses, and object point cloud are used to train a 3DGS model~\cite{Kerbl_TOG_2023}. The 3DGS representation consists of a set of 3D Gaussians, each parameterized by its mean, covariance, opacity, and view-dependent color encoded as Spherical Harmonics (SH) coefficients. The model is optimized by differentiably rendering the Gaussians onto the image plane and minimizing the photometric error against the masked training frames.

\PAR{Multi-View Stereo.}
The SfM point cloud is often too sparse to initialize stereo depth estimation reliably, particularly over untextured object regions where few or no feature correspondences can be established. Since 3DGS densification fills in these regions during optimization, we exploit the Gaussian centers as a semi-dense point cloud. Equipped with camera visibility information, this denser prior serves as input to our MVS pipeline. We run PatchMatchStereo~\cite{Bleyer_BMVC_2011} to estimate per-view depth maps from neighboring views, which are then fused into a dense point cloud. From this point cloud we recover a 3D mesh by following~\cite{Jancosek_ISRN_2014}, which discretizes space via Delaunay tetrahedralization and formulates surface extraction as a graph optimization problem that incorporates visibility constraints.

\PAR{Outputs.}
At the end of onboarding we have two assets: a 3DGS representation from which novel views of the object can be rendered, and a 3D mesh that captures the object geometry. From the 3DGS representation, we render $N_{\text{templ}}$ template images $\{\mathcal{I}_{\text{templ}}^i\}_{i=1}^{N_{\text{templ}}}$ by sampling camera poses on an icosphere centered at the object. From the mesh, we sample $N_{\text{pts}}$ 3D points via Poisson disk sampling~\cite{Bridson_SIGGRAPH_2007}, forming the template point cloud $\mathcal{T}$. Both $\{\mathcal{I}_{\text{templ}}^i\}_{i=1}^{N_{\text{templ}}}$ and $\mathcal{T}$ are passed to the downstream inference stage (see Figure~\ref{fig:pipeline}).

\subsection{Detection and Segmentation}
\label{ref:detection_segmentation}

We follow the three-stage propose-then-match workflow of CNOS~\cite{Nguyen_ICCVW_2023}\footnote{\url{https://github.com/Vision-Kek/cnos25}}, extending it in three respects. First, we replace SAM~\cite{Kirilov_ICCV_2023} / FastSAM~\cite{Zhao_ARXIV_2023} with YOLO-E~\cite{Wang_ICCV_2025} as the proposal model. Second, we replace DINOv2~\cite{Oquab_TMLR_2024} with DINOv3 ViT-L/16~\cite{Simeoni_ARXIV_2025} as the feature extractor. Third, rather than using CAD-rendered templates or masked frames from the onboarding video as done by other model-free methods, we use our rendered templates $\{\mathcal{I}_{\text{templ}}^i\}_{i=1}^{N_{\text{templ}}}$ (Section~\ref{ref:onboarding}).

\subsection{Coarse Pose Estimation}
\label{ref:coarse_pose_estimation}

We build our coarse pose estimation module on the workflow of FreeZeV2~\cite{Caraffa_ARXIV_2025}, adapting it to our model-free setting. Visual features are extracted from $\{\mathcal{I}_{\text{templ}}^i\}_{i=1}^{N_{\text{templ}}}$ using a frozen DINOv3 encoder, while geometric features are extracted from $\mathcal{T}$ (Section~\ref{ref:onboarding}) using a frozen GeDi encoder~\cite{Poiesi_PAMI_23}. Both feature types are extracted on the template and query sides, then fused into a unified per-point descriptor that encodes appearance and geometry jointly. Matching these descriptors yields a set of 3D-3D correspondences, which are passed to a RANSAC framework to produce the coarse pose $P_{\text{coarse}}$.

Compared to the original FreeZe pipeline, we relocate the symmetry-aware augmentation from the refinement stage to the coarse pose estimation stage. At this stage the full set of $N_{\text{templ}}$ views is available, providing substantially more viewpoint variation for the augmentation to exploit than the single rendered view used during refinement.

\subsection{3DGS-Based Pose Refinement}
\label{sec:refinement}

The coarse pose $P_{\text{coarse}}$ is further refined in the final stage of our pipeline through a hybrid optimization scheme that jointly considers appearance and geometric cues. The backbone of this optimization is the differentiable 3DGS representation learned during onboarding. During refinement, the Gaussian parameters are frozen, and only the object-to-camera pose is optimized.

We parameterize the rotational component using the continuous 6D representation of Zhou~\etal~\cite{Zhou_CVPR_2019}, which is defined over $SO(3)$ without singularities or discontinuities, allowing gradients to flow freely throughout the optimization. The translational component $\mathbf{t}$ is parameterized as a standard 3D vector. Both are initialized from $P_{\text{coarse}}$.

At each iteration, we render an RGB image $I_{\text{rend}}$ and a binary mask $M_{\text{rend}}$ obtained by thresholding the alpha channel at $\tau_{\alpha}$.

\PAR{Photometric Loss.}
Following GS-Pose~\cite{Cai_3DV_2025}, we define the photometric loss as the sum of the image structural similarity (SSIM)~\cite{Zhou_TIP_2004} and the multi-scale SSIM (MS-SSIM)~\cite{Wang_ACSSC_2003}, computed over the region defined by $M_{\text{rend}}$:
\begin{equation}
    \mathcal{L}_{\text{photo}} = (1 - \text{SSIM}) + (1 - \text{MS-SSIM}).
    \label{eq:photo_loss}
\end{equation}
We use 3 scales for MS-SSIM rather than the standard 5, as the latter requires a minimum image resolution that small rendered objects often fail to satisfy.

\PAR{Geometric Loss.}
We compute a bidirectional Chamfer distance between the template point cloud $\mathcal{T}$ (Section~\ref{ref:onboarding}) and a query point cloud $\mathcal{Q}$. $\mathcal{Q}$ is backprojected from the depth channel of the query image, retaining only points that fall within the segmentation mask $M_{\text{seg}}$ and whose depth lies within a range $d_{\text{range}}$ of the transformed $\mathcal{T}$. This filtering excludes table surfaces, neighboring objects, and background clutter from the query point cloud. The geometric loss is:
\begin{equation}
    \mathcal{L}_{\text{geo}} = \frac{1}{|\mathcal{T}|}\sum_{t \in \mathcal{T}} \min_{q \in \mathcal{Q}} \|t - q\|^2 + \frac{1}{|\mathcal{Q}|}\sum_{q \in \mathcal{Q}} \min_{t \in \mathcal{T}} \|q - t\|^2.
    \label{eq:geo_loss}
\end{equation}

\PAR{Total Loss.}
The two losses are combined as:
\begin{equation}
    \mathcal{L} = (1 - w_{\text{geo}}) \cdot \mathcal{L}_{\text{photo}} + w_{\text{geo}} \cdot \mathcal{L}_{\text{geo}},
    \label{eq:total_loss}
\end{equation}
where $w_{\text{geo}}$ is the geometric loss weight.

\PAR{Optimization.}
We use separate AdamW optimizers for rotational and translational components, as we found that a shared learning rate was too restrictive for the translation parameters when tuned appropriately for rotation. The two optimizers alternate every $N_{\text{alt}}$ iterations. Per-iteration corrections are clamped to an angular threshold $\alpha_{\text{thresh}}$ for rotation and $t_{\text{thresh}}$ for translation to prevent large updates from destabilizing the optimization. We retain the pose with the lowest total loss over $N_{\text{ref}}$ iterations and apply early stopping if no improvement occurs for $N_{\text{halt}}$ iterations.
}
{
    \section{Experiments}
\label{sec:experiments}

\subsection{Datasets}
\label{sec:datasets}

We evaluate NovaSplat on two complementary benchmarks spanning household and industrial objects.
 
\PAR{HOPEv2}~\cite{Tyree_IROS_2022} is the BOP version of NVIDIA's Household Objects for Pose Estimation dataset, comprising 28 toy grocery objects across 50 cluttered scenes from 10 household and office environments. For the BOP Challenge 2024~\cite{Nguyen_CVPRW_2025}, the dataset was extended with multi-view test images captured by a moving camera and provides a standardized leaderboard for direct comparison with both model-free and model-based methods under identical conditions. We note that we encountered some inconsistencies in this dataset, which are detailed in Sec.~\ref{sec:hopev2_onboarding_data_inconsistencies}.
 
\PAR{IndustryShapes}~\cite{Sapoutzoglou_ICRA_2026} is an RGB-D benchmark designed for industrial robotics, with two splits: \emph{classic} and \emph{extended}. It features five object categories with properties that are particularly challenging for pose estimation: textureless surfaces, specular reflections, thin geometries, and small scale. Scenes are captured in realistic industrial assembly settings with multiple instances per object and inter-object occlusions. The \emph{extended} split provides onboarding sequences, directly matching our model-free paradigm.

\PAR{Scope of Evaluation.} Several concurrent 3DGS-based pose estimation methods~\cite{Cai_3DV_2025, Cao_CVPR_2025} report results on Linemod (LM)~\cite{Hinterstoisser_ACCV_12}. We deliberately exclude it for three reasons. First and foremost, LM provides no onboarding videos, which are a fundamental requirement of our model-free setting and cannot be substituted by the reference images the dataset supplies. Second, its capture protocol covers only a hemisphere around each object, precluding the full 360° reconstruction that NovaSplat relies on. Third, state-of-the-art methods~\cite{Mei_ARXIV_2024} already exceed 95\% on LM, leaving little headroom to meaningfully differentiate approaches. Among the BOP-H3 benchmarks, only HOPEv2 supplies test-set depth (HOT3D and HANDAL are RGB-only), and IndustryShapes is the only other BOP-format dataset we are aware of that provides both onboarding sequences and depth, making the two the appropriate testbeds here.
\begin{table}[t]
\centering
\setlength{\tabcolsep}{4pt}
\begin{tabular}{@{}lcccc@{}}
\toprule
Method & CAD & AP $\uparrow$ & AP$_\text{MSSD}$ $\uparrow$ & AP$_\text{MSPD}$ $\uparrow$ \\
\midrule
GigaPose~\cite{Nguyen_CVPR_2024}         & \cmark & 0.167          & 0.046          & 0.288 \\
OPFormer*~\cite{Moroz_WACV_2026}         & \cmark & 0.351          & 0.272          & 0.431 \\
GenFlow~\cite{Moon_CVPR_2024}            & \cmark & 0.411          & 0.346          & 0.475 \\
Co-op (CNOS)~\cite{Moon_CVPR_2025}       & \cmark & 0.477          & 0.405          & 0.549 \\
Co-op (MUSE)~\cite{Moon_CVPR_2025}       & \cmark & \textbf{0.514} & \textbf{0.435} & \textbf{0.593} \\
\midrule
NeMO~\cite{Jung_3DV_2026}               & \xmark  & 0.302          & 0.291          & 0.312 \\
OPFormer*~\cite{Moroz_WACV_2026}        & \xmark  & 0.335          & 0.243          & 0.427 \\
GFreeDet2~\cite{Wang_ICRA_2026}         & \xmark  & 0.452          & 0.365          & 0.538 \\
\midrule
NovaSplat                               & \xmark & \textbf{0.529}  & \textbf{0.509} & \textbf{0.548} \\
NovaSplat-Lite                          & \xmark & 0.528           & 0.509          & 0.547 \\
\bottomrule
\end{tabular}
\caption{Quantitative results on \textbf{HOPEv2}. Model-based methods use CAD models at test time. Model-free methods use only onboarding videos. Best model-based and model-free results appear in \textbf{bold}. *Results taken from the BOP leaderboard. \cite{Moroz_WACV_2026} reports slightly higher numbers that do not affect the ranking. NovaSplat‑Lite is our fast inference variant (Sec.~\ref{sec:implementation}). It uses the same onboarding and object representation and differs only in inference cost.}
\label{tab:hopev2_results}
\end{table}

\subsection{Evaluation Metrics}
\label{sec:metrics}
For each of the benchmarks, we follow the proposed evaluation protocol to ensure fair comparison with the reported baselines. In particular, for HOPEv2, we adopt the BOP detection evaluation protocol~\cite{Nguyen_CVPRW_2025}, computing Average Precision ($\text{AP}$ or $\text{mAP}$) for the Maximum Symmetry-aware Surface Distance (MSSD) $\text{AP}_\text{MSSD}$ and the Maximum Symmetry-aware Projection Distance (MSPD) $\text{AP}_\text{MSPD}$. For IndustryShapes, we adopt the BOP localization protocol, computing the Average Recall ($\text{AR}$) for MSSD ($\text{AR}_\text{MSSD}$), for MSPD ($\text{AR}_\text{MSPD}$) and for the Visible Surface Discrepancy (VSD) $\text{AR}_\text{VSD}$. The overall accuracy is measured for both datasets as an average over the different metrics. We refer to Sec. 3.4 of~\cite{Nguyen_CVPRW_2025} for details.

\begin{figure*}[t]
\centering
\setlength{\tabcolsep}{1pt}
\begin{tabular}{ccccc}
\includegraphics[width=0.196\linewidth]{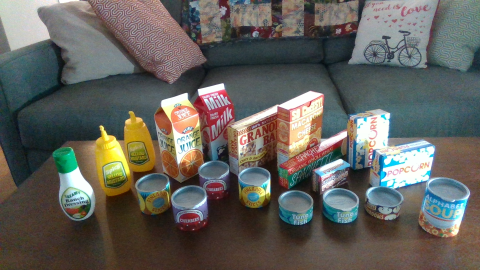} &
\includegraphics[width=0.196\linewidth]{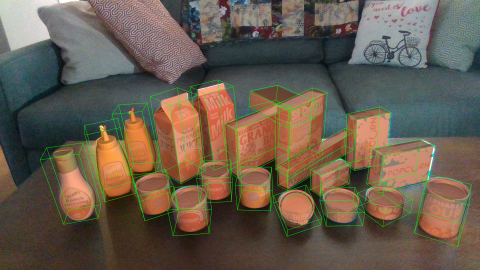} &
\includegraphics[width=0.196\linewidth]{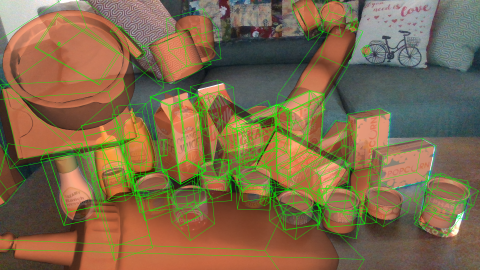} &
\includegraphics[width=0.196\linewidth]{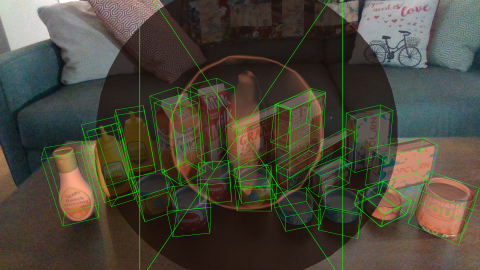} &
\includegraphics[width=0.196\linewidth]{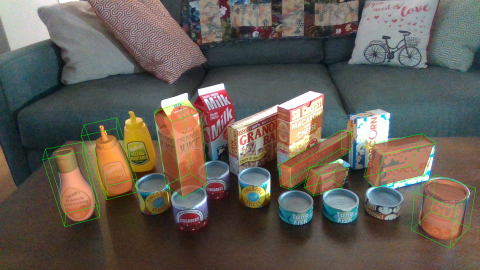} \\
\includegraphics[width=0.196\linewidth]{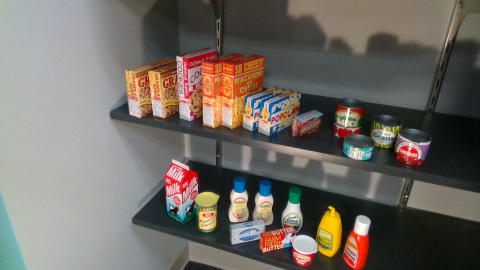} &
\includegraphics[width=0.196\linewidth]{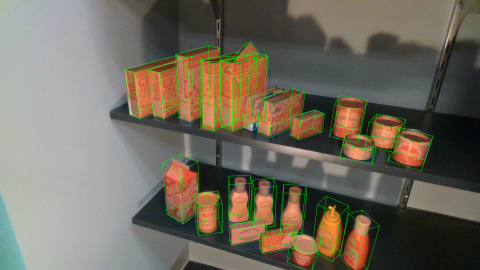} &
\includegraphics[width=0.196\linewidth]{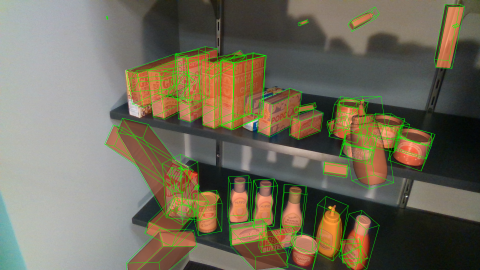} &
\includegraphics[width=0.196\linewidth]{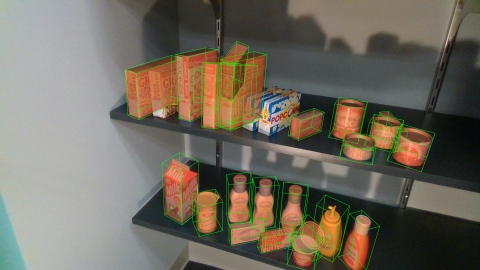} &
\includegraphics[width=0.196\linewidth]{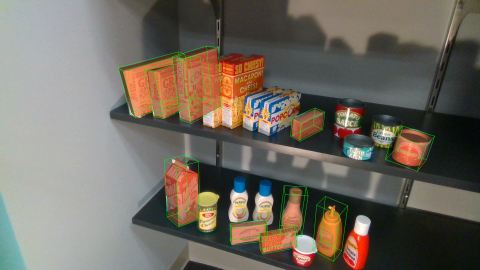} \\
\includegraphics[width=0.196\linewidth]{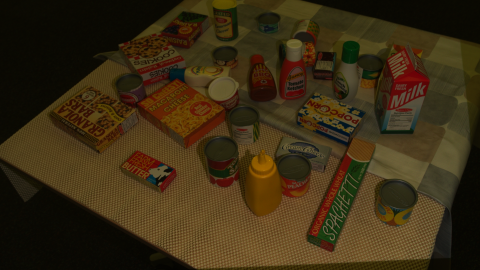} &
\includegraphics[width=0.196\linewidth]{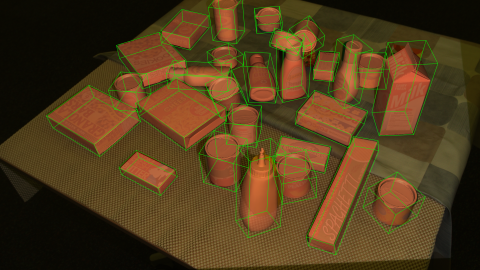} &
\includegraphics[width=0.196\linewidth]{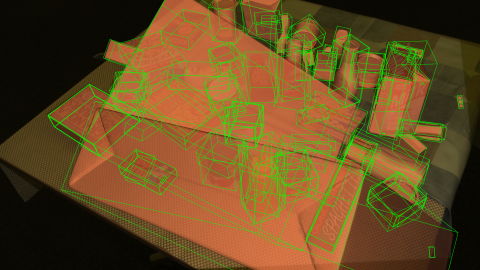} &
\includegraphics[width=0.196\linewidth]{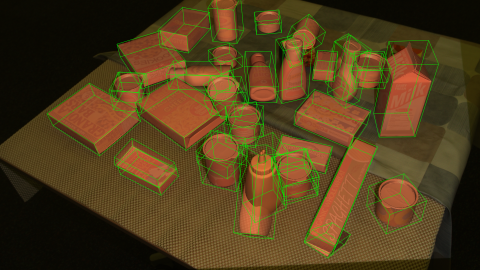} &
\includegraphics[width=0.196\linewidth]{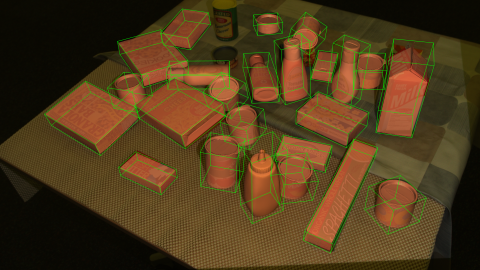} \\
Original & NovaSplat (Ours) & GFreeDet2~\cite{Wang_ICRA_2026} & OPFormer~\cite{Moroz_WACV_2026} & NeMO~\cite{Jung_3DV_2026} \\
\end{tabular}
\caption{Qualitative comparison on the \textbf{HOPEv2} dataset. NovaSplat robustly estimates poses in cluttered scenes with heavy occlusions, outperforming competing model-free methods. Bounding cuboids highlight the predicted 6D poses. Additional qualitative comparisons are provided in Sec.~\ref{sec:supp_qualitative}.}
\label{fig:hopev2_qualitative}
\end{figure*}

\subsection{Implementation Details}
\label{sec:implementation}

\PAR{Onboarding.} Onboarding videos contain $\sim$400 images per object, captured while moving the camera around the object, covering the full 360\textdegree{} azimuth and $-70$\textdegree{} to $+68$\textdegree{} elevation (upright and inverted captures). The median angular distance from a test view to its nearest onboarding view is 3.3\textdegree{}. We use the camera and object poses provided by the datasets to place all onboarding cameras into a common reference frame and triangulate masked object points into a sparse point cloud. Poses can equivalently be estimated via SfM: on both a textured HOPEv2 object and the most textureless, specular IndustryShapes object, all onboarding frames register into a single model, with median camera-pose errors of 1.1 mm and 9.5 mm, respectively. We train a 3DGS model with gsplat~\cite{Ye_JMLR_2025} and recover a mesh via OpenMVS~\cite{Cernea_CODE_2020}, rendering templates at $480 \times 480$ pixels. Video frames have a native resolution of $1920\times1440$ and are left unresized throughout SfM and 3DGS training.

\PAR{Inference.} We adapt FreeZe's coarse pose estimation module for our template-based setting and optimize the pose during refinement for at most $N_\text{ref}$ iterations with early stopping, as described in Sec.~\ref{sec:refinement}.

\PAR{Timing.} Onboarding is a one-time cost per object, completing in 7.8 minutes and producing the 3DGS representation, mesh, templates, and features required at inference time. Per-image inference totals 56.1\,s, dominated by coarse pose estimation (18.6\,s) and 3DGS-based refinement (35.4\,s). To reduce this cost, we additionally provide \textit{NovaSplat‑Lite}, an inference‑only variant that replaces the Open3D point‑cloud conversions in the coarse stage with native PyTorch operations and a single‑scale local reference frame and shortens the refinement budget with earlier stopping. Onboarding and the learned representation are unchanged. NovaSplat‑Lite runs at 18.0\,s per image, a $3.1\times$ speed‑up, at a cost of 0.001 AP on the HOPEv2 test set (0.529 vs. 0.528, Table 1). Unless stated otherwise, all reported results use the full configuration. All timings were measured on a workstation with an Intel(R) Core(TM) i7-14700K processor (28 cores) and a single NVIDIA RTX4090 GPU. 

\PAR{Memory.} Onboarding peaks at approximately 18\,GB CPU and 7.6\,GB GPU memory. Inference peaks at approximately 8.0\,GB CPU and 6\,GB GPU memory.

Sec.~\ref{sec:supp_implementation},~\ref{sec:runtime} provide (a) hyperparameters for 3DGS training and mesh recovery, (b) refiner settings and thresholds for both variants, (c) per-stage timing breakdown for both variants, (d) full NovaSplat-Lite details and (e) CPU/GPU memory profiles for onboarding and inference.

\subsection{Main Results}
\label{sec:results}

We evaluate NovaSplat against state-of-the-art baselines in both the model-based and model-free settings. For HOPEv2, baselines are selected from the BOP challenge leaderboard\footnote{\url{https://bop.felk.cvut.cz/leaderboards/modelfree-pose-detection-unseen-bop24/hopev2/}} among entries with an accompanying publication, ensuring a fair comparison under identical evaluation conditions since all results are obtained through the BOP evaluation server. For IndustryShapes, we compare against all methods reporting results on this dataset in the literature.

\begin{table*}[t]
\centering
\setlength{\tabcolsep}{4pt}
\begin{tabular}{@{}lc @{\hspace{10pt}} cccc c cccc@{}}
\toprule
& & \multicolumn{4}{c}{IndustryShapes Classic} & & \multicolumn{4}{c}{IndustryShapes Extended} \\
\cmidrule(lr){3-6} \cmidrule(lr){8-11}
Method & CAD & AR$_\text{MSPD}$ $\uparrow$ & AR$_\text{MSSD}$ $\uparrow$ & AR$_\text{VSD}$ $\uparrow$ & AR $\uparrow$ & & AR$_\text{MSPD}$ $\uparrow$ & AR$_\text{MSSD}$ $\uparrow$ & AR$_\text{VSD}$ $\uparrow$ & AR $\uparrow$ \\
\midrule
FreeZe~\cite{Caraffa_ECCV_2024}         & \cmark & 0.66 & 0.72 & \textbf{0.54} & 0.64 & & 0.75 & 0.67 & 0.42 & 0.61 \\
FoundPose~\cite{Oernek_ECCV_2024}       & \cmark & 0.31 & 0.39 & 0.22 & 0.30 & & 0.41 & 0.25 & 0.18 & 0.28 \\
FoundationPose~\cite{Wen_CVPR_2024}     & \cmark & \textbf{0.73} & \textbf{0.74} & 0.53 & \textbf{0.67} & & \textbf{0.83} & \textbf{0.74} & \textbf{0.50} & \textbf{0.69} \\
\midrule
OnePose~\cite{Sun_CVPR_2022} & \xmark &   0.08   &   0.08   &   0.06   & 0.07 & &   0.04   &  0.01  &   0.01   & 0.02 \\
OnePose++~\cite{He_NEURIPS_2022} & \xmark &  0.12    &   0.16   &  0.12 & 0.14 & &   0.26   &  0.19  &    0.14  & 0.20 \\
FoundationPose~\cite{Wen_CVPR_2024}     & \xmark & 0.26 & 0.31 & 0.18 & 0.25 & & 0.40 & 0.37 & 0.22 & 0.33 \\
\textbf{NovaSplat}                      & \xmark & \textbf{0.33} & \textbf{0.51} & \textbf{0.26} & \textbf{0.36} & & \textbf{0.48} & \textbf{0.47} & \textbf{0.25} & \textbf{0.40} \\
\bottomrule
\end{tabular}
\caption{Quantitative results on the \textbf{IndustryShapes} dataset, \textit{classic} and \textit{extended} splits. Methods are grouped by category, with model-based methods listed first and model-free methods listed below. Top-performing model-based and model-free scores are in \textbf{bold}.}
\label{tab:industryshapes}
\end{table*}

\textbf{HOPEv2 Results.}
Table~\ref{tab:hopev2_results} reports results on HOPEv2, a challenging benchmark featuring cluttered scenes with heavy occlusion and multiple instances of each object, all with annotated per-instance poses. A central contribution of NovaSplat is closing the gap between model-based and model-free pose estimation, demonstrated clearly here. NovaSplat surpasses every model-free competitor on all three metrics and, crucially, also outperforms all model-based methods on $\text{AP}$ and AP$_\text{MSSD}$ despite having no access to CAD models. Among model-free methods, NovaSplat outperforms the strongest competitor, GFreeDet2, by 17.0\% on AP (0.529 vs.\ 0.452) and by 39.5\% on AP$_\text{MSSD}$ (0.509 vs.\ 0.365) and leads OPFormer (0.509 vs.\ 0.243) and NeMO (0.509 vs.\ 0.291) by similarly large margins. Against model-based methods, NovaSplat achieves gains of 2.9\% on AP and 17.0\% on AP$_\text{MSSD}$ over Co-op (MUSE), the best CAD-model-assisted baseline, and outperforms GenFlow (AP$_\text{MSSD}$: 0.509 vs.\ 0.346) and GigaPose (0.509 vs.\ 0.046) by substantial margins. The sole exception is AP$_\text{MSPD}$, where Co-op (MUSE) leads by 8.2\% (0.593 vs.\ 0.548), while NovaSplat remains competitive with Co-op (CNOS) (0.549 vs.\ 0.548). Additional per-object quantitative breakdowns as well as qualitative results are provided in Sec.~\ref{sec:supp_per_object}.

The distinction between AP$_\text{MSSD}$ and AP$_\text{MSPD}$ is meaningful here: as noted by Hoda\v{n} et al.~\cite{Hodan_ECCVW_2020}, MSSD measures 3D surface distance and is most relevant for robotic manipulation, whereas MSPD measures 2D projection error and is more relevant for AR. Since our target application is manipulation, AP$_\text{MSSD}$ is our primary metric. We attribute this to NovaSplat's end-to-end geometric grounding: GeDi descriptors drive the coarse stage and $\mathcal{L}_\text{geo}$ constrains refinement, an integration absent in all competing model-free methods that rely on appearance cues, which explains the disproportionately large gains on AP$_\text{MSSD}$ relative to AP$_\text{MSPD}$.

Figure~\ref{fig:hopev2_qualitative} corroborates these findings: GFreeDet2 assigns multiple hypotheses per instance, OPFormer shows systematic scale errors, and NeMO misses many instances, whereas NovaSplat produces accurate and complete estimates whose residual errors trace to upstream detection rather than to pose estimation.

\textbf{IndustryShapes Results.} Table~\ref{tab:industryshapes} evaluates NovaSplat on a qualitatively different challenge. Unlike HOPEv2's textured household objects, IndustryShapes consists of texture-less, black industrial parts with specular surfaces, thin geometry, and small scale, conditions that stress-test both 3D reconstruction quality and appearance-based matching. Despite this domain shift, NovaSplat consistently surpasses all three model-free baselines, FoundationPose~\cite{Wen_CVPR_2024}, OnePose~\cite{Sun_CVPR_2022}, and OnePose++~\cite{He_NEURIPS_2022}, on both splits, demonstrating generalization across object domains. OnePose and OnePose++ rely on SfM-based 2D-3D correspondence matching (Sec.~\ref{sec:model-free-methods}) and are especially hindered by IndustryShapes' texture-less, specular surfaces, reaching only 0.07/0.02 and 0.14/0.20 AR on the \textit{classic}/\textit{extended} splits, respectively. NovaSplat improves over model-free FoundationPose by 44.0\% and 21.2\% on overall AR for the \textit{classic} and \textit{extended} splits, with the largest margin on AR$_\text{MSSD}$ (+64.5\% on \textit{classic}).

Model-based methods, however, retain a clear advantage here, with FoundationPose~\cite{Wen_CVPR_2024} achieving AR scores of 0.67 and 0.69 on the \textit{classic} and \textit{extended} splits, respectively, compared to NovaSplat's 0.36 and 0.40. That said, the gap narrows considerably on AR$_\text{MSSD}$: model-free FoundationPose trails the best model-based result by 0.43 on the \textit{classic} split, whereas NovaSplat reduces this gap to 0.23, nearly halving it. The same trend holds on the \textit{extended} split (gap reduced from 0.37 to 0.27). This is consistent with our findings on HOPEv2 and reinforces that NovaSplat's geometric grounding is the key driver of progress, as AR$_\text{MSSD}$ is the metric most sensitive to 3D accuracy.

The overall performance gap relative to model-based methods on this dataset is larger than on HOPEv2, and we attribute this to the fundamental difficulty of the objects' nature. Texture-less, uniformly black surfaces with specular highlights provide little signal for appearance-based feature matching and yield degraded 3D reconstructions that fail to capture fine geometric detail. This explanation is further supported by the qualitative results reported in the appendix. Improving NovaSplat's robustness to such objects, for instance, through geometry-only reconstruction pipelines, is a natural direction for future work.

\subsection{Ablation Study}
\label{sec:ablation}

\begin{table}[t]
\centering
\setlength{\tabcolsep}{4pt}
\begin{tabular}{@{}lcccc@{}}
\toprule
Configuration & AP  $\uparrow$ & AP$_\text{MSSD}$  $\uparrow$ & AP$_\text{MSPD}$  $\uparrow$ \\
\midrule
CPE                                            & 0.724          & 0.703          & 0.744 \\
+ ICP                                          & 0.471          & 0.384          & 0.557 \\
\midrule
CPE + 3DGS (only $\mathcal{L}_{\text{photo}}$) & 0.724          & 0.706          & 0.743 \\
CPE + 3DGS (only $\mathcal{L}_{\text{geo}}$)   & 0.666          & 0.651          & 0.682 \\ % without refinement is 0.708 0.690 0.726
CPE + 3DGS (full $\mathcal{L}$)                & \textbf{0.736} & \textbf{0.718} & \textbf{0.754} \\
\bottomrule
\end{tabular}
\caption{Ablation study on the \textbf{HOPEv2} validation set. We evaluate the effect of the refinement module and its loss components.}
\label{tab:ablation}
\end{table}

Table~\ref{tab:ablation} isolates the contribution of each component of our refinement module on the HOPEv2 validation set. The coarse pose estimation (CPE) stage alone achieves a strong baseline of 0.724 AP and 0.703 AP$_\text{MSSD}$. Refining this initialization with standard Iterative Closest Point (ICP), as employed by FreeZe~\cite{Caraffa_ECCV_2024}, severely degrades performance across all metrics: AP$_\text{MSSD}$ drops by 45.4\% (0.703 to 0.384) and AP by 34.9\% (0.724 to 0.471). This highlights the unreliability of zero-shot geometry extracted from monocular video for standard point-to-point registration: the resulting point clouds are too noisy for ICP, which is highly susceptible to local minima under such conditions.

Our 3DGS-based refinement takes a fundamentally different approach by directly optimizing the pose against the captured scene rather than relying on point cloud alignment. Using only the photometric loss $\mathcal{L}_\text{photo}$ yields a marginal improvement over CPE (AP$_\text{MSSD}$: 0.703 to 0.706), as texture ambiguities can still trap the optimizer in poor local minima. Isolating the geometric term, CPE + 3DGS (only $\mathcal{L}_{\text{geo}}$) degrades performance below CPE (AP$_\text{MSSD}$: 0.703 to 0.651; AP: 0.724 to 0.666), mirroring the ICP result: without the complementary photometric signal to anchor it, geometric alignment alone inherits the same sensitivity to noisy zero-shot geometry. Introducing the full dual loss, combining photometric and geometric (depth) terms, effectively constrains the optimization and outperforms both individual components, achieving the best performance across all metrics (AP$_\text{MSSD}$: 0.718, AP: 0.736). Figure~\ref{fig:refinement} illustrates the qualitative improvement.
%, where the Ketchup and Mustard objects show a clear correction of misaligned coarse estimates after refinement. 
Together, these results validate our design choice to use the 3DGS representation not only for template rendering but also as a differentiable backbone for pose refinement.

\PAR{Onboarding View Density.} Since pose accuracy is fundamentally bounded by the quality of the onboarding reconstruction, we additionally study how the number of onboarding views affects downstream accuracy by varying the fraction of video frames used for SfM and 3DGS training. As detailed in Sec.~\ref{sec:supp_onboarding_density}, accuracy is essentially unaffected down to 25\% of the original views, degrades moderately at 5\%, and collapses at the extreme 1\% ratio, where insufficient view coverage prevents SfM and 3DGS from recovering a faithful object reconstruction. 

\PAR{3DGS Training Budget.} We further study how the onboarding reconstruction budget affects downstream accuracy by varying the 3DGS training iterations. As detailed in Sec.~\ref{sec:3dgs-budget}, accuracy improves steadily up to 5k iterations, reaching a plateau beyond that. We retain 14k as the setting under which all results were produced. 5k is the recommended setting under the BOP onboarding budget.

\begin{figure}[t]
    \centering
    \begin{subfigure}[b]{0.49\linewidth}
        \centering
        \includegraphics[height=2.8cm]{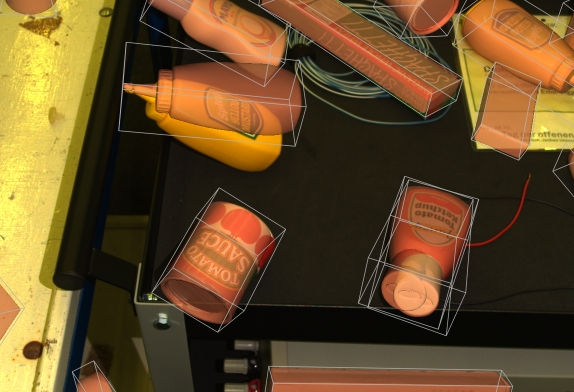}
        \caption{Before refinement}
        \label{fig:refinement_before}
    \end{subfigure}
    \hfill
    \begin{subfigure}[b]{0.49\linewidth}
        \centering
        \includegraphics[height=2.8cm]{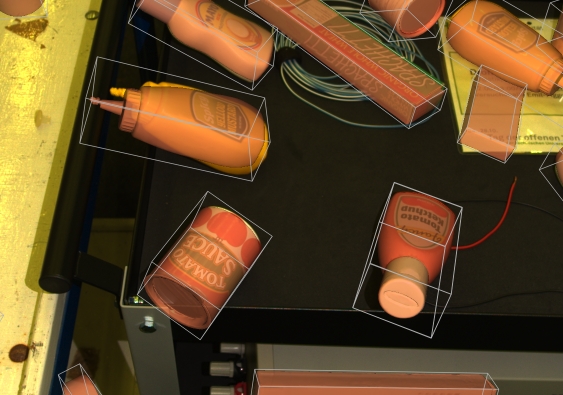}
        \caption{After refinement}
        \label{fig:refinement_after}
    \end{subfigure}
    \caption{Effect of pose refinement on \textbf{HOPEv2}. The improvement is most pronounced on the Ketchup and Mustard objects (center), where the initial coarse estimates~\ref{fig:refinement_before} are visibly misaligned, and refinement~\ref{fig:refinement_after} brings the projected bounding boxes into close agreement with the observed geometry.}
    \label{fig:refinement}
\end{figure}
}
{
    \section{Conclusions}
\label{sec:conclusions}

We presented NovaSplat, a model-free pipeline for 6D object pose estimation that replaces the CAD model requirement with a compact onboarding procedure: SfM and MVS recover object geometry from a short smartphone video, which is then used to train a 3D Gaussian Splatting representation for both template rendering and differentiable pose refinement. NovaSplat is, to our knowledge, the first model-free method to surpass model-based competitors on a BOP benchmark, exceeding the best model-based baseline on HOPEv2 by 2.9\% on AP and 17.0\% on AP$_\text{MSSD}$, and outperforms the best model-free competitor on IndustryShapes while narrowing the gap to model-based methods. We additionally provide NovaSplat‑Lite, an inference variant that is $3.1\times$ faster at a cost of 0.001 AP.

Our primary limitation is sensitivity to object appearance: on IndustryShapes, textureless, specular surfaces degrade both reconstruction quality and feature matching, leaving a gap to model-based methods. Future work includes improving robustness to such objects, exploring our refiner as a tracker, and replacing our separate MVS meshing step with direct mesh extraction from 3DGS, as in recent approaches like MeshSplatting~\cite{Held_CVPR_2026}.
}
{
    \section*{Acknowledgements}
\label{sec:acknowledgements}

This work was funded by the HEU programmes SOPRANO (GA No 101120990) and PANDORA (GA No 101135775).
}
{
    \small
    \bibliographystyle{ieeenat_fullname}
    \bibliography{main}
}
{
    \clearpage
    \appendix
    \section*{Appendix}
    \section{Full Implementation Details}
\label{sec:supp_implementation}

We specify two inference configurations. \textit{NovaSplat} is the full configuration and produces all results reported in the main paper unless stated otherwise. \textit{NovaSplat-Lite} is an inference-only variant that streamlines coarse pose estimation and photometric refinement for a $3.1\times$ speed-up (Sec.~\ref{sec:runtime}). The two share the same onboarding, the same learned object representation, and the same detection module and differ only in the settings listed in Table~\ref{tab:impl}.

\PAR{Onboarding.} We use the gsplat library~\cite{Ye_JMLR_2025} to train the splat for 14k iterations, allowing densification for the first 7k iterations and have the depth loss enabled. Our MVS pipeline is based on OpenMVS~\cite{Cernea_CODE_2020}. We render $N_{\text{templ}}=336$ templates at $480 \times 480$ pixels and sample $N_{\text{pts}}=\text{5k}$ points for $\mathcal{T}$. Onboarding frames are native $1920 \times 1440$ and are left unresized throughout SfM and 3DGS training. The $480 \times 480$ figure is the rendered template resolution only. Onboarding is identical for both configurations. Section~\ref{sec:3dgs-budget} studies the effect of the 14k budget.

\PAR{Coarse pose estimation.} We adapt FreeZe's coarse pose estimation module to our template-based setting, and omit its point filtering and PCA dimensionality reduction, as we found them to have no effect on system performance. NovaSplat computes the GeDi local reference frame at two scales. NovaSplat-Lite uses a single scale ($r_{\text{lrf}}=0.15$). NovaSplat-Lite additionally replaces the Open3D point-cloud conversions preceding descriptor extraction with native PyTorch operations (Sec.~\ref{sec:runtime-opt}). This is an implementation change only and leaves the descriptor formulation and its numeric parameters unchanged.

\PAR{Refinement.} In the refiner, the alpha channel is thresholded at $\tau_{\alpha}=0.001$, and the point cloud is filtered using a distance threshold of $d_{\text{range}}=30$\,mm. Pose optimization uses two interleaved AdamW~\cite{Loshchilov_ICLR_2019} optimizers with learning rates of $0.01$ for rotation and $5.0$ for translation. Rotation and translation updates alternate every $N_{\text{alt}}=10$ steps. Per-iteration corrections are clamped to $\alpha_{\text{thresh}}=10^{\circ}$ for rotation and $t_{\text{thresh}}=\pm10$\,mm for lateral translation. The geometric loss weight is set to $w_{\text{geo}}=0.001$, reflecting the scale difference between the SSIM-based $\mathcal{L}_{\text{photo}}$ and the Chamfer-based $\mathcal{L}_{\text{geo}}$. NovaSplat runs the optimizer for at most $N_{\text{ref}}=200$ iterations with early stopping after $N_{\text{halt}}=30$ iterations without improvement. NovaSplat-Lite reduces these to $N_{\text{ref}}=20$ and $N_{\text{halt}}=3$. All other refiner settings are shared.

\begin{table}[t]
\centering
\small
\setlength{\tabcolsep}{4pt}
\begin{tabular}{@{}llcc@{}}
\toprule
& Parameter & NovaSplat & -Lite \\
\midrule
\multirow{5}{*}{\rotatebox{90}{Onboard}}
 & 3DGS iterations       & \multicolumn{2}{c}{14k (densify $<$7k)} \\
 & Depth loss            & \multicolumn{2}{c}{enabled} \\
 & $N_{\text{templ}}$    & \multicolumn{2}{c}{336} \\
 & Template resolution   & \multicolumn{2}{c}{$480 \times 480$} \\
 & $N_{\text{pts}}$      & \multicolumn{2}{c}{5k} \\
\midrule
\multirow{3}{*}{\rotatebox{90}{Detect}}
 & Proposal model        & \multicolumn{2}{c}{YOLO-E} \\
 & Feature extractor     & \multicolumn{2}{c}{DINOv3 ViT-L/16} \\
 & Templates             & \multicolumn{2}{c}{ours (rendered)} \\
\midrule
\multirow{3}{*}{\rotatebox{90}{CPE}}
 & Geometric encoder     & \multicolumn{2}{c}{GeDi (frozen)} \\
 & LRF scales            & 2  & \textbf{1} \\
 & Point-cloud ops       & Open3D & \textbf{PyTorch} \\
\midrule
\multirow{7}{*}{\rotatebox{90}{Refinement}}
 & $N_{\text{ref}}$      & 200 & \textbf{20} \\
 & $N_{\text{halt}}$     & 30  & \textbf{3} \\
 & $N_{\text{alt}}$      & \multicolumn{2}{c}{10} \\
 & LR (rot./transl.)     & \multicolumn{2}{c}{0.01 / 5.0 (AdamW)} \\
 & $\alpha_{\text{thresh}}$ / $t_{\text{thresh}}$ & \multicolumn{2}{c}{$10^{\circ}$ / $\pm10$\,mm} \\
 & $\tau_{\alpha}$ / $d_{\text{range}}$ & \multicolumn{2}{c}{0.001 / 30\,mm} \\
 & $w_{\text{geo}}$ / MS-SSIM scales & \multicolumn{2}{c}{0.001 / 3} \\
\bottomrule
\end{tabular}
\caption{Implementation settings for both configurations. Entries spanning both columns are identical. Bold entries mark where NovaSplat-Lite differs.}
\label{tab:impl}
\end{table}

\section{Runtime Analysis}
\label{sec:runtime}

\subsection{Measurement Protocol}
All timings are measured on a workstation with an Intel(R) Core(TM) i7-14700K processor (28 cores) and a single NVIDIA RTX~4090 GPU. Inference timings are reported on the HOPEv2 validation set, which comprises $457$ images containing $9{,}873$ annotated object instances, an average of $21.6$ instances per image. Onboarding timings are per object and are a one-time cost. We report wall-clock time with all model weights already resident in GPU memory, excluding dataset I/O and one-time process startup.

Because HOPEv2 images are densely populated, the per-image figure aggregates $21.6$ instance-level pose estimates on average. We therefore report both per-image and per-instance timings, as the latter is the quantity that scales with scene complexity in a deployment setting.

\subsection{Per-Stage Breakdown}
Table~\ref{tab:runtime} reports the per-stage runtime of the full pipeline. Onboarding is unaffected by the inference-side optimizations described in Sec.~\ref{sec:runtime-opt} and is therefore reported once. For inference, we report both configurations: \textit{NovaSplat}, which produces all results in the main paper unless stated otherwise, and \textit{NovaSplat-Lite}, the streamlined variant.

\begin{table}[t]
\centering
\small
\begin{tabular}{lr}
\toprule
\multicolumn{2}{l}{\textbf{Onboarding} (per object, one-time)} \\
\midrule
SfM                            & 10.4\,s \\
MVS                            & 77.7\,s \\
3DGS training (14k iterations) & 326.7\,s \\
Poisson subsampling            & 11.5\,s \\
Template rendering             & 14.5\,s \\
Feature extraction             & 27.4\,s \\
\midrule
Total                          & 468.2\,s \\
\bottomrule
\end{tabular}

\vspace{1.0em}

\begin{tabular}{lrrr}
\toprule
\textbf{Inference} (per image) & NovaSplat & \textbf{-Lite} & Speed-up \\
\midrule
Detection \& segmentation & 2.1\,s  & 1.2\,s  & 1.8$\times$ \\
Coarse pose estimation    & 18.6\,s & 9.8\,s  & 1.9$\times$ \\
3DGS refinement           & 35.4\,s & 7.0\,s  & 5.1$\times$ \\
\midrule
Total                     & 56.1\,s & \textbf{18.0\,s} & \textbf{3.1$\times$} \\
Per object instance       & 2.60\,s & \textbf{0.83\,s} & \textbf{3.1$\times$} \\
\bottomrule
\end{tabular}
\caption{Per-stage runtime breakdown. Onboarding is a one-time per-object cost and is identical for both configurations. NovaSplat-Lite streamlines coarse pose estimation and 3DGS refinement (Sec.~\ref{sec:runtime-opt}) for a $3.1\times$ end-to-end speed-up. Per-instance figures divide the per-image total by the $21.6$ object instances per image on the HOPEv2 validation set. Detection and segmentation is identical in both configurations, and the two figures are drawn from separate profiling runs. The difference between them is not attributable to the optimizations described here, which affect only coarse pose estimation and refinement.}
\label{tab:runtime}
\end{table}

\subsection{Where Coarse Pose Estimation Spends Its Time}
Coarse pose estimation is the dominant inference stage in NovaSplat-Lite, accounting for $9.8$\,s of the $18.0$\,s per-image budget. Table~\ref{tab:cpe-substeps} decomposes it into its constituent operations.

The decomposition makes the cost structure unambiguous. GeDi descriptor extraction alone accounts for $91.0\%$ of coarse pose estimation and $49.3\%$ of total per-image inference time. The geometric verification machinery that follows it, similarity computation, correspondence matching, RANSAC, and hypothesis scoring, together consumes $0.45$\,s per image, or $4.6\%$ of the stage.

\subsection{Optimizations}
\label{sec:runtime-opt}
The optimizations behind NovaSplat-Lite target both dominant inference stages: coarse pose estimation and photometric refinement.

\begin{table}[t]
\centering
\small
\begin{tabular}{lrrr}
\toprule
CPE sub-step & Total & s/image & \% \\
\midrule
Point-cloud preparation      & 79.6\,s   & 0.174 & 1.8 \\
GeDi descriptor extraction   & 4058.3\,s & 8.880 & 91.0 \\
Visual--geometric fusion     & 114.9\,s  & 0.251 & 2.6 \\
Feature similarity           & 94.2\,s   & 0.206 & 2.1 \\
Correspondence matching      & 8.7\,s    & 0.019 & 0.2 \\
RANSAC                       & 19.7\,s   & 0.043 & 0.4 \\
Hypothesis scoring           & 82.7\,s   & 0.181 & 1.9 \\
\midrule
Total                        & 4458.1\,s & 9.758 & 100.0 \\
\bottomrule
\end{tabular}
\caption{Sub-step breakdown of coarse pose estimation for NovaSplat-Lite, aggregated over the $457$ images and $9{,}873$ object instances of the HOPEv2 validation set. Geometric descriptor extraction dominates, consuming $91\%$ of the CPE budget and $49\%$ of total inference time. Every remaining sub-step costs under $0.3$\,s per image.}
\label{tab:cpe-substeps}
\end{table}

\PAR{Coarse pose estimation.} Profiling revealed that descriptor extraction was dominated not by descriptor computation itself but by data marshalling. The full configuration converts point clouds to Open3D structures for filtering, voxel downsampling, and radius-neighbor queries, incurring two GPU$\rightarrow$CPU$\rightarrow$Open3D round-trips per object instance. NovaSplat-Lite replaces these conversions with native PyTorch operations that keep the point cloud resident on the GPU throughout: non-finite filtering, a scatter-mean voxel downsampling, and a vectorized radius-neighbor construction. It further reduces cost by computing the local reference frame at a single scale ($r_\mathrm{lrf} = 0.15$) rather than two. Together these changes reduce coarse pose estimation from $18.6$\,s to $9.8$\,s per image, a $1.9\times$ speed-up, without altering the descriptor formulation.

\PAR{Photometric refinement.} NovaSplat-Lite reduces the pose optimization budget from $N_\mathrm{ref} = 200$ to $N_\mathrm{ref} = 20$ maximum iterations and tightens the early-stopping patience from $N_\mathrm{halt} = 30$ to $N_\mathrm{halt} = 3$. The photometric objective converges quickly from a well-initialized coarse pose: the dominant pose correction is captured within the first few iterations, and the long tail of the full schedule contributes minor improvements. This reduces refinement from $35.4$\,s to $7.0$\,s per image, a $5.1\times$ speed-up, and is the single largest contributor to the end-to-end gain.

\PAR{Accuracy cost.} The combined effect is a $3.1\times$ end-to-end speed-up, from $56.1$\,s to $18.0$\,s per image ($2.60$\,s to $0.83$\,s per object instance), at a cost of $0.001$ AP on the HOPEv2 test set ($0.529$ vs.\ $0.528$, Table~1 of the main paper). We note that refinement is batchable across object instances, so the per-image figure is a conservative upper bound for the static-scene pick-and-place scenarios NovaSplat targets.

\subsection{Memory Profile}
\label{sec:memory}
We continuously monitored host (RAM) and GPU memory utilization across the full HOPEv2 validation processing timeline, for both the onboarding and the inference stage. Figure~\ref{fig:memory} shows the resulting traces. Because the 3DGS training budget is the dominant onboarding cost (Sec.~\ref{sec:3dgs-budget}), we profile both the $14$k-iteration default and the reduced $5$k-iteration setting.

\begin{figure}
\centering
\includegraphics[width=\linewidth]{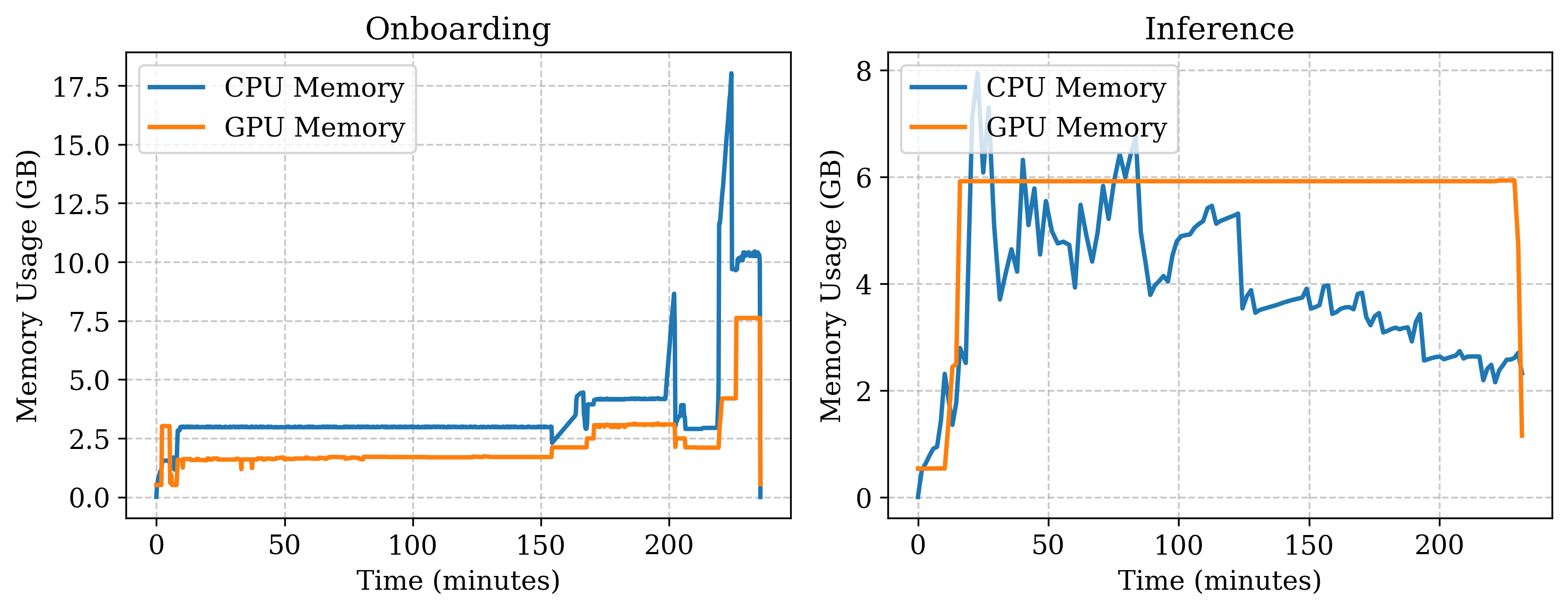}\\[0.6em]
\includegraphics[width=\linewidth]{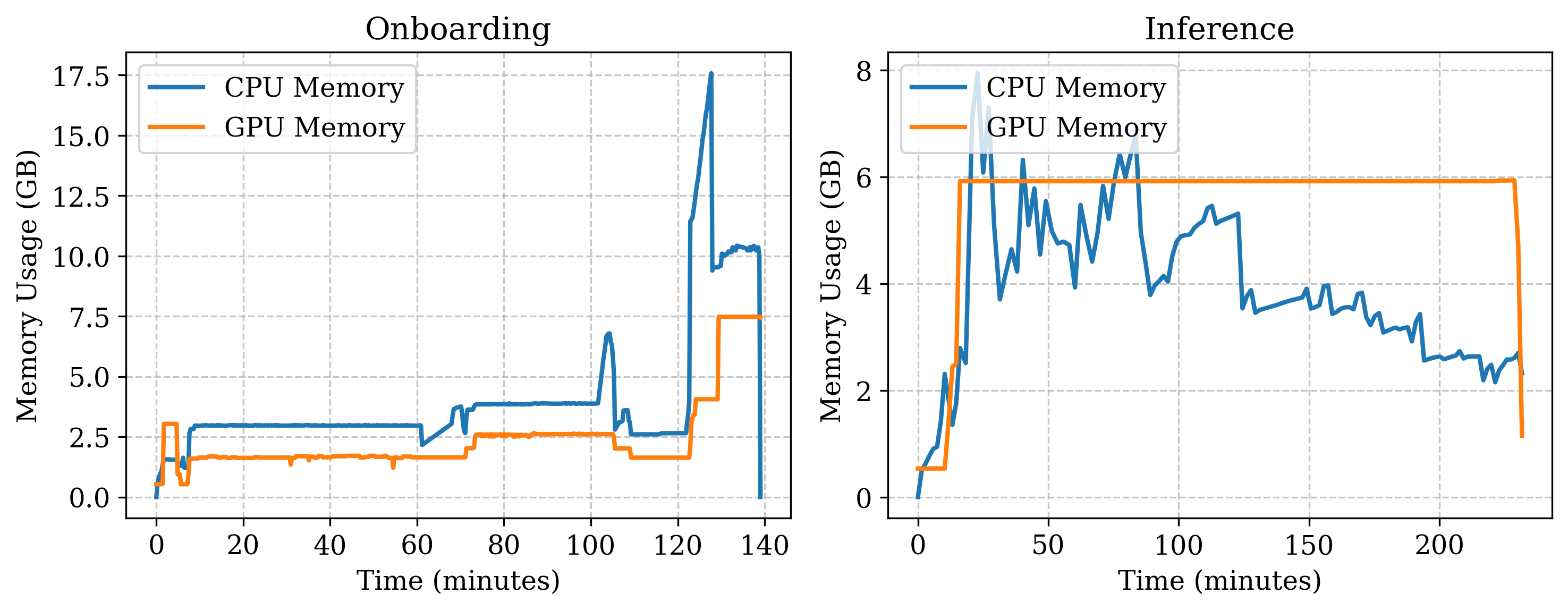}
\caption{Host and GPU memory utilization over the full HOPEv2 validation timeline. Top: $14$k 3DGS iterations (default). Bottom: $5$k iterations. Onboarding covers all $28$ HOPEv2 objects sequentially. Inference covers all $457$ validation images. Reducing the 3DGS budget shortens onboarding from $218$ to $140$ minutes while leaving the memory envelope essentially unchanged.}
\label{fig:memory}
\end{figure}

\PAR{Onboarding.} For most of its duration onboarding is inexpensive, holding a sustained working set of $3.0$--$4.2$\,GB host and $1.6$--$3.1$\,GB device memory. The $18.0$\,GB host peak is a narrow transient occurring in the final minutes of each object's onboarding, during geometry extraction and feature export, and is not representative of the stage as a whole. Device memory rises in discrete steps that track the pipeline's stage transitions, reaching $7.6$\,GB at the end.

\PAR{Inference.} Device memory rises to $6.0$\,GB within the first minutes and holds a flat plateau there for the entire run, reflecting the fixed cost of the resident detection, descriptor, and 3DGS models. Host memory shows a transient $8.0$\,GB peak shortly after start-up, then a working set that oscillates between $3.5$ and $6.8$\,GB before decaying to $2.1$--$3.0$\,GB over the second half of the run.

\PAR{Effect of the 3DGS budget.} Memory consumption is essentially invariant to the 3DGS training budget: reducing from $14$k to $5$k iterations changes the onboarding peak by $0.4$\,GB host and $0.1$\,GB device, while shortening the onboarding pass over all $28$ objects from $218$ to $140$ minutes. The reduced budget studied in Sec.~\ref{sec:3dgs-budget} therefore costs nothing in memory terms.

\medskip
\noindent
Taken together, a sustained device footprint of $6.0$\,GB places NovaSplat within reach of 8\,GB-class consumer GPUs, and the host requirement is modest except for a brief onboarding transient. The method can be deployed on standard consumer-grade hardware without out-of-memory issues.

\section{Onboarding Reconstruction Budget}
\label{sec:3dgs-budget}
Sec.~\ref{sec:runtime} shows that onboarding costs $468.2$\,s per object, of which 3DGS training accounts for $326.7$\,s, or $70\%$. We therefore study how the 3DGS training budget trades against downstream pose accuracy, re-running onboarding at iteration counts from $500$ to our $14$k default and re-evaluating the full pipeline on the HOPEv2 validation set. Table~\ref{tab:3dgs-budget} reports the result.

\begin{table}[t]
\centering
\small
\begin{tabular}{lrrrr}
\toprule
Iterations & Time & AP $\uparrow$ & $\mathrm{AP}_\mathrm{MSSD}$ $\uparrow$ & $\mathrm{AP}_\mathrm{MSPD}$ $\uparrow$ \\
\midrule
500  & 153\,s & 0.655 & 0.632 & 0.678 \\
1k   & 165\,s & 0.706 & 0.684 & 0.728\\
5k   & 258\,s & \textbf{0.741} & \textbf{0.722} & \textbf{0.761} \\
14k  & 468\,s & 0.736 & 0.718 & 0.754 \\
\bottomrule
\end{tabular}
\caption{Effect of the 3DGS training budget on downstream pose accuracy (HOPEv2 validation set). \textquote{Time} is the full per-object onboarding cost.}
\label{tab:3dgs-budget}
\end{table}

Accuracy improves steadily up to $5$k iterations and then plateaus. Relative to the $14$k default, the $5$k setting is marginally \emph{better} on all three metrics (AP $0.741$ vs.\ $0.736$; $\mathrm{AP}_\mathrm{MSSD}$ $0.722$ vs.\ $0.718$; $\mathrm{AP}_\mathrm{MSPD}$ $0.761$ vs.\ $0.754$), differences of under one percent that we regard as within run-to-run variation rather than a genuine gain from training less. Below $5$k the picture changes sharply: $1$k loses $4.1\%$ AP relative to the default and $500$ iterations loses $11.0\%$, as the reconstruction is no longer faithful enough to support template matching and photometric refinement.

The practical consequence is that the onboarding budget can be cut substantially at no measurable accuracy cost. At $5$k iterations the estimated per-object onboarding time falls to $258$\,s, bringing it within the BOP protocol's five-minute per-object budget, against $468.2$\,s at the $14$k default.

\section{Effect of Onboarding View Density}
\label{sec:supp_onboarding_density}

\begin{table}[t]
\centering
\begin{tabular}{@{}lccc@{}}
\toprule
Ratio & AP $\uparrow$ & AP$_\text{MSSD}$ $\uparrow$ & AP$_\text{MSPD}$ $\uparrow$ \\
\midrule
100\% (baseline) & 0.736          & 0.718          & 0.754 \\
25\%             & \textbf{0.738} & \textbf{0.722} & \textbf{0.754} \\
5\%              & 0.655          & 0.635          & 0.674 \\
1\%              & 0.212          & 0.203          & 0.221 \\
\bottomrule
\end{tabular}
\caption{Effect of onboarding view sampling ratio on pose accuracy, evaluated on the HOPEv2 validation set. Accuracy is stable down to 25\% of the original views, degrades moderately at 5\%, and collapses at 1\%.}
\label{tab:supp_onboarding_density}
\end{table}

To assess how the number of onboarding views affects downstream pose accuracy, we re-run onboarding while randomly subsampling the input video frames prior to SfM and 3DGS training, testing sampling ratios of 100\% (baseline), 25\%, 5\%, and 1\% of the original frame count, and re-evaluate the full NovaSplat pipeline on the HOPEv2 validation set. Table~\ref{tab:supp_onboarding_density} reports the resulting accuracy, and Figure~\ref{fig:ablation_templates} shows the corresponding rendered templates.

\begin{figure*}[t]
\centering
\setlength{\tabcolsep}{1pt}
\begin{tabular}{c *{10}{c}}

\rotatebox{90}{100\%} &
\includegraphics[width=0.09\linewidth, trim={100 100 100 100}, clip]{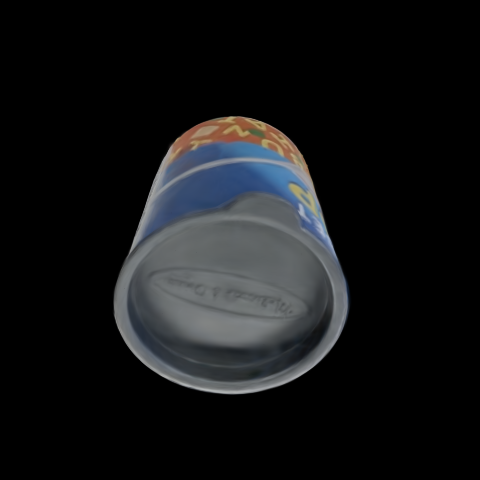} &
\includegraphics[width=0.09\linewidth, trim={100 100 100 100}, clip]{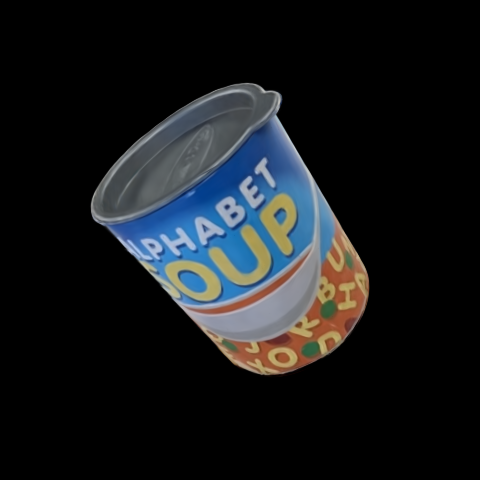} &
\includegraphics[width=0.09\linewidth, trim={100 100 100 100}, clip]{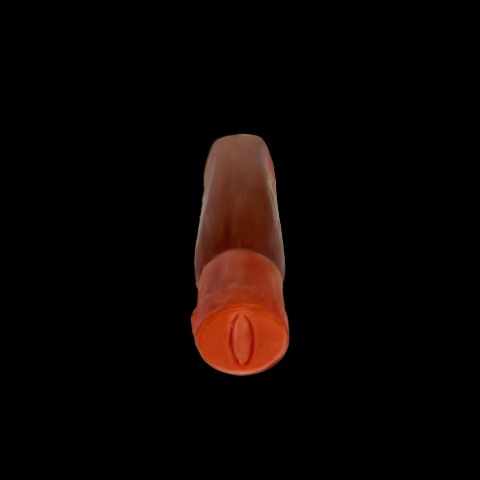} &
\includegraphics[width=0.09\linewidth, trim={100 100 100 100}, clip]{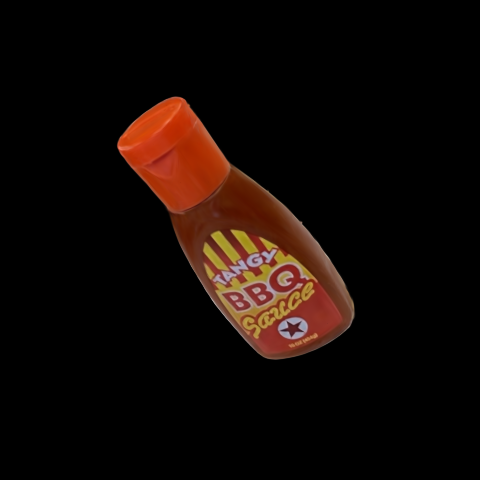} &
\includegraphics[width=0.09\linewidth, trim={100 100 100 100}, clip]{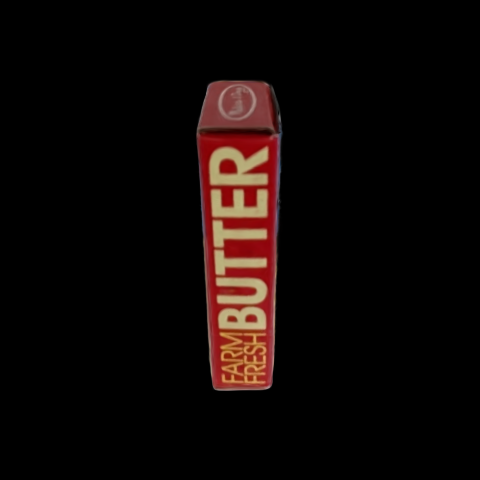} &
\includegraphics[width=0.09\linewidth, trim={100 100 100 100}, clip]{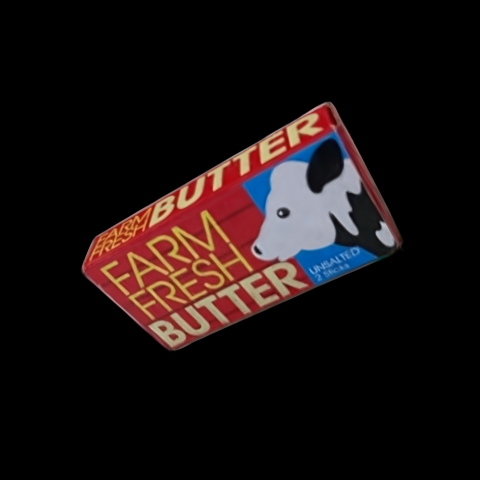} &
\includegraphics[width=0.09\linewidth, trim={100 100 100 100}, clip]{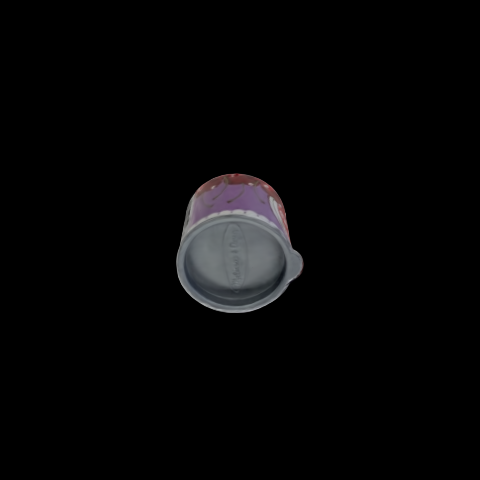} &
\includegraphics[width=0.09\linewidth, trim={100 100 100 100}, clip]{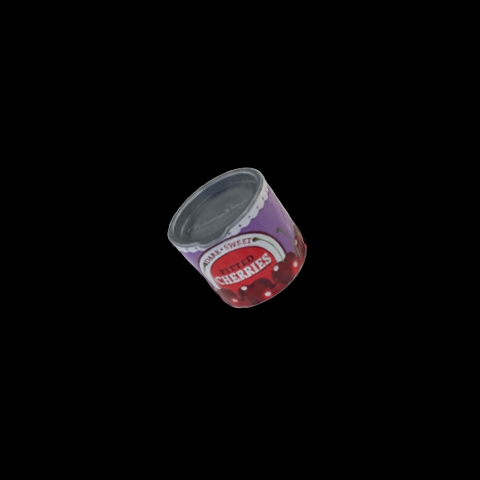} &
\includegraphics[width=0.09\linewidth, trim={100 100 100 100}, clip]{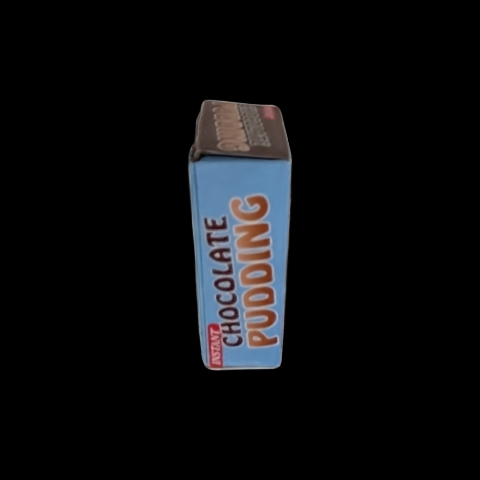} &
\includegraphics[width=0.09\linewidth, trim={100 100 100 100}, clip]{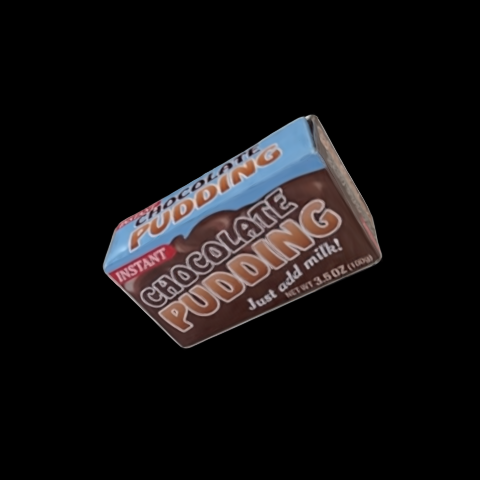} \\

\rotatebox{90}{25\%} &
\includegraphics[width=0.09\linewidth, trim={100 100 100 100}, clip]{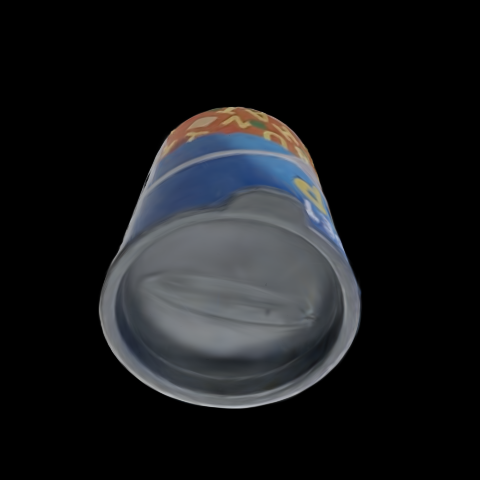} &
\includegraphics[width=0.09\linewidth, trim={100 100 100 100}, clip]{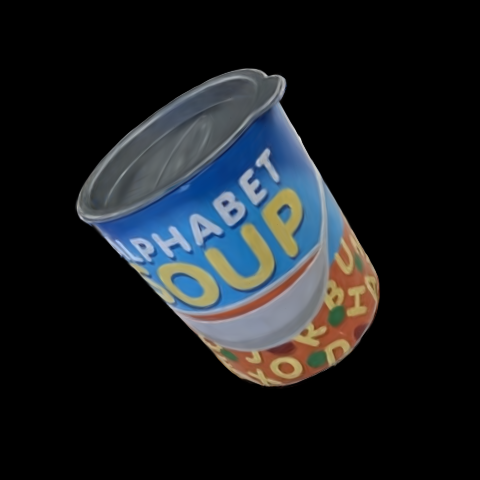} &
\includegraphics[width=0.09\linewidth, trim={100 100 100 100}, clip]{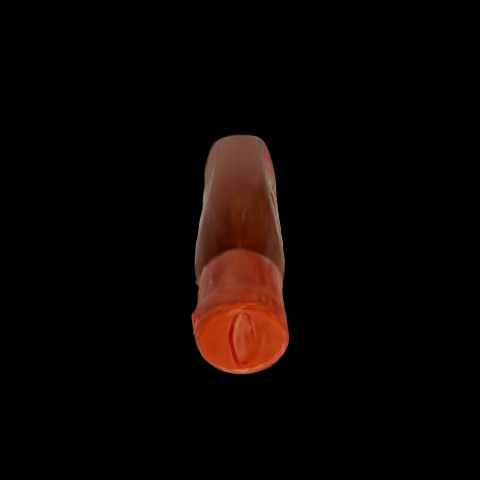} &
\includegraphics[width=0.09\linewidth, trim={100 100 100 100}, clip]{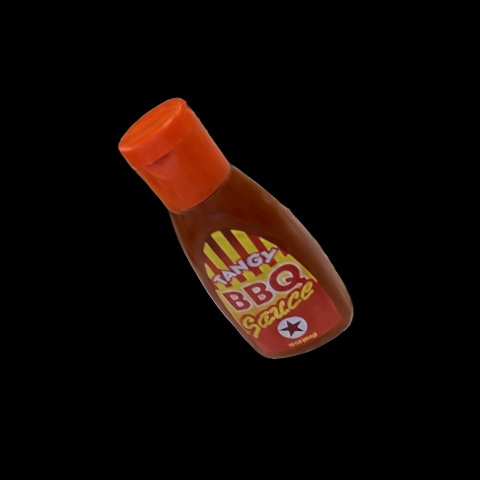} &
\includegraphics[width=0.09\linewidth, trim={100 100 100 100}, clip]{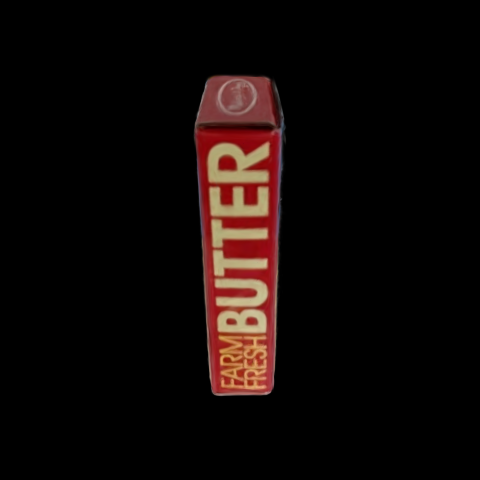} &
\includegraphics[width=0.09\linewidth, trim={100 100 100 100}, clip]{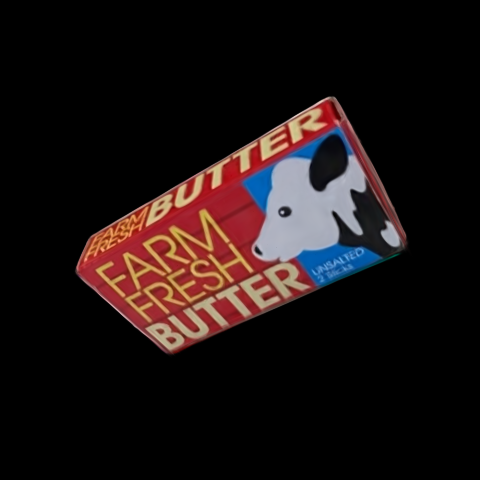} &
\includegraphics[width=0.09\linewidth, trim={100 100 100 100}, clip]{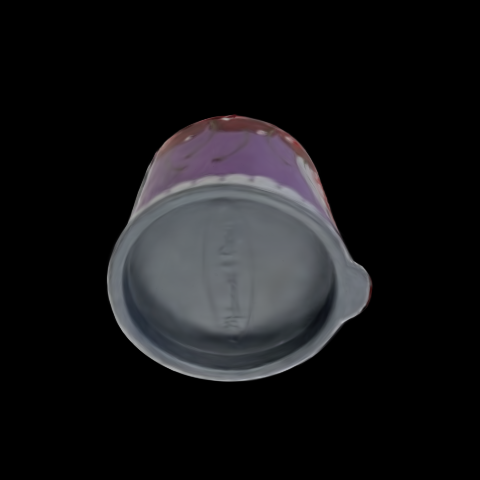} &
\includegraphics[width=0.09\linewidth, trim={100 100 100 100}, clip]{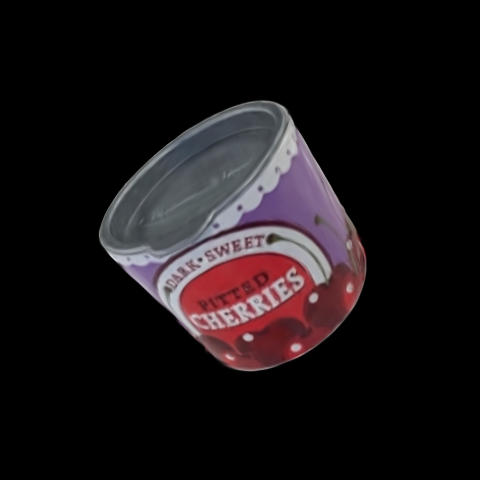} &
\includegraphics[width=0.09\linewidth, trim={100 100 100 100}, clip]{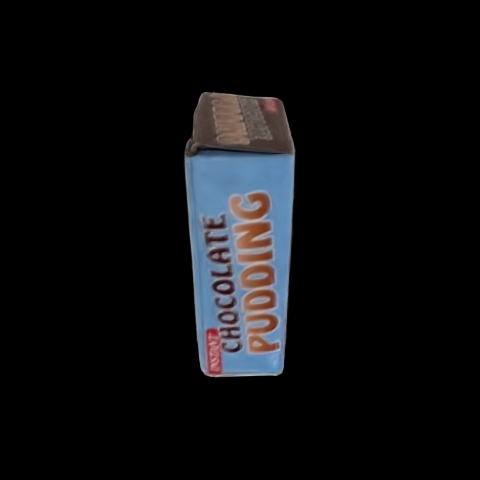} &
\includegraphics[width=0.09\linewidth, trim={100 100 100 100}, clip]{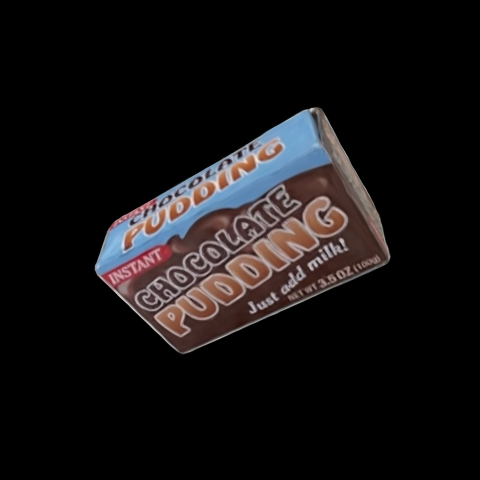} \\

\rotatebox{90}{5\%} &
\includegraphics[width=0.09\linewidth, trim={100 100 100 100}, clip]{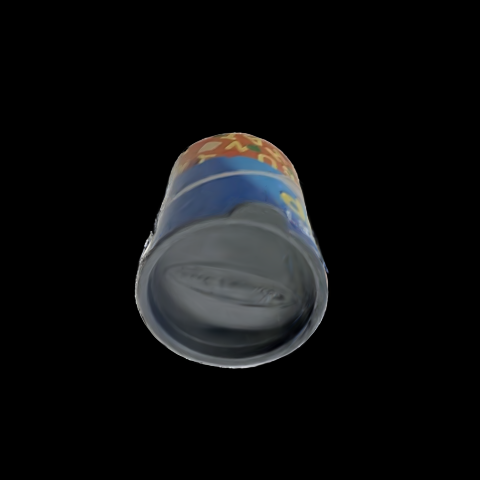} &
\includegraphics[width=0.09\linewidth, trim={100 100 100 100}, clip]{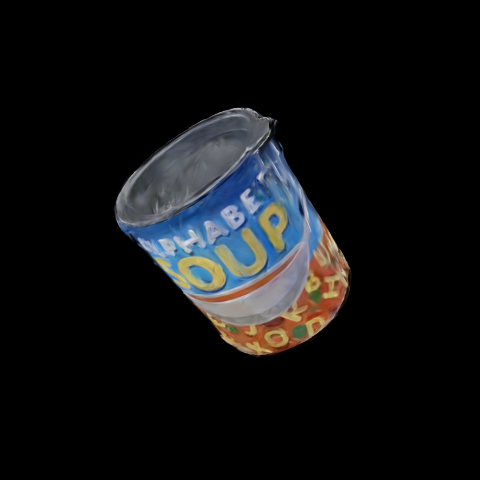} &
\includegraphics[width=0.09\linewidth, trim={100 100 100 100}, clip]{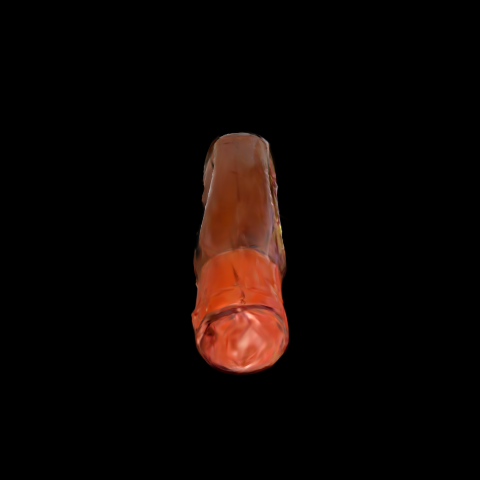} &
\includegraphics[width=0.09\linewidth, trim={100 100 100 100}, clip]{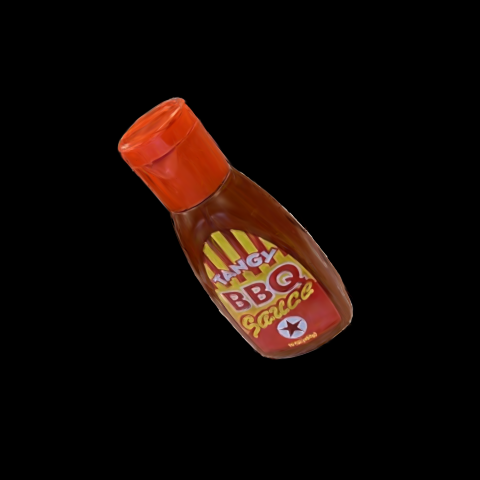} &
\includegraphics[width=0.09\linewidth, trim={100 100 100 100}, clip]{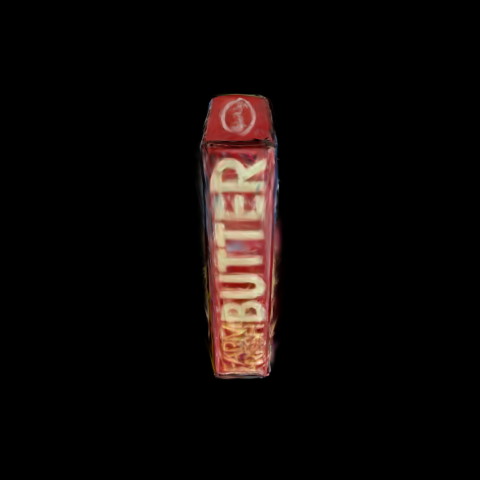} &
\includegraphics[width=0.09\linewidth, trim={100 100 100 100}, clip]{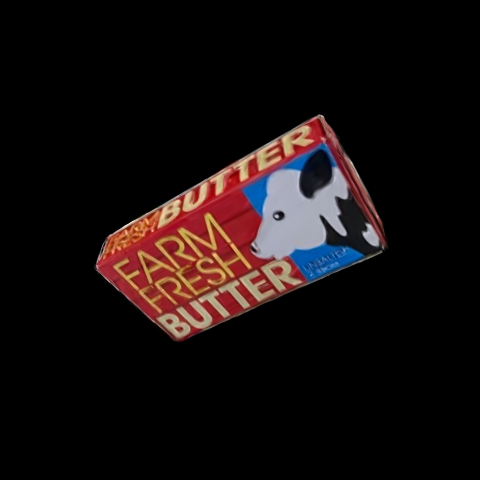} &
\includegraphics[width=0.09\linewidth, trim={100 100 100 100}, clip]{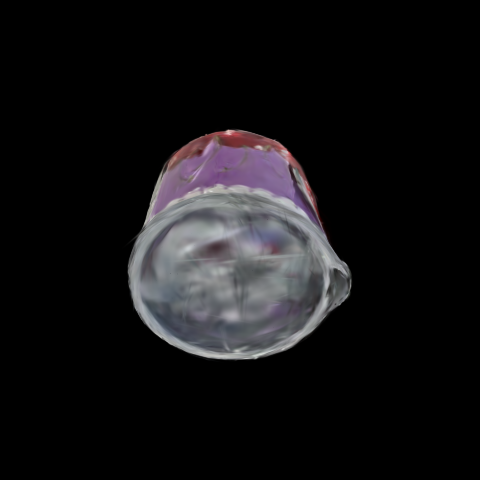} &
\includegraphics[width=0.09\linewidth, trim={100 100 100 100}, clip]{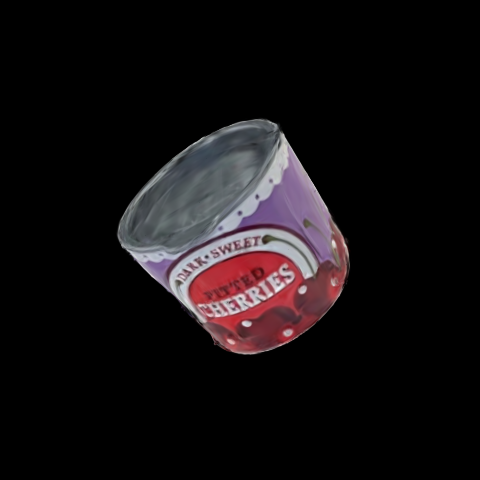} &
\includegraphics[width=0.09\linewidth, trim={100 100 100 100}, clip]{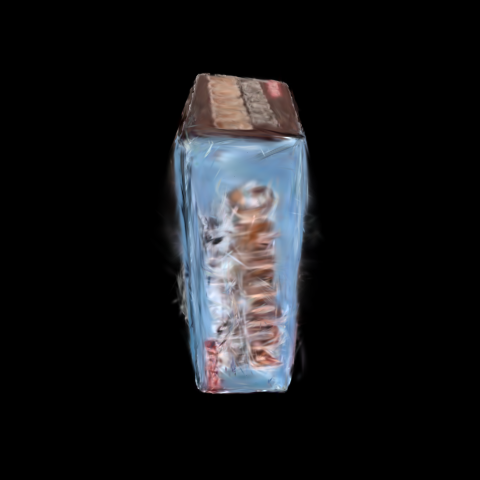} &
\includegraphics[width=0.09\linewidth, trim={100 100 100 100}, clip]{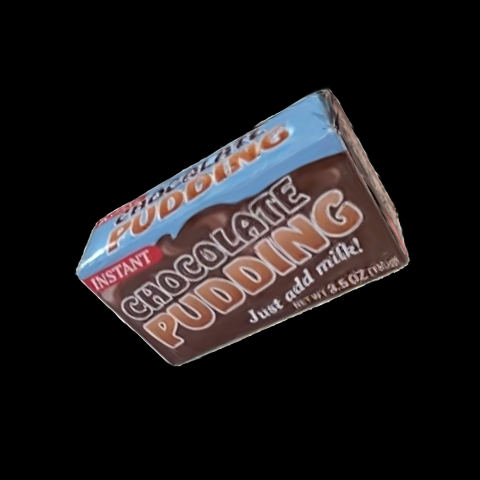} \\

\rotatebox{90}{1\%} &
\includegraphics[width=0.09\linewidth, trim={100 100 100 100}, clip]{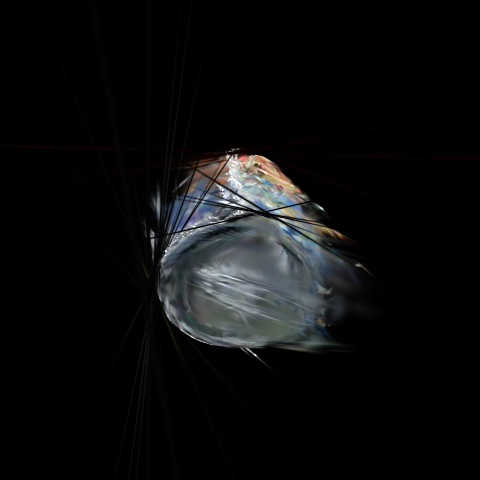} &
\includegraphics[width=0.09\linewidth, trim={100 100 100 100}, clip]{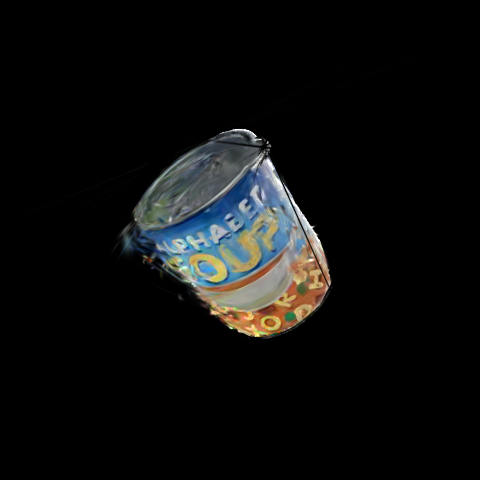} &
\includegraphics[width=0.09\linewidth, trim={100 100 100 100}, clip]{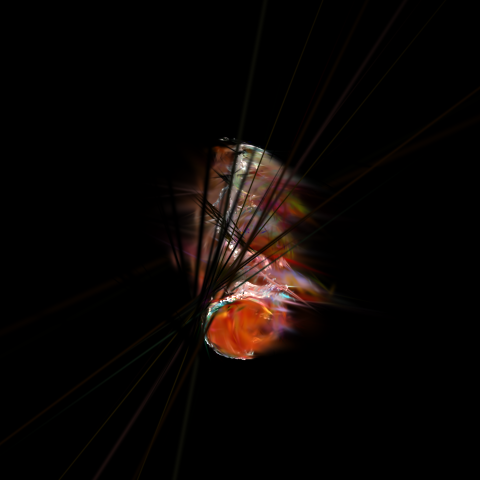} &
\includegraphics[width=0.09\linewidth, trim={100 100 100 100}, clip]{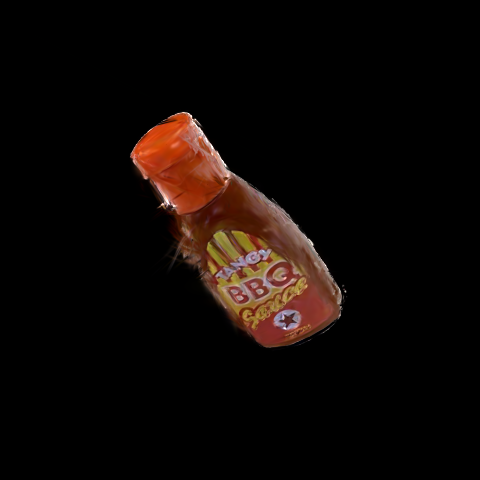} &
\includegraphics[width=0.09\linewidth, trim={100 100 100 100}, clip]{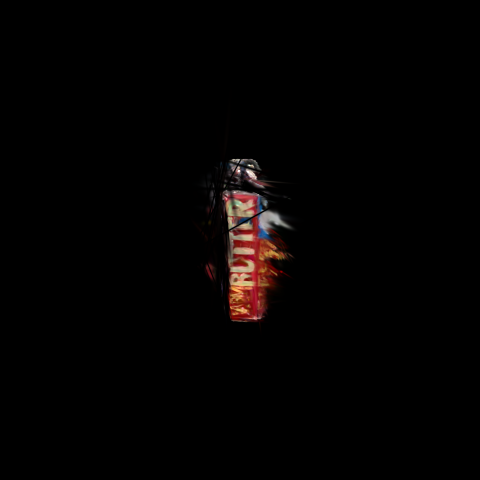} &
\includegraphics[width=0.09\linewidth, trim={100 100 100 100}, clip]{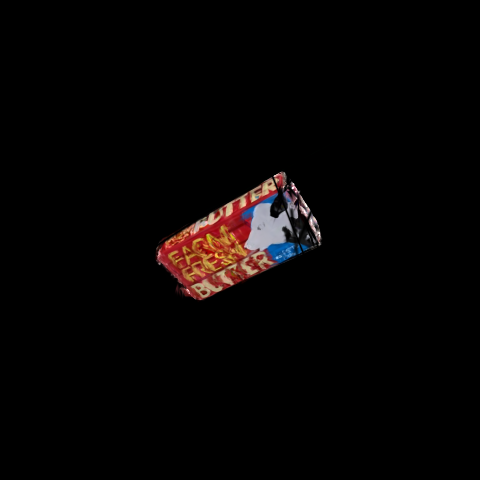} &
\includegraphics[width=0.09\linewidth, trim={100 100 100 100}, clip]{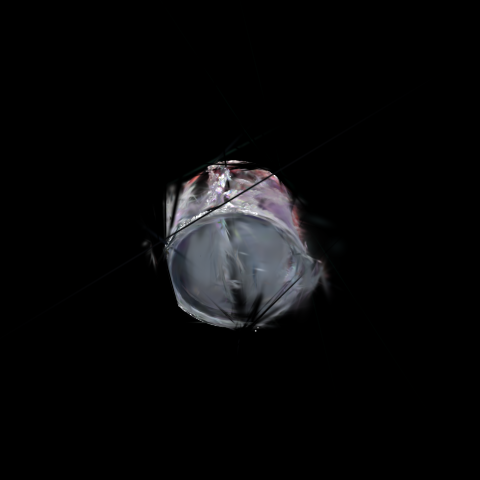} &
\includegraphics[width=0.09\linewidth, trim={100 100 100 100}, clip]{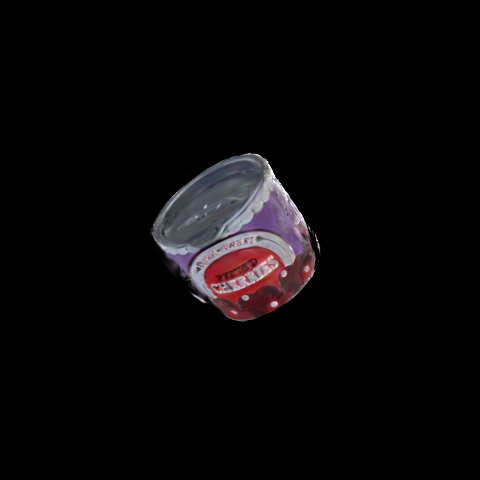} &
\includegraphics[width=0.09\linewidth, trim={100 100 100 100}, clip]{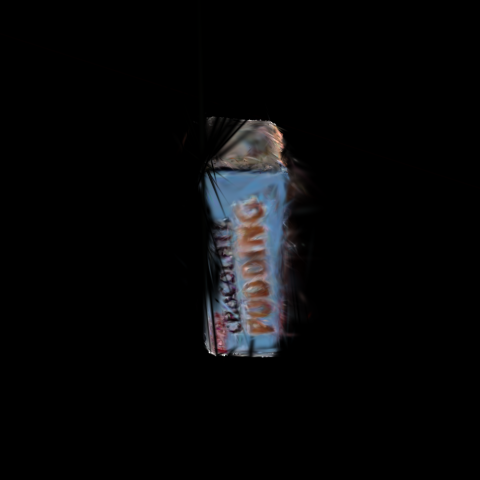} &
\includegraphics[width=0.09\linewidth, trim={100 100 100 100}, clip]{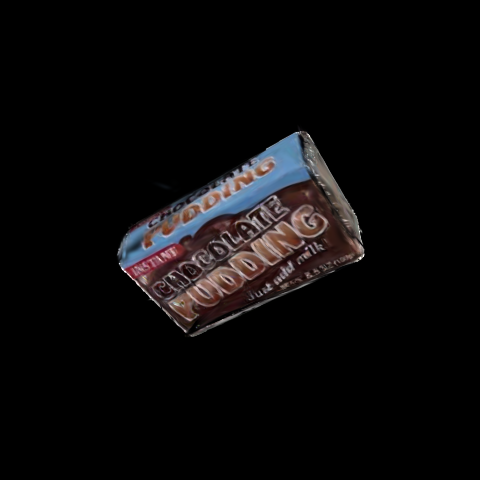} \\

\end{tabular}
\caption{Qualitative comparison of rendered templates under different onboarding sampling ratios on the HOPEv2 dataset. The rendering fidelity degrades significantly at extreme sampling ratios (1\%), resulting in artifacts and loss of detail, while 25\% maintains comparable quality to the baseline.}
\label{fig:ablation_templates}
\end{figure*}

At 25\% of the original views, accuracy matches or marginally exceeds the 100\% baseline (AP: 0.738 vs.\ 0.736; AP$_\text{MSSD}$: 0.722 vs.\ 0.718), indicating that our onboarding videos, captured densely for convenience, contain substantial redundancy and that a much sparser capture would suffice without harming reconstruction quality. Reducing further to 5\% causes a moderate drop (AP: 0.655, a 11.0\% relative decrease), as the sparser point cloud and viewpoint coverage begin to degrade the 3DGS reconstruction. At the extreme 1\% ratio, accuracy collapses (AP: 0.212), consistent with the qualitative degradation visible in Figure~\ref{fig:ablation_templates}: the rendered templates exhibit severe artifacts and loss of detail, reflecting insufficient multi-view coverage for both SfM triangulation and 3DGS optimization to converge to a faithful reconstruction. These results confirm that final pose accuracy is directly tied to onboarding reconstruction quality, which in turn depends on having sufficient view coverage, while also showing that NovaSplat is robust to substantial frame subsampling before quality begins to degrade.

\section{Onboarding Poses from Structure-from-Motion}
\label{sec:supp_sfm}
The onboarding protocol of BOP-H3 supplies camera poses together with the onboarding videos, and all results in the main paper are produced with those poses. This raises a fair question about how much of our performance depends on having them. This section shows that the poses can be recovered from the onboarding video alone and that this remains true for the textureless, specular objects the method targets, the case in which SfM is most often expected to fail.

\subsection{Protocol}
\label{sec:supp_sfm_protocol}
We discard the provided poses entirely and re-estimate the onboarding trajectory from the RGB frames with COLMAP features and matches~\cite{Schoenberger_CVPR_2016} followed by the GLOMAP global mapper~\cite{Pan_ECCV_2024}. Every onboarding frame is used (no temporal subsampling), images are \emph{not} masked to the object, loop detection is disabled, and the camera model is PINHOLE with the intrinsics shipped with each dataset, held fixed (no intrinsic refinement). 

We run this on two objects chosen to bracket the difficulty range of the benchmark: \texttt{obj\_000001} of HOPEv2, a textured grocery item, and \texttt{obj\_000005} of IndustryShapes, the most textureless and specular object in that dataset.

The estimated trajectory lives in an arbitrary gauge (the global mapper fixes the world frame and the overall scale), so it cannot be compared to the reference poses directly. We recover the similarity transform between the two sets of camera centers in closed form with the Umeyama algorithm, apply it to the estimated poses, and report the following:
\begin{itemize}
  \setlength{\itemsep}{0pt}
  \item \textbf{Absolute rotation error}: the geodesic angle between the reference camera orientation and the aligned estimate.
  \item \textbf{Absolute position error}: the Euclidean distance between the reference and aligned camera centers, in millimeters.
  \item \textbf{Relative rotation error}: the same angle computed on the relative pose between consecutive frames. This quantity is invariant to the alignment and measures the local multi-view consistency on which the MVS and 3DGS stages actually depend.
\end{itemize}

\subsection{Reconstruction}
\label{sec:supp_sfm_reconstruction}
\begin{table}[t]
\centering
\small
\setlength{\tabcolsep}{4pt}
\begin{tabular}{lcc}
\toprule
 & HOPEv2 & IndustryShapes \\
 & \texttt{obj\_000001} & \texttt{obj\_000005} \\
\midrule
Onboarding frames              & 178      & 522 \\
Frames registered              & 178 (100\%) & 522 (100\%) \\
Reconstructed sub-models       & 1        & 1 \\
Mean reprojection error (px)   & 0.50     & 0.67 \\
\midrule
Sparse 3D points               & 78{,}973 & 5{,}802 \\
\quad per registered image     & 444      & 11 \\
\quad inside object bbox       & ---      & 1{,}117 (19.3\%) \\
\quad within bbox\,$+25$\,mm   & ---      & 1{,}481 (25.5\%) \\
\midrule
Feature extraction (s)         & 10.6     & 15.1 \\
Matching (s)                   & 85.6     & 7.5 \\
Global mapping (s)             & 84.7     & 29.3 \\
\textbf{Total (s)}             & \textbf{181.0} & \textbf{51.9} \\
\bottomrule
\end{tabular}
\caption{SfM reconstruction of the onboarding sequences, with no ground-truth poses used. Both objects register fully into a single model. The textureless object yields a sparse cloud $40 \times$ of magnitude thinner, $80\%$ of which lie off the object.}
\label{tab:sfm_reconstruction}
\end{table}

Table~\ref{tab:sfm_reconstruction} reports the reconstruction itself, before any comparison to ground truth. Both sequences register \emph{completely}: every selected frame is placed into a \emph{single} model, with no sub-model fragmentation and sub-pixel mean reprojection error. Registration is therefore not the failure mode on textureless objects.

What changes dramatically is the density of the sparse point cloud. The IndustryShapes reconstruction recovers $5{,}802$ points from $522$ images ($11$ points per image) against $78{,}973$ points from $178$ images ($444$ points per image) on HOPEv2, a fortyfold drop in points yield per frame, which is precisely the signature of a surface that offers no repeatable features.

To locate those points, we transform the sparse cloud into the object frame with the similarity transform of Sec.~\ref{sec:supp_sfm_protocol} and test each point against the object's bounding box. Only $1{,}117$ of the $5{,}802$ points ($19.3\%$) fall inside it, and $1{,}481$ ($25.5\%$) fall within a $25$\,mm dilation of it. $80\%$ of the reconstruction is the surrounding scene.

\begin{table}[t]
\centering
\small
\setlength{\tabcolsep}{4pt}
\begin{tabular}{lcc}
\toprule
 & HOPEv2 & IndustryShapes \\
 & \texttt{obj\_000001} & \texttt{obj\_000005} \\
\midrule
Frames compared & 178 & 482 \\
Mean viewing distance (mm) & --- & 447 \\
\midrule
\multicolumn{3}{l}{\emph{Absolute rotation error} ($^\circ$)} \\
\quad median  & 0.32 & 1.36 \\
\quad mean    & 0.38 & 1.37 \\
\quad p95     & 0.89 & 2.23 \\
\quad max     & 1.25 & 3.77 \\
\midrule
\multicolumn{3}{l}{\emph{Absolute position error} (mm)} \\
\quad median  & 1.09 & 9.47 \\
\quad mean    & 1.37 & 9.14 \\
\quad p95     & 3.22 & 14.70 \\
\quad max     & 4.66 & 22.88 \\
\midrule
\multicolumn{3}{l}{\emph{Relative rotation error, consecutive frames} ($^\circ$)} \\
\quad median  & 0.07 & 0.19 \\
\quad p95     & 0.31 & 1.06 \\
\midrule
\multicolumn{3}{l}{\emph{Relative direction error, consecutive frames} ($^\circ$)} \\
\quad median  & 0.61 & 13.95 \\
\bottomrule
\end{tabular}
\caption{Estimated versus provided onboarding poses, after similarity alignment of the camera centers. Absolute errors are alignment-dependent. The relative rotation error is not and measures the local consistency that the MVS and 3DGS stages rely on.}
\label{tab:sfm_accuracy}
\end{table}

This is the mechanism that makes the onboarding stage robust here, and it is worth stating explicitly: \emph{the camera trajectory is not recovered from the object.} Onboarding sequences image the object together with its supporting surface, the calibration target, and the surrounding workspace, and those features (not the object's own appearance) constrain the poses. The object's geometry is then recovered by the MVS and 3DGS stages, which are driven by the estimated cameras rather than by sparse keypoints on the object. A textureless object, consequently, yields a thin, sparse cloud without yielding a poor camera trajectory.

\subsection{Pose Accuracy}
\label{sec:supp_sfm_accuracy}

\begin{figure}
\centering
\begin{subfigure}{\linewidth}
  \centering
  \includegraphics[width=\textwidth]{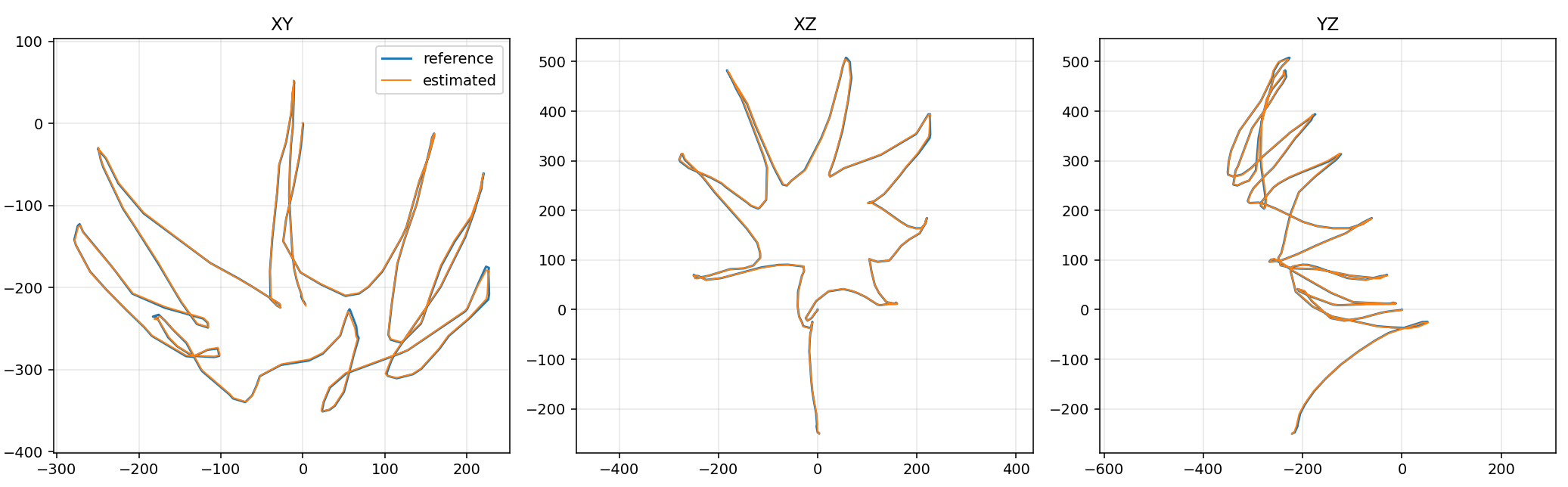}
  \caption{HOPEv2 \texttt{obj\_000001} (textured).}
  \label{fig:sfm_traj_hopev2}
\end{subfigure}\\[4pt]
\begin{subfigure}{\linewidth}
  \centering
  \includegraphics[width=\textwidth]{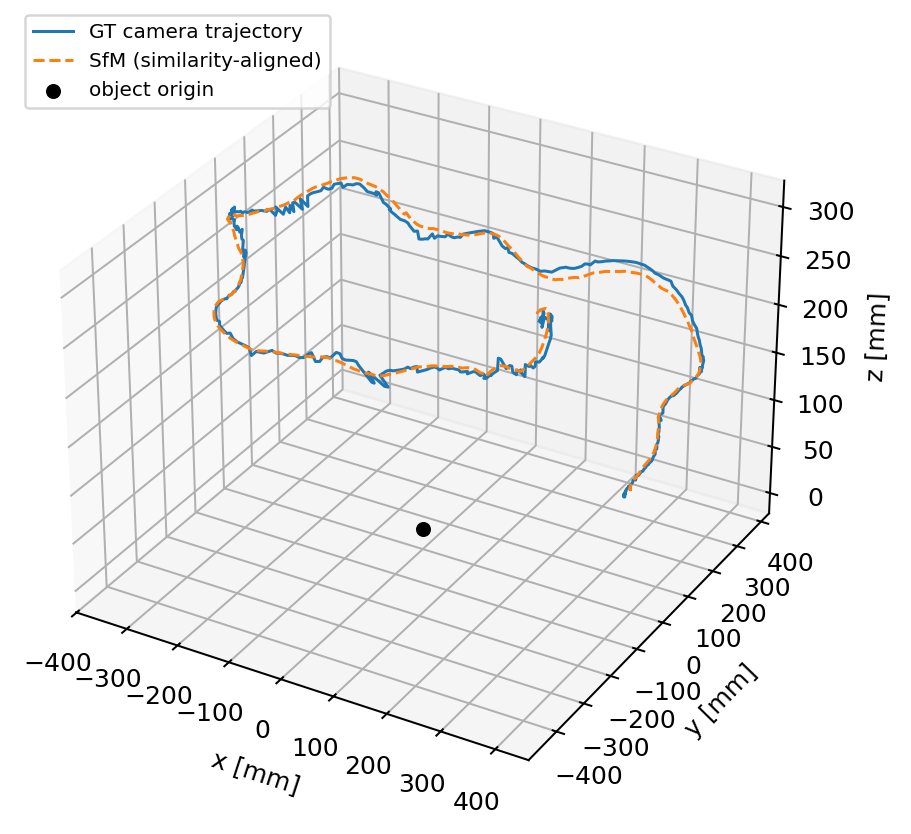}
  \caption{IndustryShapes \texttt{obj\_000005} (textureless, specular).}
  \label{fig:sfm_traj_industryshapes}
\end{subfigure}
\caption{Onboarding camera trajectories, shown as orthographic projections onto
the three coordinate planes (millimeters). The estimated trajectory (orange) is
the structure-from-motion reconstruction after similarity alignment; the
reference (blue) is the trajectory provided with the dataset. The two overlap
throughout both sequences, including on the textureless object, where the
sparse point cloud is forty times thinner per frame.}
\label{fig:sfm_trajectory}
\end{figure}

\begin{figure}
\centering
\begin{subfigure}{\linewidth}
  \centering
  \includegraphics[width=\textwidth]{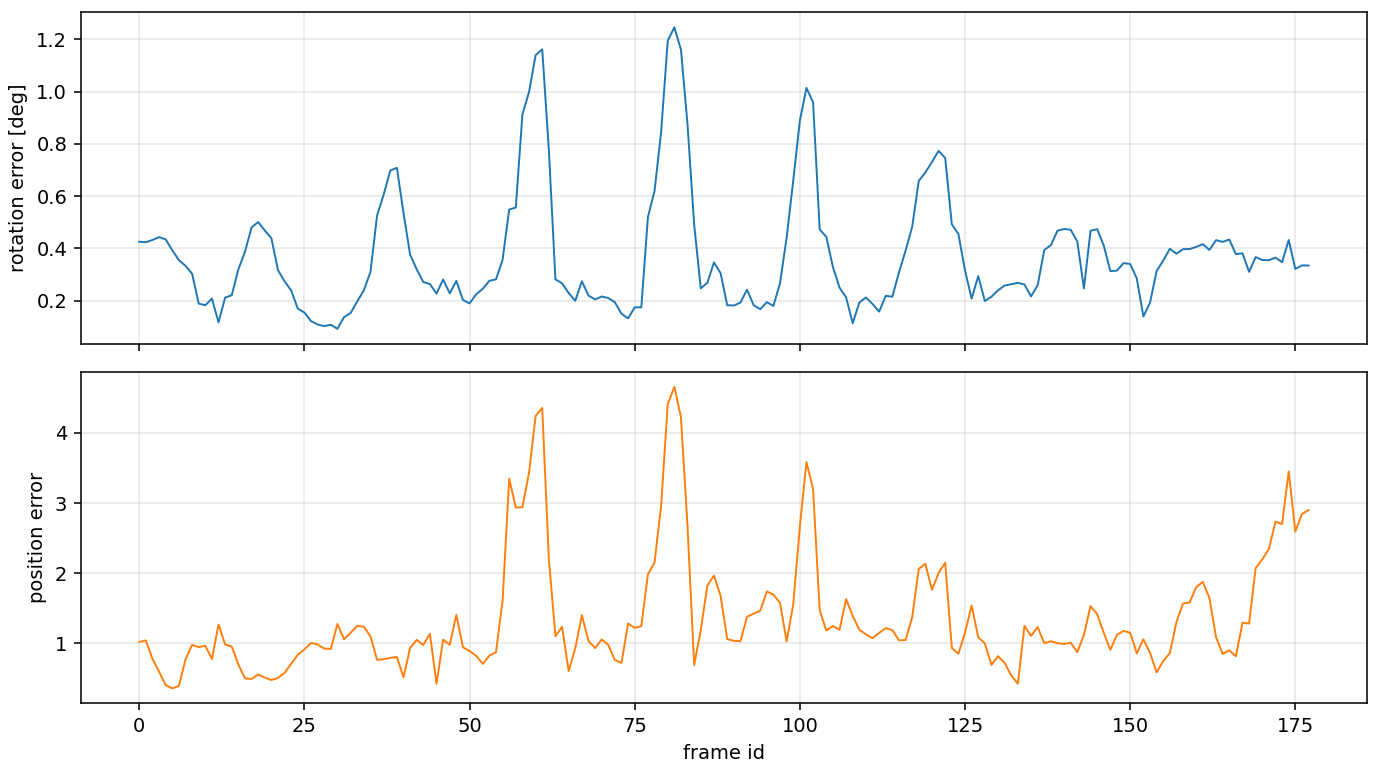}
  \caption{HOPEv2 \texttt{obj\_000001}: median $0.32^\circ$ / $1.09$\,mm.}
  \label{fig:sfm_err_hopev2}
\end{subfigure}\\[4pt]
\begin{subfigure}{\linewidth}
  \centering
  \includegraphics[width=\textwidth]{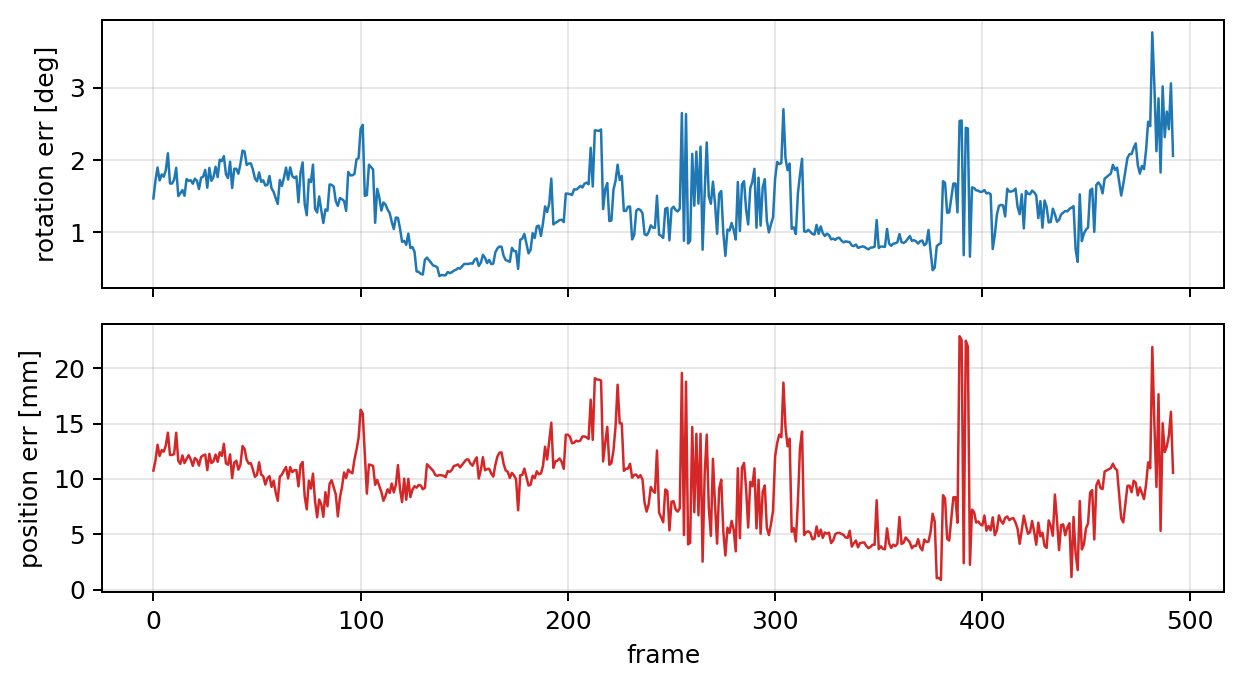}
  \caption{IndustryShapes \texttt{obj\_000005}: median $1.36^\circ$ / $9.47$\,mm.}
  \label{fig:sfm_err_industryshapes}
\end{subfigure}
\caption{Absolute rotation and position error per frame, after similarity
alignment. Note the difference in vertical scale between the two objects. On
both sequences the error is bounded and oscillates about a stable baseline
rather than accumulating along the trajectory, indicating a globally consistent
reconstruction. No frame fails to register on either object.}
\label{fig:sfm_error}
\end{figure}

Table~\ref{tab:sfm_accuracy} compares the estimated trajectories against the poses provided with the datasets. Ground truth is available for $482$ of the $522$ registered frames, and the comparison is restricted to those.

On the textured object the agreement is close to the annotation noise floor: $0.32^\circ$ and $1.09$\,mm median error. On the textureless object, the error grows to $1.36^\circ$ and $9.47$\,mm at a mean viewing distance of $447$\,mm. It is roughly $4\times$ worse in rotation and $9\times$ worse in position, and it is consistent with the limitation stated in the main paper. What matters is the \emph{form} the degradation takes:
\begin{itemize}
  \setlength{\itemsep}{0pt}
  \item \textbf{No frame fails.} The maximum position error over $482$ frames is $22.9$\,mm, and the maximum rotation error is $3.77^\circ$. There is no heavy tail and no diverging segment.
  \item \textbf{No drift.} Figures~\ref{fig:sfm_error} and~\ref{fig:sfm_trajectory} show the error varying smoothly along the sequence and returning to its baseline, rather than accumulating, the signature of a globally consistent reconstruction rather than a drifting one.
  \item \textbf{Local consistency is preserved.} The frame-to-frame relative rotation error stays at $0.19^\circ$ (median), an order of magnitude below the absolute error. Neighboring views, which is what MVS matching and 3DGS training consume, remain mutually consistent even where the trajectory as a whole is slightly deformed.
\end{itemize}

The relative \emph{direction} error is the one statistic that is much larger on IndustryShapes ($13.95^\circ$ vs $0.61^\circ$ median). This is an artifact of sampling density rather than of accuracy: the IndustryShapes sequence covers a single slow orbit in $522$ frames, so the baseline between consecutive cameras is very short and the direction of a near-zero displacement is poorly conditioned. The HOPEv2 sequence covers a much wider trajectory in $178$ frames. The quantity is reported for completeness and should not be read as a reconstruction failure.

\subsection{Discussion}
\label{sec:supp_sfm_discussion}
For context, the tightest BOP MSSD threshold is $5\%$ of the object diameter. The diameter of \texttt{obj\_000005} is $\sim$314\,mm, so the median onboarding position error of $9.47$\,mm is $3.0\%$ of the diameter and the $95$th percentile is $4.7\%$. Two practical consequences follow. First, the method does not depend on dataset-supplied onboarding poses and can be deployed on a video captured in the wild. Second, the requirement it does impose is on the \emph{scene}, not on the object: onboarding should be captured with the object in a feature-bearing environment. This is already what the BOP onboarding protocol prescribes, and it is a far weaker requirement than demanding texture on the object itself.

\section{Per-Object Quantitative Results}
\label{sec:supp_per_object}

This section reports per-object quantitative results that complement the aggregate results presented in the main paper.

For HOPEv2, Table~\ref{tab:hopev2_perobj} reports the per-object metrics of NovaSplat on the validation set. We use the validation set for this per-object analysis because ground-truth annotations are available for this split.

\begin{table}[t]
\centering
\setlength{\tabcolsep}{8pt}
\begin{tabular}{@{}l ccc}
\toprule
Object & \textbf{AP} $\uparrow$ & AP$_\text{MSSD}$ $\uparrow$ & AP$_\text{MSPD}$ $\uparrow$ \\
\midrule
Obj 01 & 0.77 & 0.71 & 0.84 \\ 
Obj 02 & 0.83 & 0.75 & 0.92 \\ 
Obj 03 & 0.83 & 0.76 & 0.91 \\ 
Obj 04 & 0.48 & 0.38 & 0.58 \\ 
Obj 05 & 0.75 & 0.69 & 0.80 \\ 
Obj 06 & 0.65 & 0.60 & 0.69 \\ 
Obj 07 & 0.46 & 0.42 & 0.50 \\ 
Obj 08 & 0.53 & 0.52 & 0.55 \\ 
Obj 09 & 0.75 & 0.67 & 0.84 \\ 
Obj 10 & 0.58 & 0.49 & 0.66 \\
Obj 11 & 0.93 & 0.87 & 0.98 \\
Obj 12 & 0.55 & 0.48 & 0.63 \\
Obj 13 & 0.87 & 0.81 & 0.94 \\
Obj 14 & 0.88 & 0.80 & 0.95 \\
Obj 15 & 0.56 & 0.51 & 0.61 \\
Obj 16 & 0.75 & 0.67 & 0.82 \\
Obj 17 & 0.57 & 0.51 & 0.62 \\
Obj 18 & 0.83 & 0.75 & 0.91 \\
Obj 19 & 0.89 & 0.81 & 0.97 \\
Obj 20 & 0.68 & 0.58 & 0.78 \\
Obj 21 & 0.69 & 0.63 & 0.75 \\
Obj 22 & 0.76 & 0.65 & 0.86 \\
Obj 23 & 0.42 & 0.37 & 0.47 \\
Obj 24 & 0.70 & 0.61 & 0.79 \\
Obj 25 & 0.55 & 0.47 & 0.64 \\
Obj 26 & 0.78 & 0.68 & 0.88 \\
Obj 27 & 0.67 & 0.57 & 0.77 \\
Obj 28 & 0.77 & 0.65 & 0.88 \\
\bottomrule
\end{tabular}
\caption{Per-object Average Precision (AP) of NovaSplat on the HOPEv2 validation set, along with the MSSD and MSPD components.}
\label{tab:hopev2_perobj}
\end{table}

As reported in Table~\ref{tab:hopev2_perobj}, Object 23 yields the lowest pose estimation performance across the dataset. We focus our failure analysis specifically on this object because it serves as the clearest example of the fundamental difficulty in resolving pose ambiguities from visual or geometric cues. Object 23 exhibits near-perfect physical symmetry along its major axes, with only minor variations distinguishing its canonical orientation. For any visual pose estimation method, capturing these minor geometric or textural variations presents a significant challenge. During the rendering and refinement process, NovaSplat can occasionally converge to a local minimum that corresponds to a symmetric counterpart of the true pose. 

Figure~\ref{fig:symmetry_failure} illustrates an example where NovaSplat predicts a pose that is flipped by approximately 180 degrees relative to the ground-truth annotation. While the predicted overlay tightly matches the visible object boundaries and appears visually correct, the underlying rotation matrix incurs a large angular error. This highlights a common failure mode in nearly symmetric objects, demonstrating the extreme difficulty of visually resolving poses when only minor asymmetric variations exist.

\begin{figure}[ht]
\centering
\setlength{\tabcolsep}{1pt}
\begin{tabular}{ccc}

\includegraphics[trim=603 176 917 504, clip, width=0.32\linewidth]{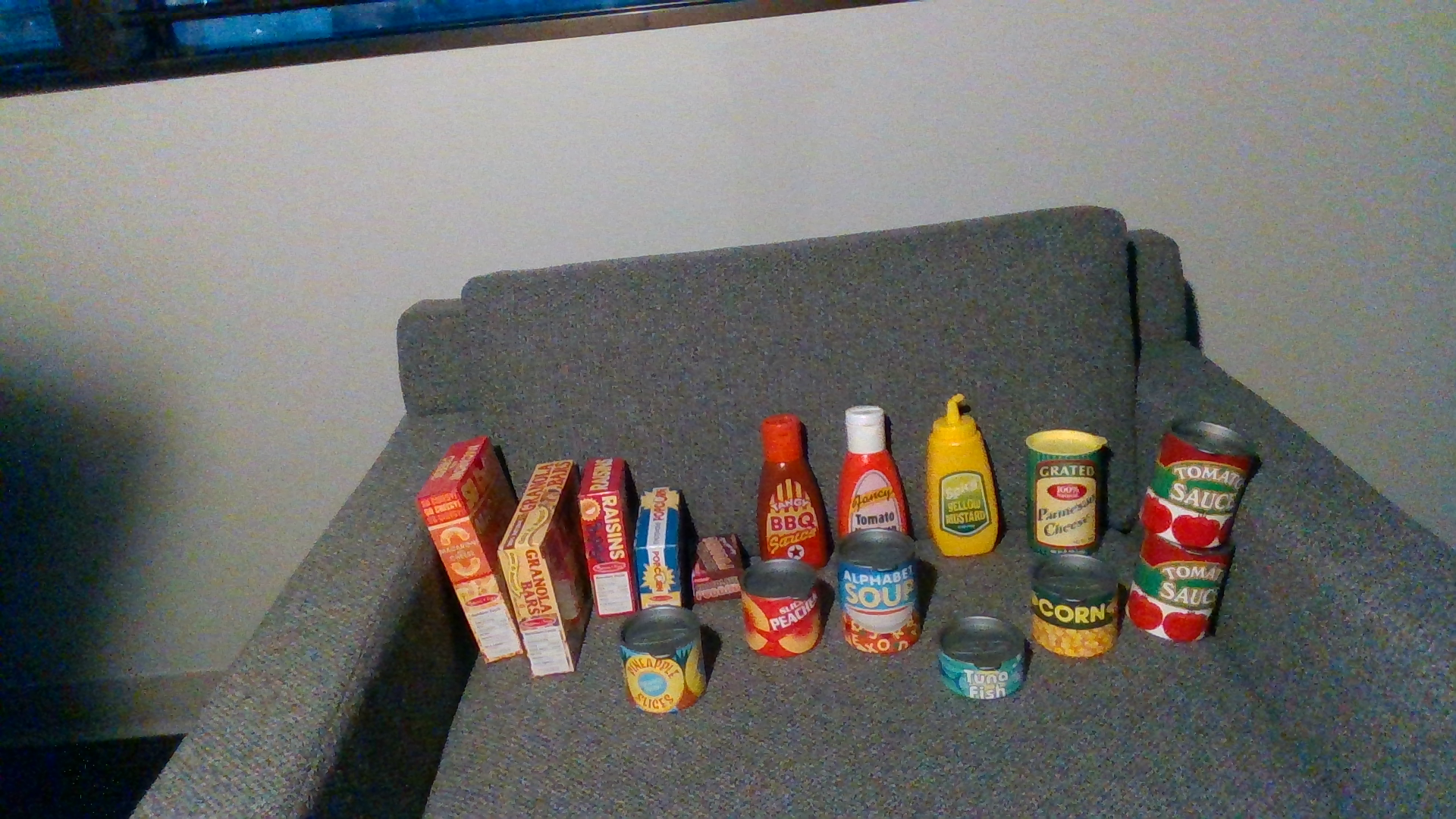} &
\includegraphics[trim=603 176 917 504, clip, width=0.32\linewidth]{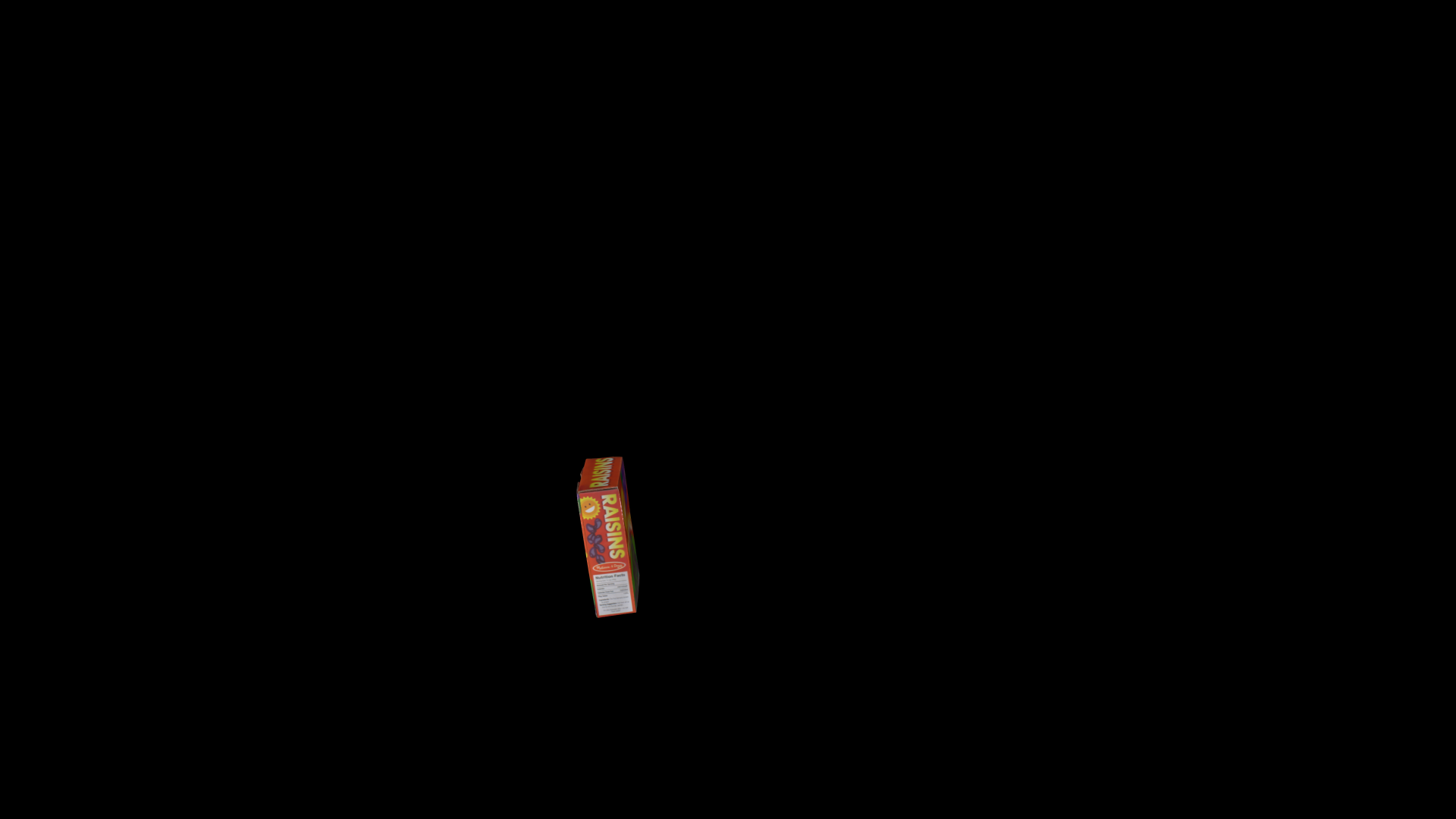} &
\includegraphics[trim=603 176 917 504, clip, width=0.32\linewidth]{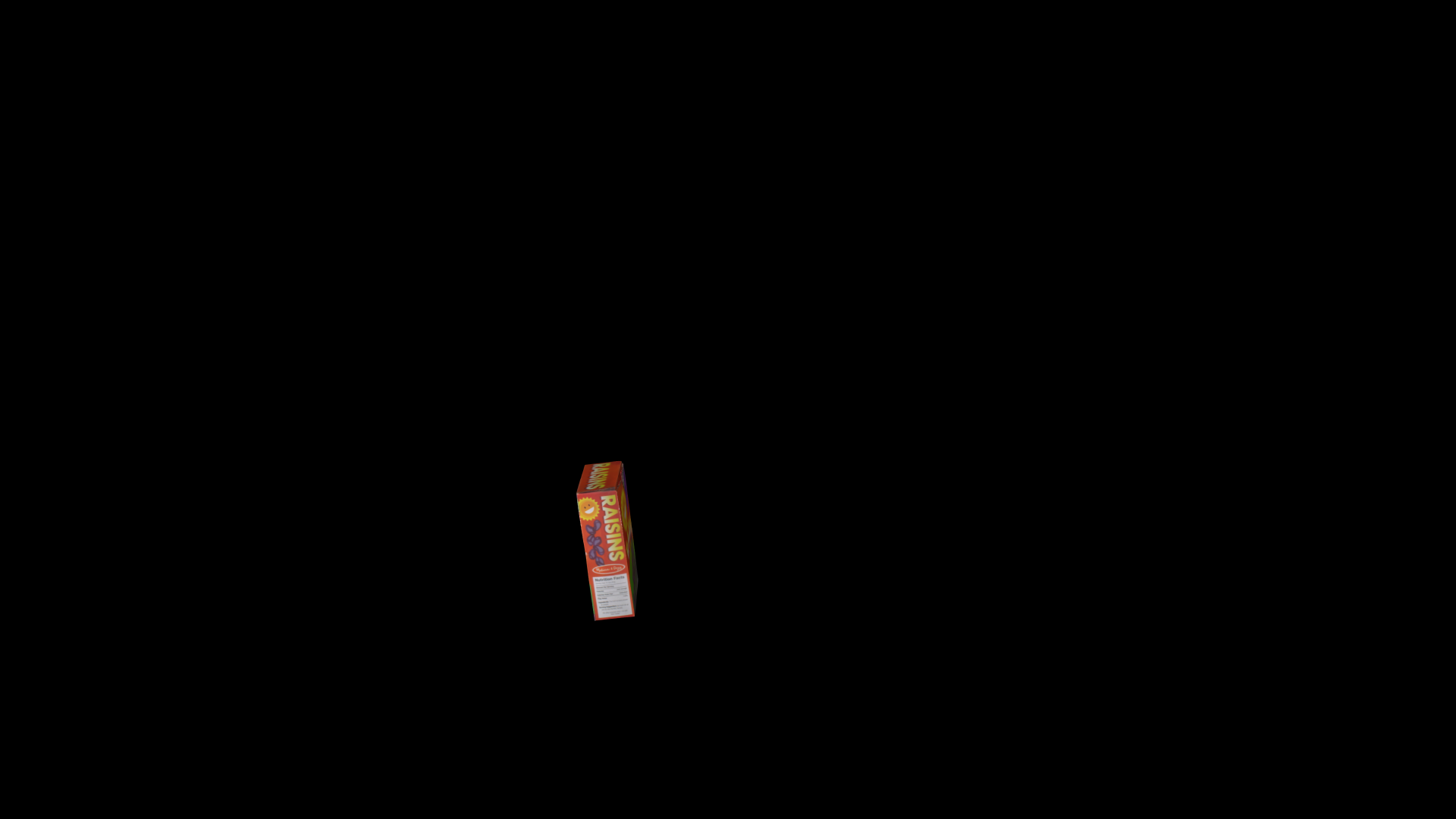} \\

\includegraphics[trim=194 285 1326 395, clip, width=0.32\linewidth]{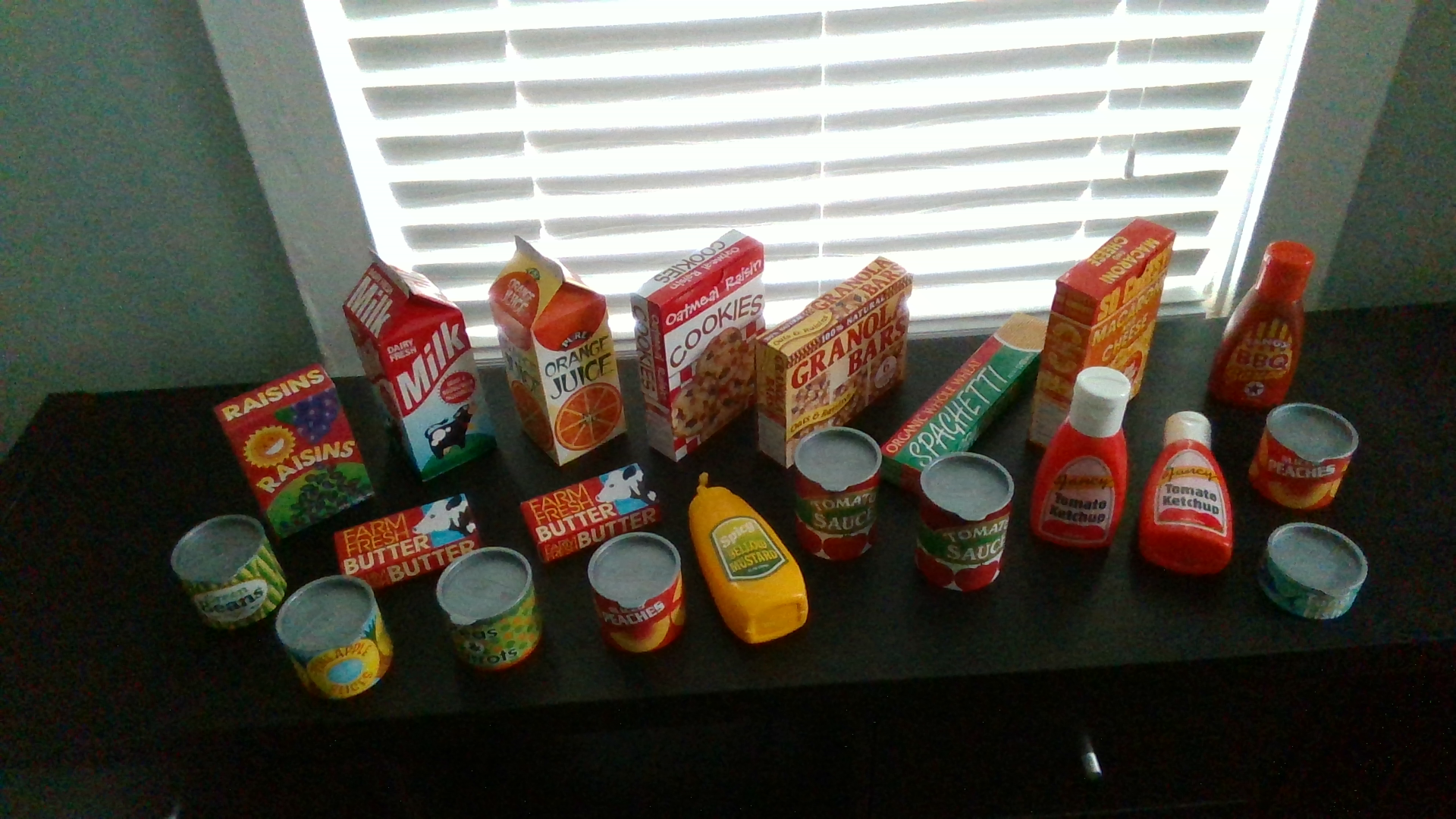} &
\includegraphics[trim=194 285 1326 395, clip, width=0.32\linewidth]{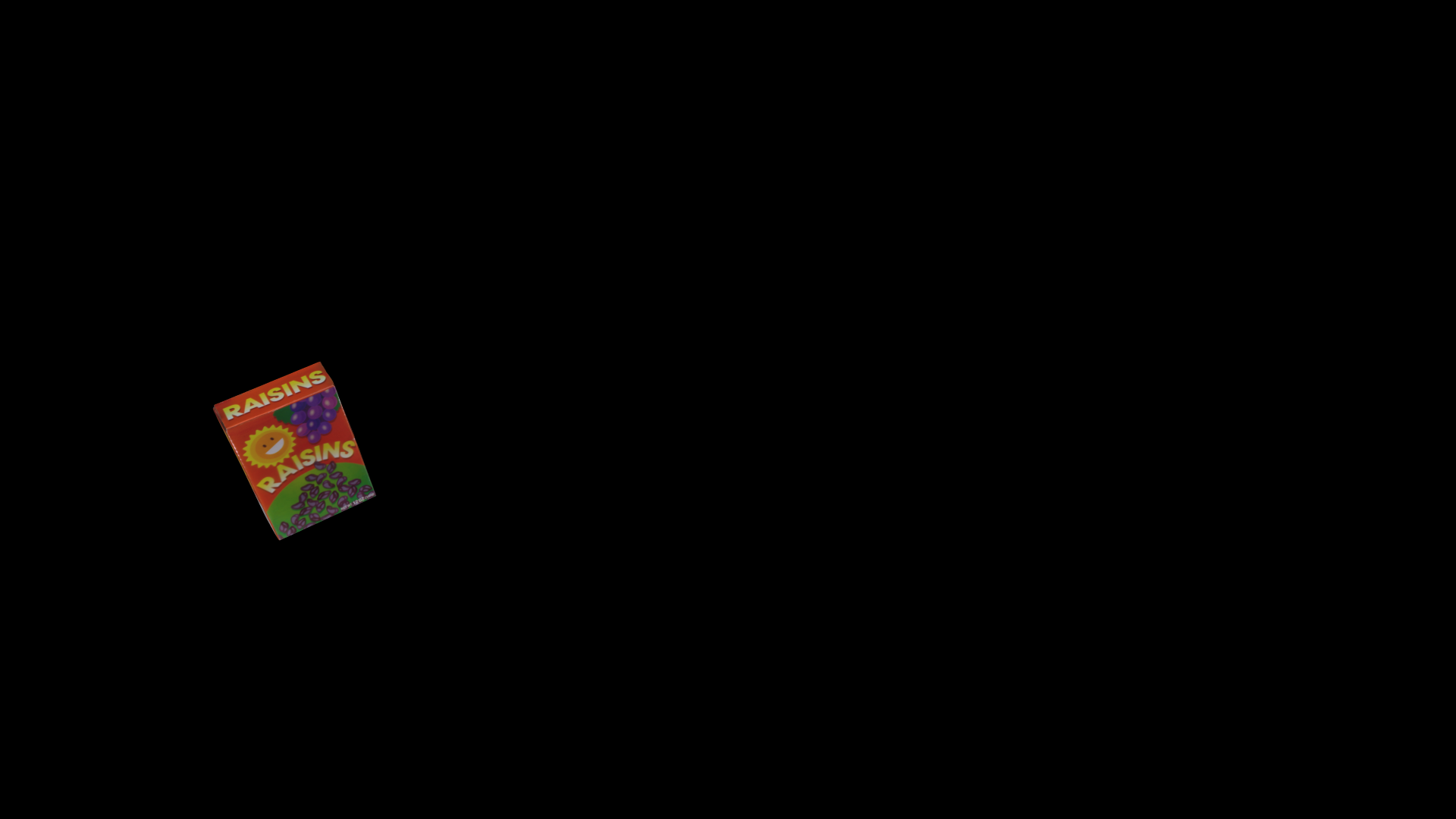} &
\includegraphics[trim=194 285 1326 395, clip, width=0.32\linewidth]{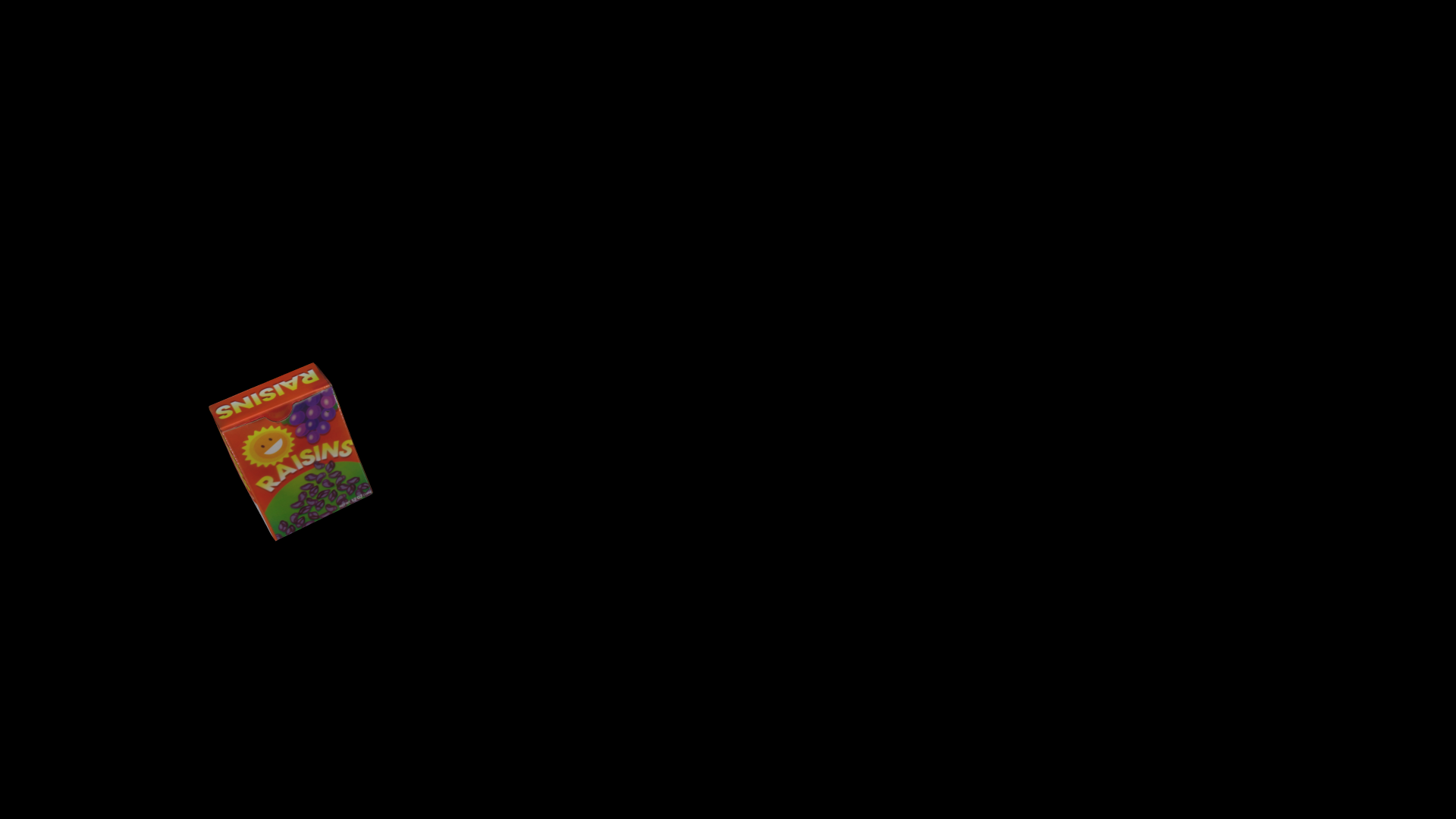} \\

\includegraphics[trim=813 94 707 586, clip, width=0.32\linewidth]{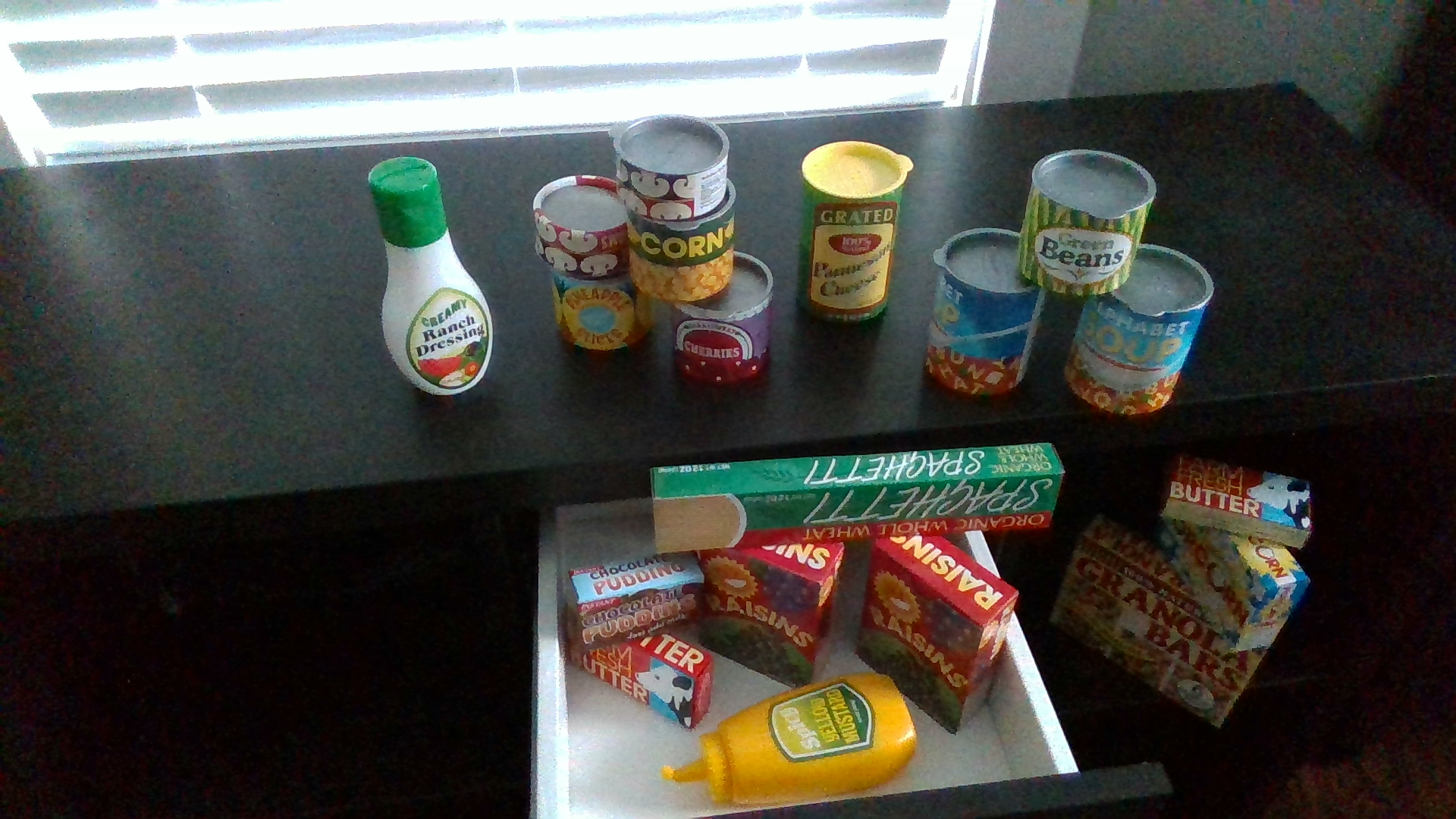} &
\includegraphics[trim=813 94 707 586, clip, width=0.32\linewidth]{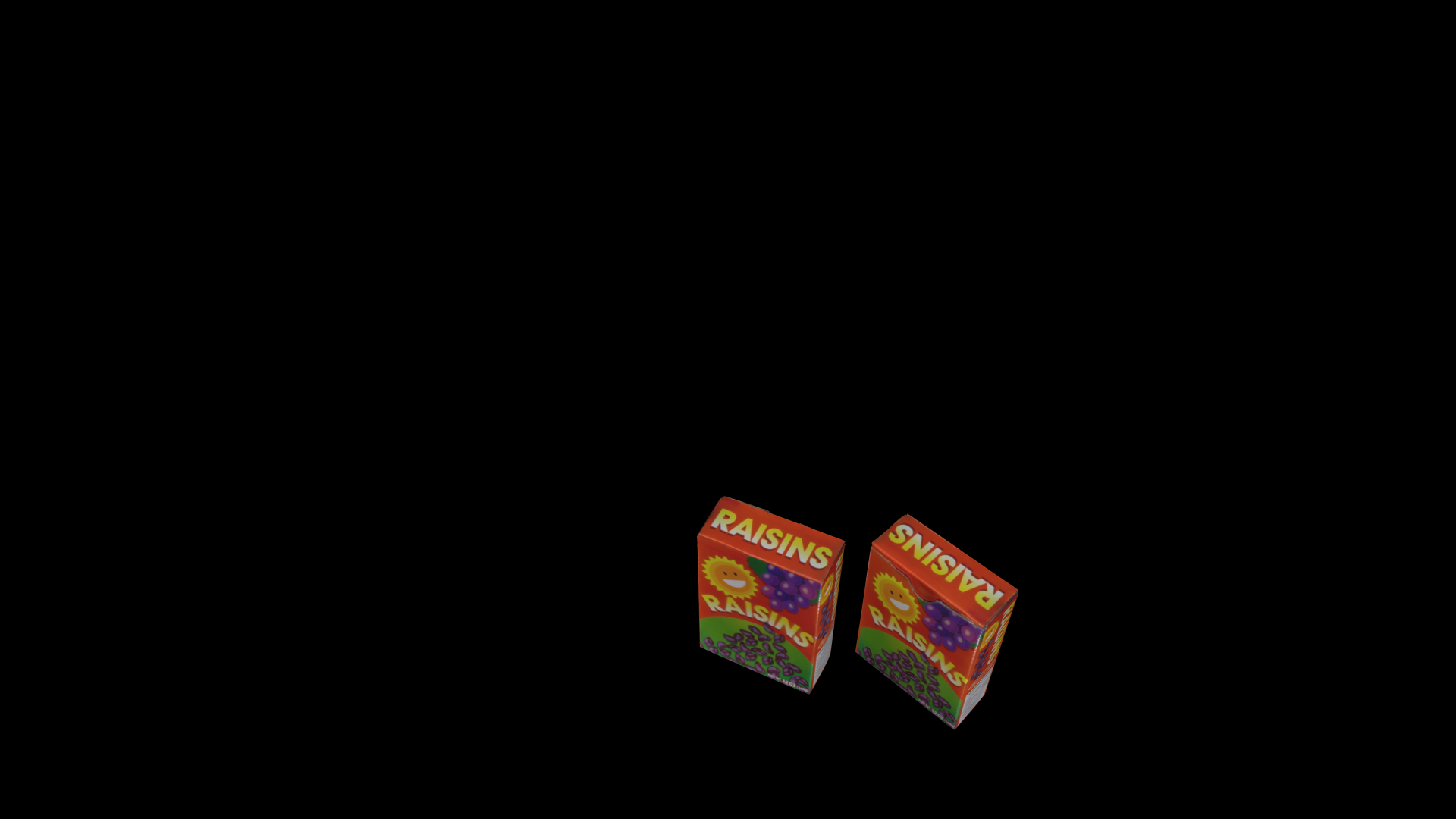} &
\includegraphics[trim=813 94 707 586, clip, width=0.32\linewidth]{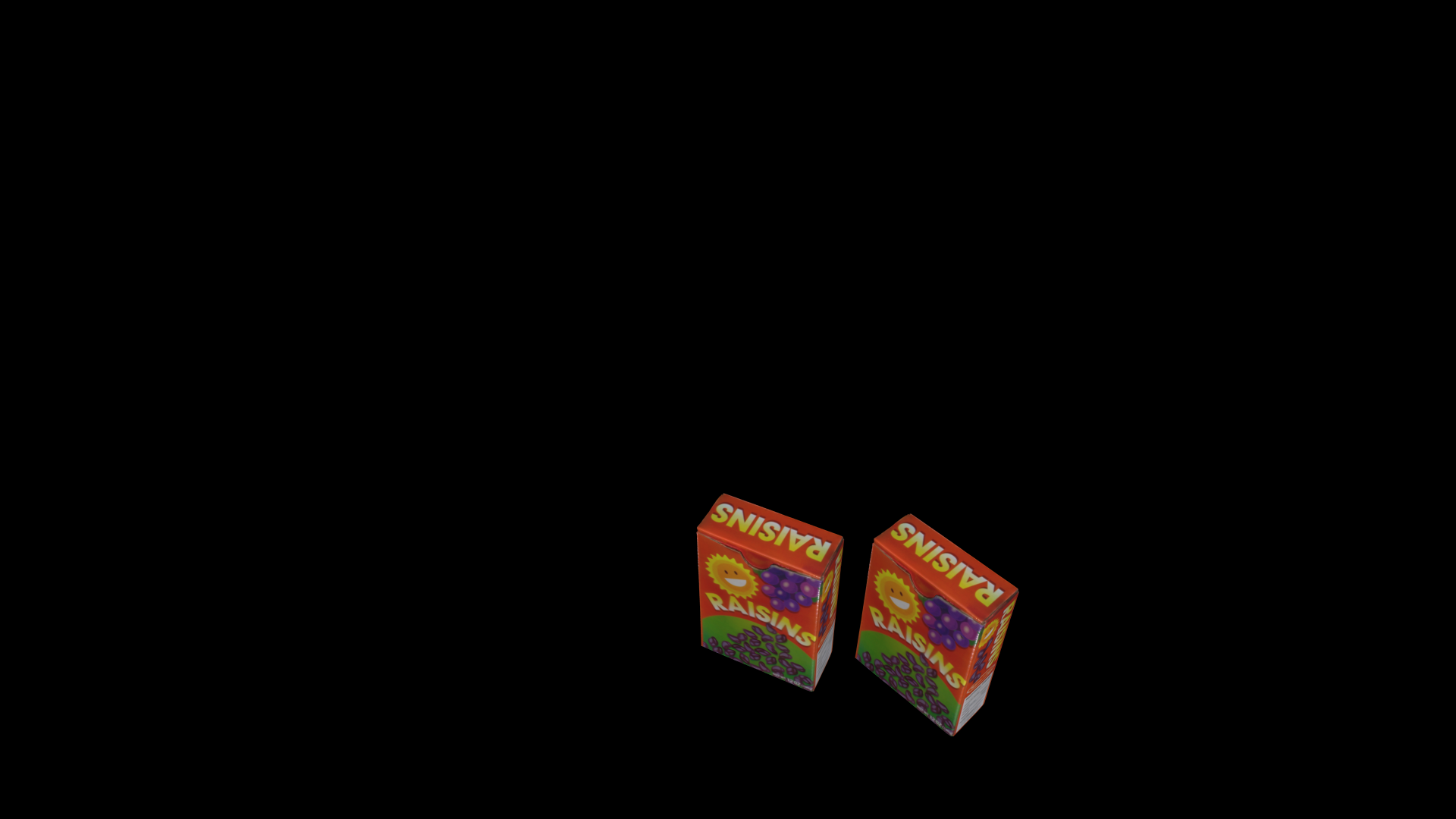} \\

Original & Ground Truth  & NovaSplat (Ours) \\

\end{tabular}
\caption{Impact of object symmetry on NovaSplat performance (HOPEv2, Object 23). Across three diverse viewpoints, the object's near-symmetry causes NovaSplat to predict an inverted pose ($\sim 180$ degrees error).}
\label{fig:symmetry_failure}
\end{figure}

\section{HOPEv2 Onboarding Data Inconsistencies}
\label{sec:hopev2_onboarding_data_inconsistencies}

The HOPEv2 dataset contains object instances whose physical configuration differs between the onboarding and query images. In particular, articulated or loosely attached components are not always in the same relative pose across the two splits. Figure~\ref{fig:hopev2_inconsistency} shows an example in which the cap orientation relative to the printed label differs between the images from the onboarding and validation sets. In the onboarding image, the cap is rotated sideways, whereas in the image from the validation set, it is aligned above the label.

This inconsistency creates a mismatch between the object geometry reconstructed from the onboarding sequence and the object in the query images. For methods that build an explicit 3D reference representation from the onboarding data, such as 3DGS-based refinement approaches, this mismatch can affect pose refinement. The rendered reference model may contain local geometric structures whose relative arrangement does not correspond to the query observation, causing the refinement objective to optimize against inconsistent geometry.

Methods that rely primarily on global learned features or 2D image-space cues may be less affected by these local 3D discrepancies. Consequently, results on such instances should be interpreted with care, as the object in the query images may not exactly match the onboarding-derived reference geometry.

\section{Qualitative Comparisons}
\label{sec:supp_qualitative}

\begin{figure}[t]
\centering
\includegraphics[trim=500 460 700 260, clip, width=0.48\linewidth]{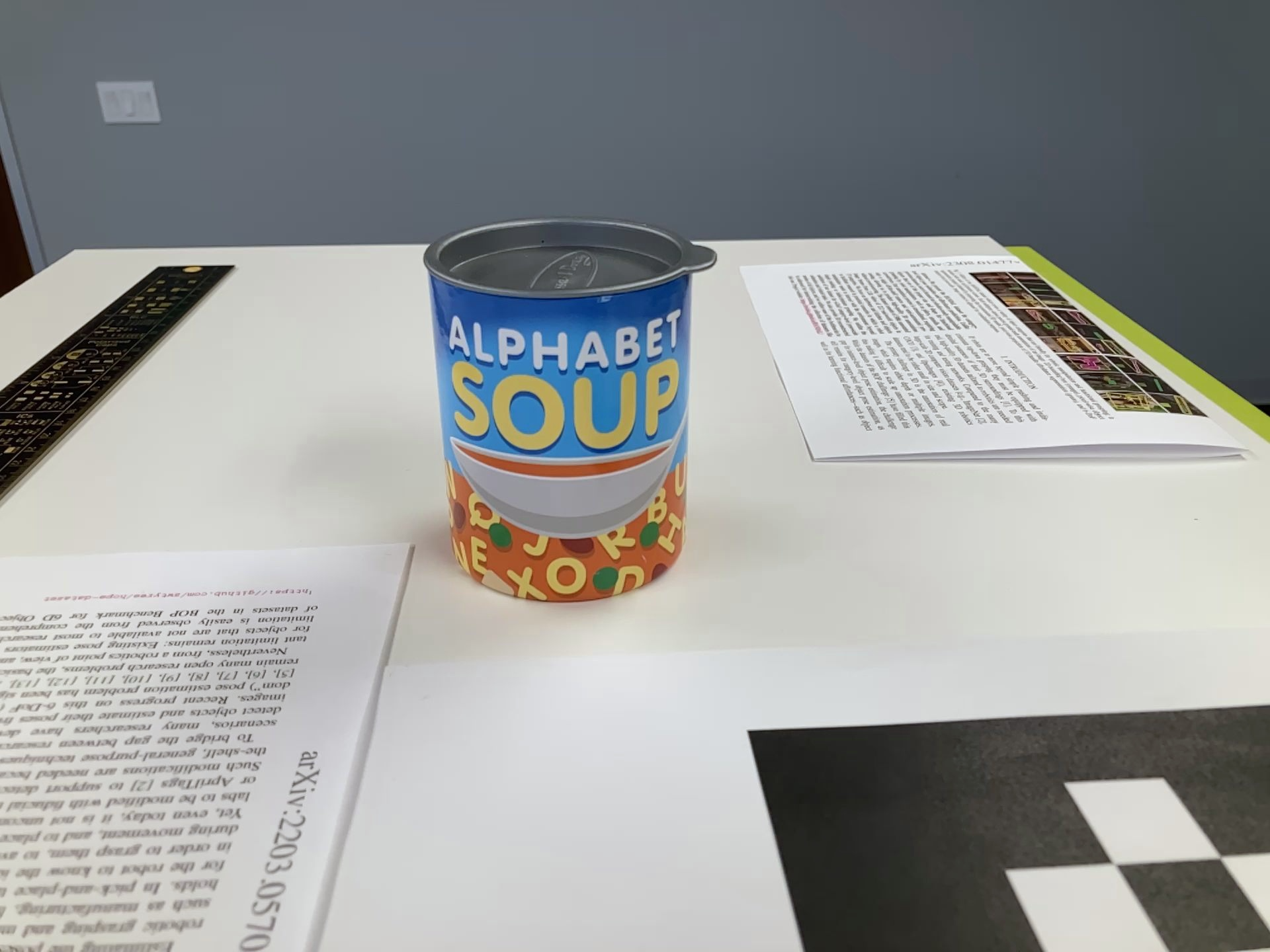}
\hfill
\includegraphics[trim=1000 150 600 610, clip, width=0.48\linewidth]{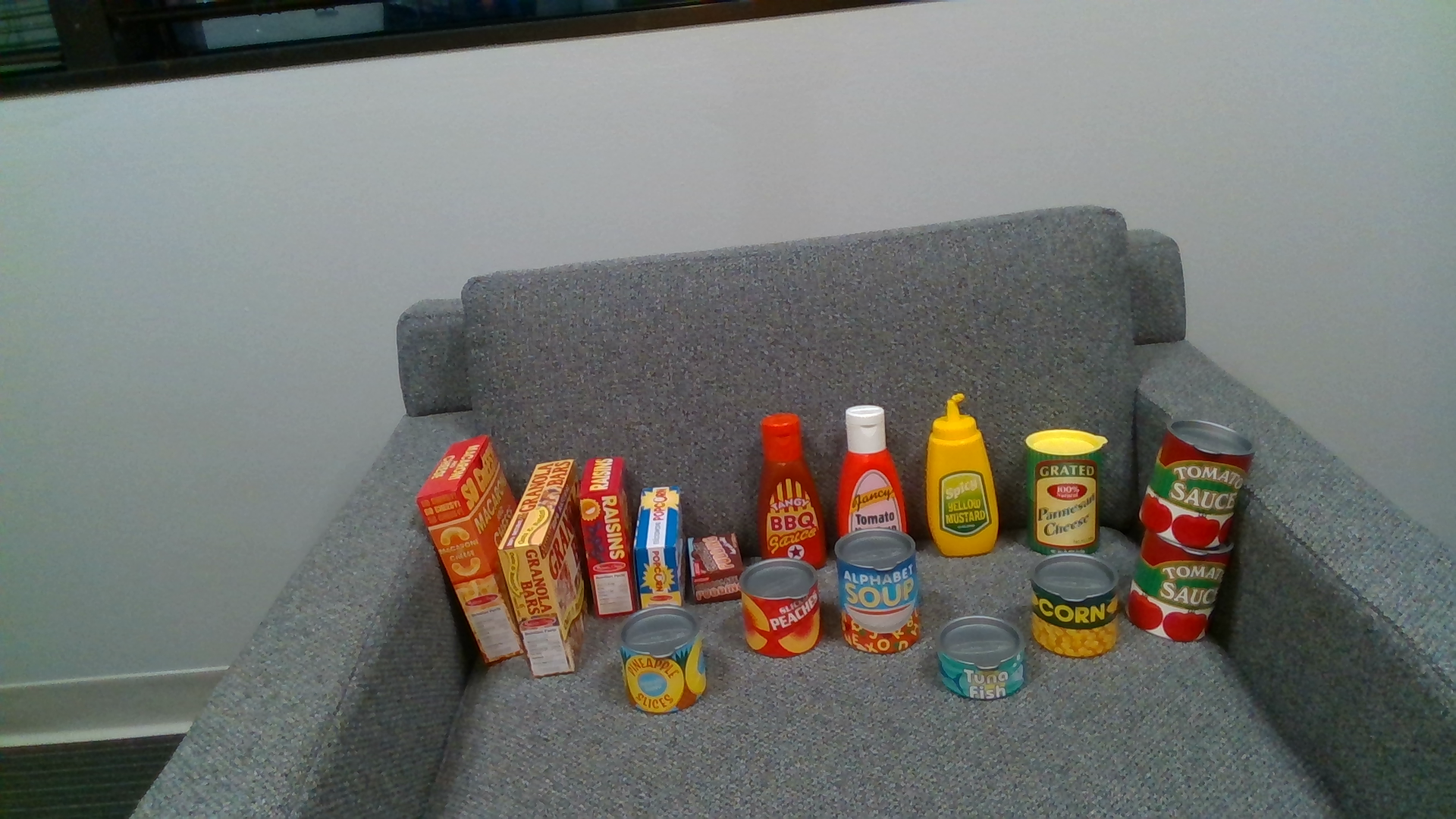}
\caption{Example of a geometric inconsistency in HOPEv2. Left: onboarding image. Right: image from validation set. The cap orientation relative to the printed label differs between the two splits, resulting in a mismatch between the onboarding-derived reference geometry and the validation observation.}
\label{fig:hopev2_inconsistency}
\end{figure}

Figure~\ref{fig:hopev2_qualitative} shows additional qualitative comparisons on the HOPEv2 test set, complementing the results presented in the main paper. The examples include challenging scenes with multiple objects, partial occlusions, clutter, and visually similar packaging, where accurate 6D pose estimation requires both reliable object localization and precise pose refinement.

Across these examples, NovaSplat produces predictions that are better aligned with the visible object extent and orientation than the competing model-free baselines. By optimizing the pose against the reconstructed 3D representation, NovaSplat yields tighter alignment between the projected object geometry and the image evidence.

The comparison also highlights common failure modes of the baseline methods. GFreeDet2, OPFormer, and NeMO can produce plausible detections, but their predicted poses are often less consistent with the object boundaries, especially under occlusion or when only a limited portion of the object is visible. In cluttered scenes, these methods may also exhibit larger orientation errors or less stable box projections. In contrast, NovaSplat maintains more geometrically consistent predictions, suggesting that the 3DGS-based reference representation provides useful constraints during refinement.

Like all evaluated methods, NovaSplat may occasionally produce false positive predictions. These predictions are primarily inherited from misdetections made by the initial object detection. Our method attempts to align these erroneous proposals, ultimately preserving them in the final scene prediction. While NovaSplat mitigates many of the failure modes seen in competing methods, explicitly filtering out these inherited misdetections remains an interesting direction for future work.

Furthermore, Figures~\ref{fig:is_classic_qualitative} and~\ref{fig:is_extended_qualitative} present the qualitative results on the IndustryShapes Classic and Extended datasets, respectively. These visualizations demonstrate that our model-free NovaSplat robustly handles challenging, textureless industrial parts, achieving highly accurate pose alignments that are competitive with state-of-the-art model-based methods, even under extreme viewpoint variations.

\begin{figure*}[t]
\centering
\setlength{\tabcolsep}{1pt}
\begin{tabular}{ccccc}

\includegraphics[width=0.196\linewidth]{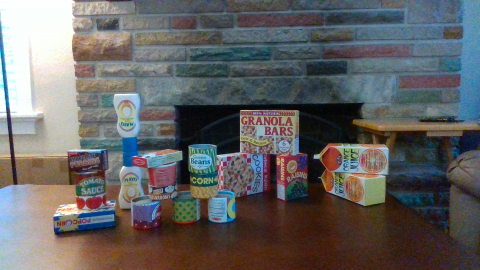} &
\includegraphics[width=0.196\linewidth]{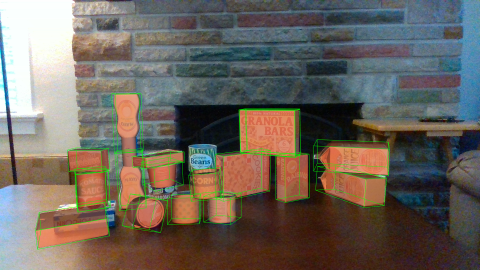} &
\includegraphics[width=0.196\linewidth]{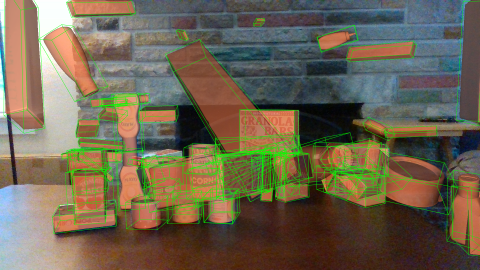} &
\includegraphics[width=0.196\linewidth]{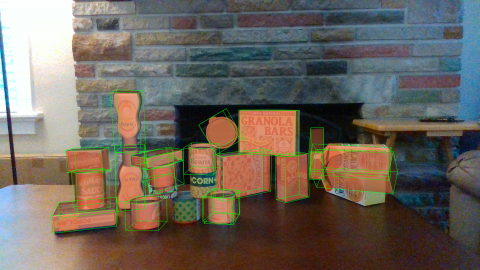} &
\includegraphics[width=0.196\linewidth]{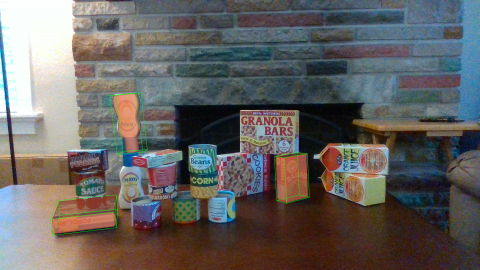} \\

\includegraphics[width=0.196\linewidth]{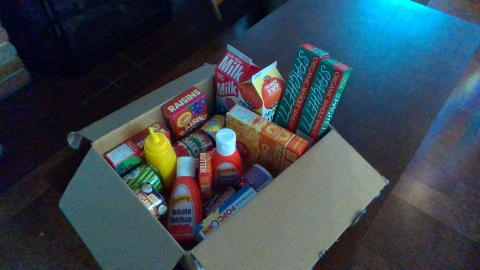} &
\includegraphics[width=0.196\linewidth]{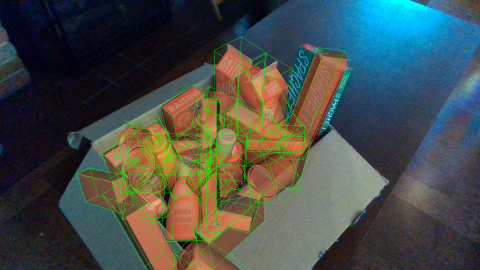} &
\includegraphics[width=0.196\linewidth]{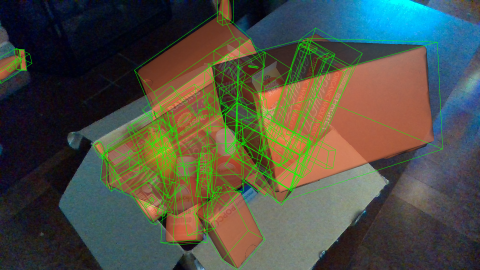} &
\includegraphics[width=0.196\linewidth]{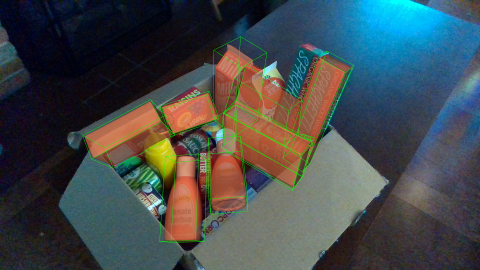} &
\includegraphics[width=0.196\linewidth]{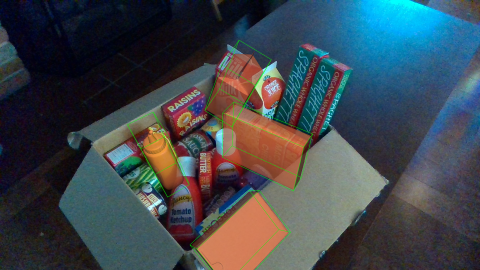} \\

\includegraphics[width=0.196\linewidth]{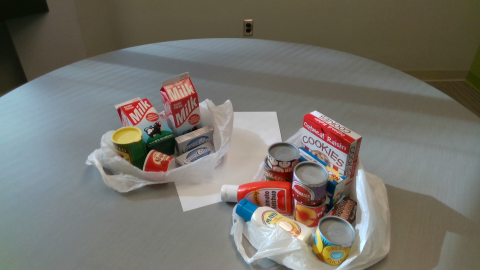} &
\includegraphics[width=0.196\linewidth]{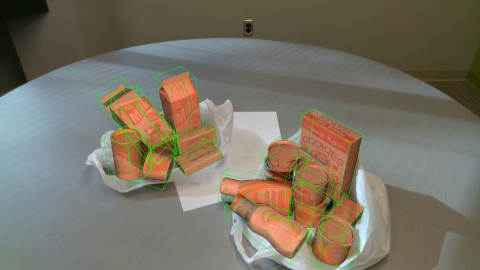} &
\includegraphics[width=0.196\linewidth]{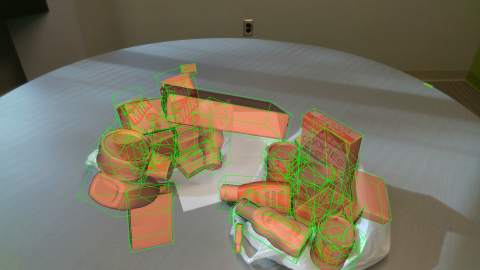} &
\includegraphics[width=0.196\linewidth]{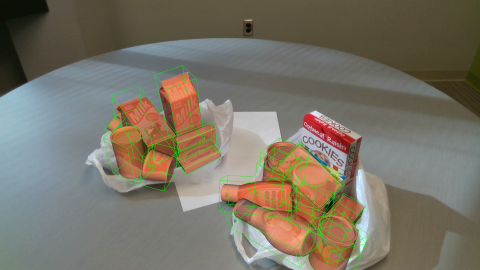} &
\includegraphics[width=0.196\linewidth]{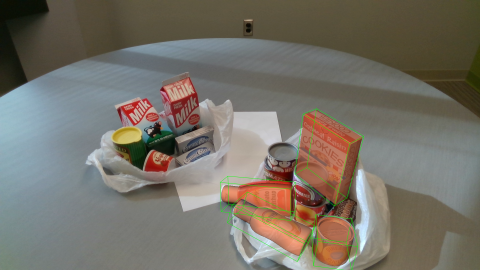} \\

\includegraphics[width=0.196\linewidth]{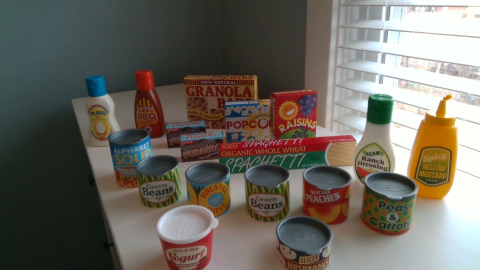} &
\includegraphics[width=0.196\linewidth]{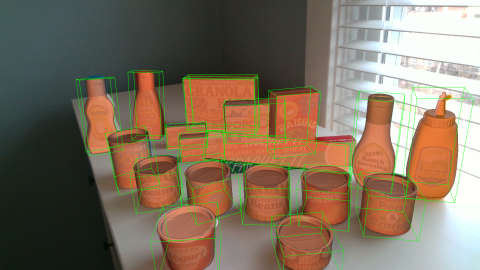} &
\includegraphics[width=0.196\linewidth]{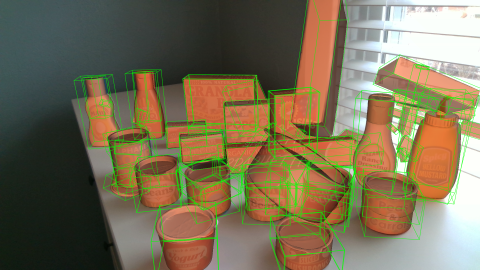} &
\includegraphics[width=0.196\linewidth]{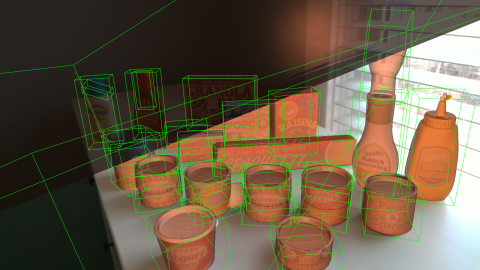} &
\includegraphics[width=0.196\linewidth]{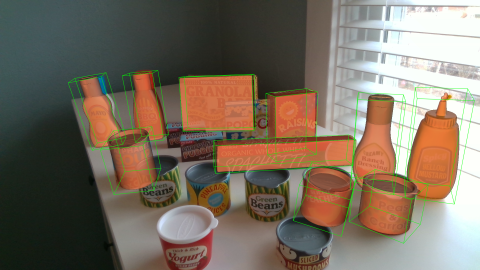} \\

\includegraphics[width=0.196\linewidth]{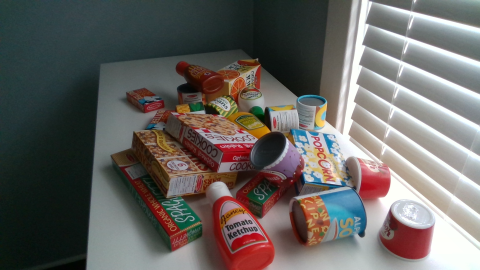} &
\includegraphics[width=0.196\linewidth]{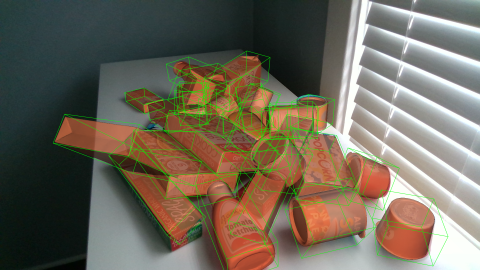} &
\includegraphics[width=0.196\linewidth]{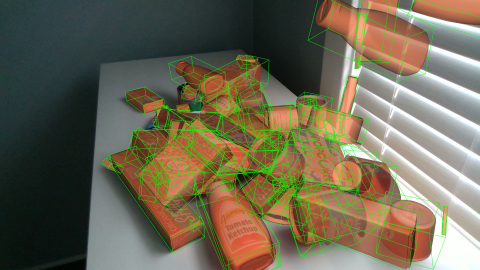} &
\includegraphics[width=0.196\linewidth]{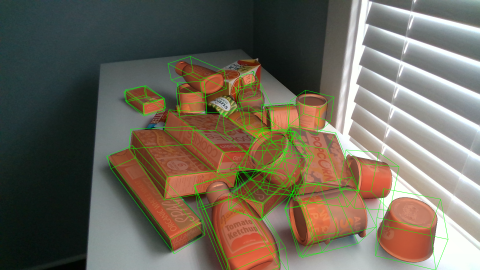} &
\includegraphics[width=0.196\linewidth]{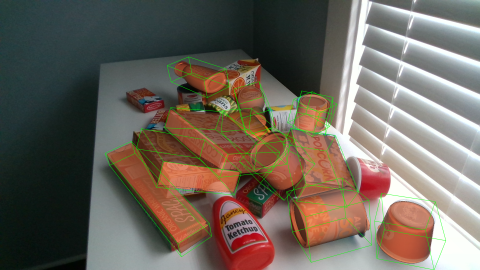} \\

\includegraphics[width=0.196\linewidth]{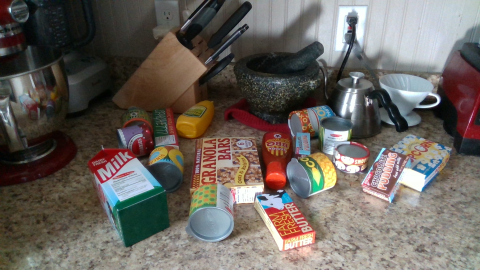} &
\includegraphics[width=0.196\linewidth]{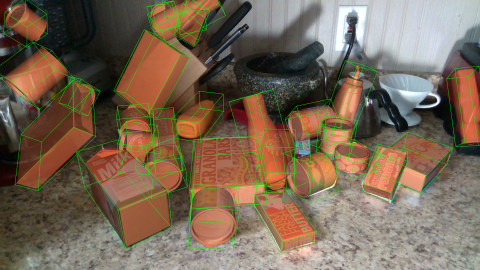} &
\includegraphics[width=0.196\linewidth]{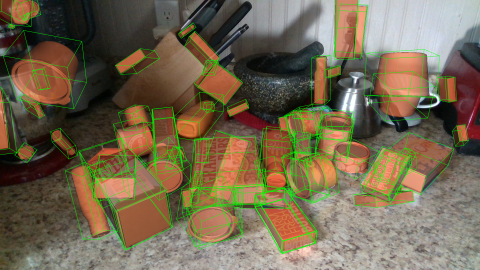} &
\includegraphics[width=0.196\linewidth]{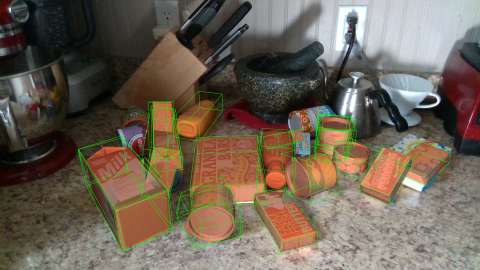} &
\includegraphics[width=0.196\linewidth]{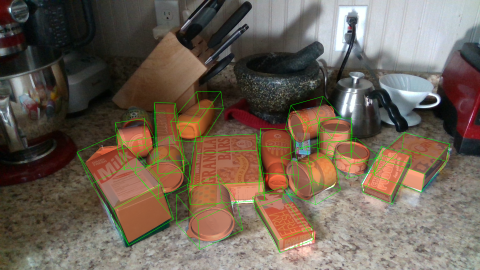} \\

\includegraphics[width=0.196\linewidth]{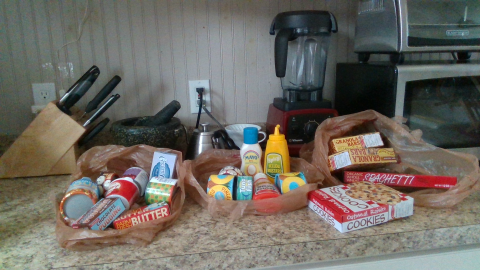} &
\includegraphics[width=0.196\linewidth]{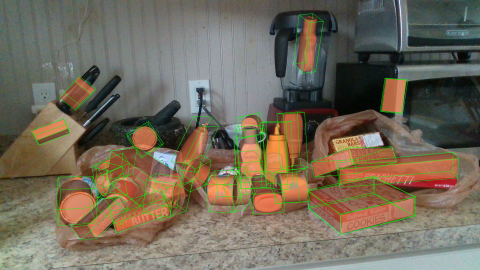} &
\includegraphics[width=0.196\linewidth]{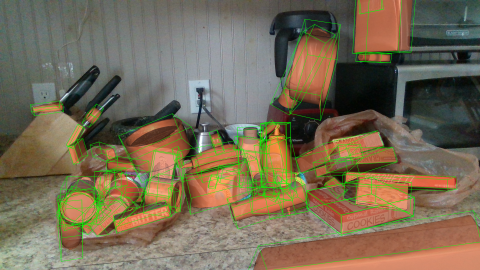} &
\includegraphics[width=0.196\linewidth]{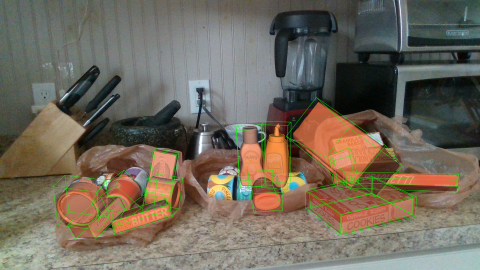} &
\includegraphics[width=0.196\linewidth]{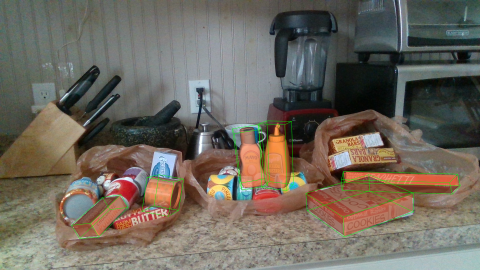} \\

\includegraphics[width=0.196\linewidth]{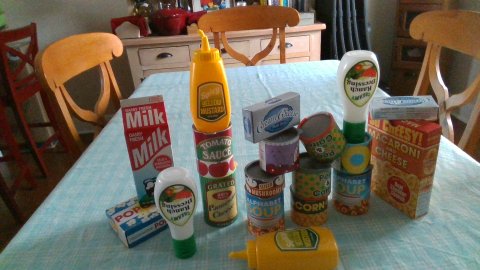} &
\includegraphics[width=0.196\linewidth]{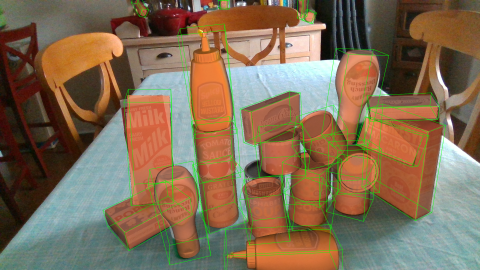} &
\includegraphics[width=0.196\linewidth]{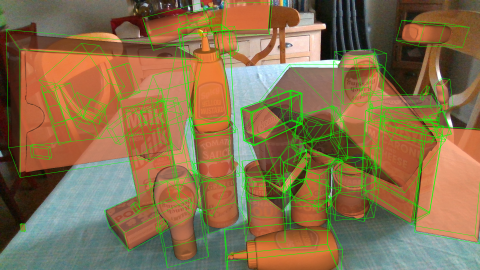} &
\includegraphics[width=0.196\linewidth]{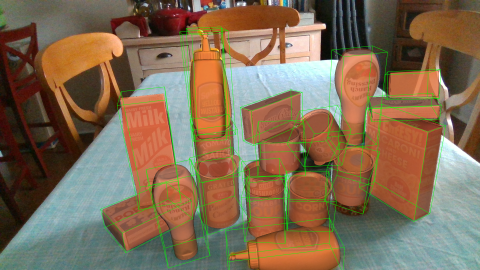} &
\includegraphics[width=0.196\linewidth]{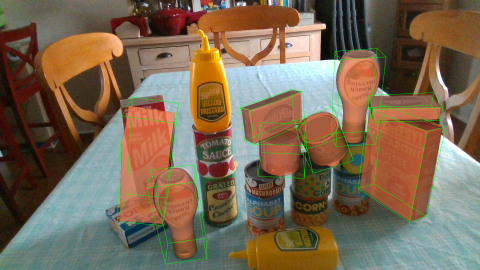} \\

\includegraphics[width=0.196\linewidth]{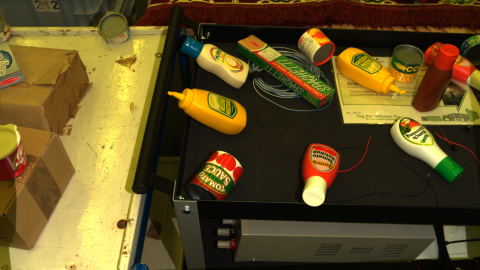} &
\includegraphics[width=0.196\linewidth]{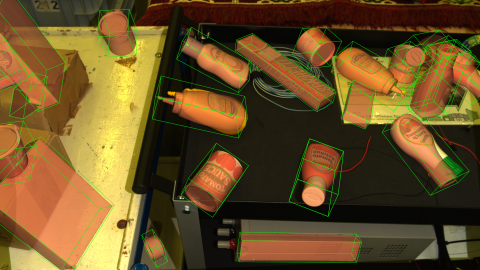} &
\includegraphics[width=0.196\linewidth]{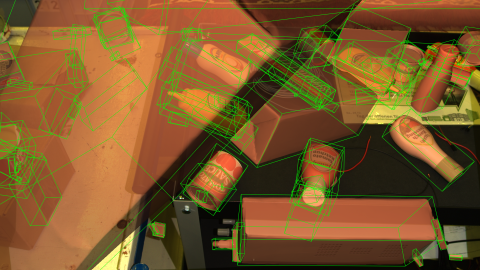} &
\includegraphics[width=0.196\linewidth]{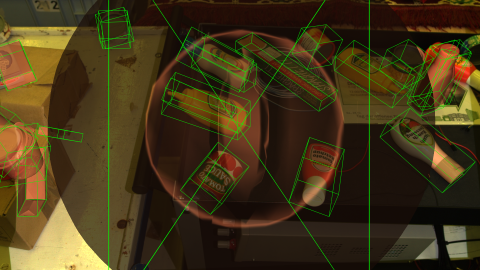} &
\includegraphics[width=0.196\linewidth]{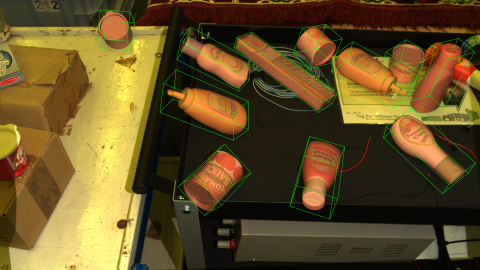} \\

Original & NovaSplat (Ours) & GFreeDet2~\cite{Wang_ICRA_2026} & OPFormer~\cite{Moroz_WACV_2026} & NeMO~\cite{Jung_3DV_2026} \\

\end{tabular}
\caption{Qualitative comparison on the HOPEv2 dataset. Bounding boxes highlight the predicted 6D poses.}
\label{fig:hopev2_qualitative}
\end{figure*}

\begin{figure*}[t]
\centering
\setlength{\tabcolsep}{1pt}
\begin{tabular}{cccc}

\includegraphics[width=0.245\linewidth]{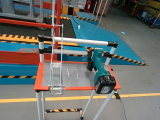} &
\includegraphics[width=0.245\linewidth]{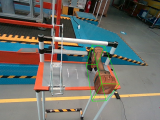} &
\includegraphics[width=0.245\linewidth]{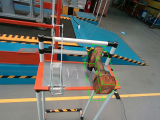} &
\includegraphics[width=0.245\linewidth]{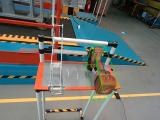} \\

\includegraphics[width=0.245\linewidth]{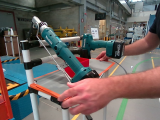} &
\includegraphics[width=0.245\linewidth]{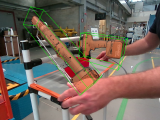} &
\includegraphics[width=0.245\linewidth]{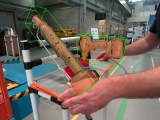} &
\includegraphics[width=0.245\linewidth]{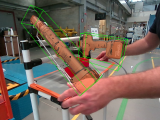} \\

\includegraphics[width=0.245\linewidth]{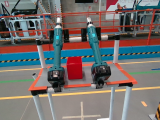} &
\includegraphics[width=0.245\linewidth]{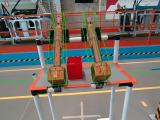} &
\includegraphics[width=0.245\linewidth]{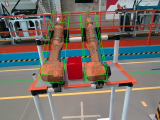} &
\includegraphics[width=0.245\linewidth]{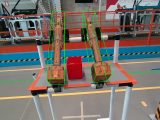} \\

\includegraphics[width=0.245\linewidth]{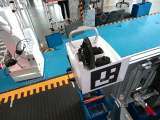} &
\includegraphics[width=0.245\linewidth]{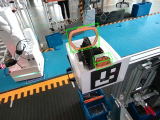} &
\includegraphics[width=0.245\linewidth]{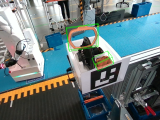} &
\includegraphics[width=0.245\linewidth]{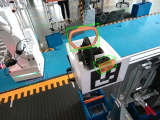} \\

\includegraphics[width=0.245\linewidth]{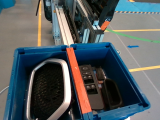} &
\includegraphics[width=0.245\linewidth]{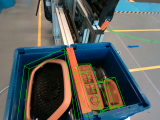} &
\includegraphics[width=0.245\linewidth]{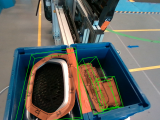} &
\includegraphics[width=0.245\linewidth]{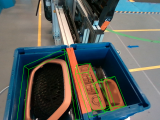} \\

\includegraphics[width=0.245\linewidth]{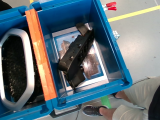} &
\includegraphics[width=0.245\linewidth]{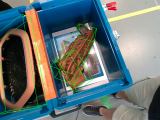} &
\includegraphics[width=0.245\linewidth]{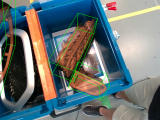} &
\includegraphics[width=0.245\linewidth]{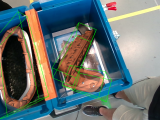} \\

Original & Ground Truth & NovaSplat (Ours) & FreeZe~\cite{Caraffa_ARXIV_2025} \\

\end{tabular}
\caption{Qualitative comparison on the IndustryShapes Classic dataset. Bounding boxes highlight the predicted 6D poses.}
\label{fig:is_classic_qualitative}
\end{figure*}

\begin{figure*}[t]
\centering
\setlength{\tabcolsep}{1pt}
\begin{tabular}{cccc}

\includegraphics[width=0.245\linewidth]{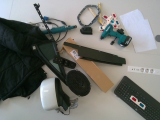} &
\includegraphics[width=0.245\linewidth]{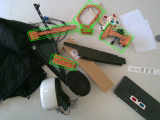} &
\includegraphics[width=0.245\linewidth]{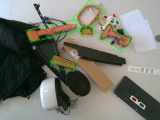} &
\includegraphics[width=0.245\linewidth]{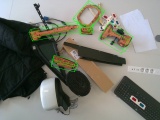} \\

\includegraphics[width=0.245\linewidth]{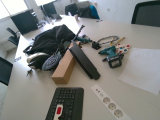} &
\includegraphics[width=0.245\linewidth]{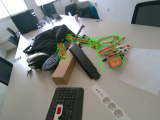} &
\includegraphics[width=0.245\linewidth]{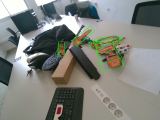} &
\includegraphics[width=0.245\linewidth]{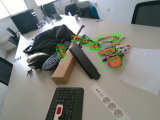} \\

\includegraphics[width=0.245\linewidth]{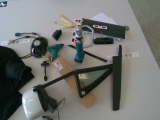} &
\includegraphics[width=0.245\linewidth]{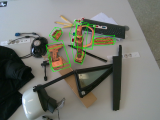} &
\includegraphics[width=0.245\linewidth]{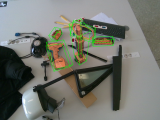} &
\includegraphics[width=0.245\linewidth]{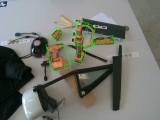} \\

\includegraphics[width=0.245\linewidth]{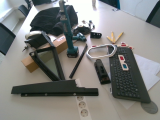} &
\includegraphics[width=0.245\linewidth]{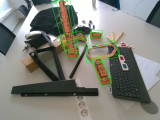} &
\includegraphics[width=0.245\linewidth]{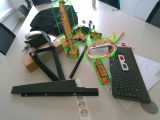} &
\includegraphics[width=0.245\linewidth]{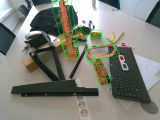} \\

\includegraphics[width=0.245\linewidth]{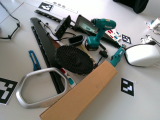} &
\includegraphics[width=0.245\linewidth]{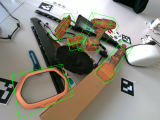} &
\includegraphics[width=0.245\linewidth]{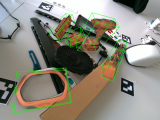} &
\includegraphics[width=0.245\linewidth]{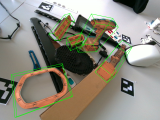} \\

Original & Ground Truth & NovaSplat (Ours) & FreeZe~\cite{Caraffa_ARXIV_2025} \\

\end{tabular}
\caption{Qualitative comparison on the IndustryShapes Extended dataset.  Bounding boxes highlight the predicted 6D poses.}
\label{fig:is_extended_qualitative}
\end{figure*}
}
\end{document}